%% file: main.tex
\documentclass{article} 
\usepackage[dvipsnames]{xcolor}
\usepackage{iclr2022_conference,times}

\input{math_commands.tex}

\usepackage{hyperref}
\usepackage{url}
\usepackage{soul}
\usepackage{makecell}
\usepackage{threeparttable,subcaption}
\usepackage{tikz}
\usepackage{multirow}
\usepackage{booktabs,wrapfig}
\usepackage{amsthm}
\usepackage{setspace}
\usepackage{parskip}
\usepackage{tcolorbox}

\newtcolorbox{mybasecolorbox}[1][]{%
  colback=gray!25, colframe=gray!25,
  coltitle=black, fonttitle=\bfseries,
  sharp corners,
  width=(\linewidth-30pt),
  title=#1}

\newenvironment{mytitlebox}[1][]{%
  \centering
  \mybasecolorbox[#1]
}{%
  \endmybasecolorbox
}

\newenvironment{example}{%
  \mytitlebox[\centering Correction note]
}{%
  \endmytitlebox
}

\usepackage{floatrow}
\newfloatcommand{capbtabbox}{table}[][\FBwidth]

\def\model{Time Control} 
\def\modelabbr{TC} 
\def\vae{Variational Auto-Encoder}
\def\vaeabbr{VAE} 
\def\infonce{Implicit Dynamics}
\def\infonceabbr{ID} 
\def\brownian{Brownian motion}
\def\brownianabbr{BM} 
\def\static{\textsc{Static}} 
\def\original{\textsc{Original}}

\usepackage{colortbl}
\newcommand\graycell{\cellcolor[rgb]{0.9,0.9,0.9}}

\title{Language modeling  via stochastic processes}

\author{
Rose E. Wang, Esin Durmus, Noah Goodman, Tatsunori B. Hashimoto
\\
Stanford University 
\\
\texttt{\{rewang, edurmus, ngoodman,thashim\}@cs.stanford.edu} \\
}

\newcommand{\revise}[1]{\textcolor{black}{ #1 }}

\iclrfinalcopy 
\begin{document}

\maketitle

\begin{abstract}
Modern language models can generate high-quality short texts. However, they often meander or are incoherent when generating longer texts. 
These issues arise from the next-token-only language modeling objective.
Recent work in self-supervised learning suggests that models can learn good latent representations via contrastive learning, which can be effective for discriminative tasks.
Our work analyzes the application of contrastive representations for generative tasks, like long text generation.
We propose one approach for leveraging constrastive representations, which we call \model\ (\modelabbr).  \modelabbr\ first learns a contrastive representation of the target text domain, then generates text by decoding from these representations.
Compared to domain-specific methods and fine-tuning GPT2 across a variety of text domains, \modelabbr\ performs competitively to methods specific for learning sentence representations on discourse coherence. On long text generation settings, \modelabbr\ preserves the text structure both in terms of ordering (up to $+15\%$ better) and text length consistency (up to $+90\%$ better)\footnote{Please find our code at \url{https://github.com/rosewang2008/language_modeling_via_stochastic_processes}

\begin{example}
    This is a revised version of the original ICLR 2022 paper. 
    During post-publication code review, we discovered that the original version of the code did not leverage goal-directedness during decoding.
    While we still find that contrastive representations lead to gains in our evaluations, this error affects other claims made in the paper on goal-directed decoding.
    To correct this and help improve our understanding of goal-directed decoding, this updated version of the manuscript contains results on both goal-directed and non goal-directed baselines. 
    We detail the difference between the original work and this updated work in Appendix~\ref{sec:version_diffs}. 
    The original version of this work can be found here: \url{https://arxiv.org/abs/2203.11370v1}.
\end{example}
}. 
\end{abstract}

\section{Introduction}
Large language models (LLM) such as GPT-2 have been extremely successful in text generation \citep{radford2019language,brown2020language}. 
However, LLMs are known to generate incoherent \textit{long} texts. One reason is that they are unable to plan ahead or represent long-range text dynamics \citep{kiddon2016globally,fan_strategies_2019,hua2020pair,10.3115/1073012.1073035,10.3115/1218955.1218966, tamkin2020language}. 
As a result, they oftentimes produce wandering content with poor discourse structure and low relevance  \citep{hua2020pair,zhao-etal-2017-learning,xu-etal-2020-discourse}; the text reads as if the model has no anchored goal when generating.
These problems with coherence are further exacerbated when forcing autoregressive models to generate \textit{longer} texts as the model struggles to extrapolate beyond its expected text end point. 
These problems suggest that LLMs currently fail to properly capture how documents evolve from beginning to end.
Doing so is critical for succeeding in goal-oriented tasks such as story, dialog or recipe generation. 

Prior work has explored the use of planning-based methods for generating globally coherent text \citep{kiddon2016globally,fan_strategies_2019,hua2020pair,10.3115/1073012.1073035,10.3115/1218955.1218966}. 
However, these methods rely on \textit{manually} defining text dynamics for specific domains. 
Other work has attempted to use sentence representations for modeling  text, such as with variational auto-encoders \citep{bowman2016generating} or contrastive learning \citep{gao2021simcse,devlin_bert_2019}.
However, most of these works focus on discriminative tasks like classifying the order of sentences. 
Methods including \citet{oord2019representation} have tried decoding from latent contrastive representations, however they fail to generate coherent long sequences.

We propose one way of leveraging contrastive learning representations for generation, which we call \model\ (\modelabbr).
We begin by assuming that meandering text generated without a goal can be represented as Brownian motion in latent space;
this motion enforces the embeddings of neighboring sentences to be similar to each other, whereas those of distant sentences to be dissimilar. 
Goal-directed behavior can be incorporated into this model by conditioning on a fixed start and end point. In this case, the Brownian motion becomes a Brownian bridge and the resulting latent trajectories abide by simple, closed-form dynamics.

We derive a novel contrastive objective for learning a latent space with Brownian bridge dynamics. 
We can then use this latent space to generate text that retains local coherence and has improved global coherence. 
To perform text generation, \model\ first generates a sequence of latent embeddings sampled from the Brownian bridge process pinned at a start and end point.
It then conditionally generates sentences using this latent sequence. 
In our work, we decode the latent sequences by fine-tuning GPT2 to generate text conditioned on \model's latent sequence. 

In summary, our work's contributions are the following:
\begin{itemize}
    \item We derive a novel contrastive learning objective based on Brownian bridge stochastic processes.
    \item We explore how contrastive representations can be used in both the discriminative and generative setting with our model \model\ which decodes from these learned representations.
    \item  Across a range of text domains, we show that \model\ generates more or equally coherent text on tasks including \revise {forced long text generation}, compared to task-specific methods. 
    \item We validate  that our latent representations capture text dynamics competitively by evaluating discourse coherence.  
    \item We ablate our method to understand the importance of the contrastive objective, enforcing Brownian bridge dynamics, explicitly modeling latent dynamics, and decoding from a latent plan. 
\end{itemize}

\section{Related works}

Generating long, coherent text is conceptually difficult for autoregressive models because they lack the ability to model text structure and dynamics \citep{lin-etal-2021-limitations}. 
This means that they struggle to plan and look ahead which leads to generating globally incoherent text.
Forcing autoregressive models to generate \textit{longer} texts exacerbates this incoherence because the models struggle to extrapolate beyond their expected text end point. 
Prior work has tried to address the problem of generating globally coherent text with  planning-based approaches \citep{puduppully2019data,moryossef-etal-2019-step,fan_strategies_2019,kiddon2016globally}. 
However, planning-based approaches rely on domain-specific heuristics for capturing text structure and dynamics.

Our work uses a contrastive objective to learn latent dynamics in text without domain-specific heuristics.
Contrastive objectives have been applied to several domains, including language \citep{devlin_bert_2019,iter_pretraining_2020,liu_simcls_2021}, vision \citep{chen2020simple}, and general time series data \citep{hyvarinen_unsupervised_2016,hyvarinen2019nonlinear}. In particular for language, contrastive objectives have been applied to the next-sentence prediction task for improving BERT embeddings \citep{devlin_bert_2019} and to the discourse coherence setting \citep{nie_dissent_2019,mchen-discoeval-19} for evaluating how coherent pairs of sentences are. 
\revise{However, they are not used for generation and are limited to classification tasks like discourse coherence.}
Prior work has also tried fitting latent variable models \citep{bowman2016generating}, however these generally result in poor language generation \citep{he2018lagging} or are domain-specific \citep{weber2020generating,arora2016latent}. 

Our work is closely related to Contrastive Predictive Coding (CPC) from \citet{oord2019representation}.
\revise{The key difference is CPC \textit{implicitly} learns \textit{unconditioned} latent dynamics, whereas our contrastive learning objective imposes goal-conditioned dynamics on our latent space.} 
Additionally, our method builds off of recent findings that contrastive objectives can be used to approximate local transition kernels of stochastic processes \citep{liu_contrastive_2021}.
The main difference between \citet{liu_contrastive_2021} and our work is that they focus on provable conditions for latent recovery; 
we focus on empirically effective methods that leverage similar insights for recovering latent representations from language.
Finally, our use of stochastic processes draws similarities to diffusion models \citep{song2020score, sohl2015deep} which apply a chain of diffusion steps onto the data and learn to reverse the diffusion process. However, our application is conceptually different: diffusion processes characterize properties of our latent space and are not a fixed inference method in our work.

\section{Methods}
\label{model}

\begin{figure*}[t]
    \centering
    \small
    \newcommand{\gw}{130mm}
    \includegraphics[width=\gw]{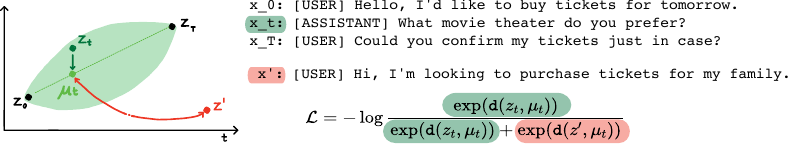}
    \caption{
    \small
    Latent space for a positive triplet of sentences $(x_0, x_t, x_T)$ that are part of the same conversation. 
    \revise{The encoder } maps positive triplets to a smooth Brownian bridge trajectory. It embeds $z_t$ close to the expected embedding $\mu_t$ pinned by $z_0, z_T$. 
    The green oval area illustrates the uncertainty over $z_t$ as a function of how close $t$ is to $0$ and $T$.
    In contrast, a negative random sentence $x'$ from a different conversation is not coherent with $x_0$ and $x_T$; thus, it is embedded far from $\mu_t$. This is captured by our contrastive loss, $\mathcal{L}$. 
    }
    \label{fig:model}
\end{figure*}

This section details our novel contrastive learning objective based on Brownian bridge dynamics.
First, we discuss training an encoder via our contrastive learning objective to map sentences to a Brownian bridge \citep{revuz2013continuous} latent space.
Next, we discuss how a decoder learns to reconstruct sentences from this latent space.
Finally, we discuss how to generate text at inference time.

\subsection{Training an encoder with Brownian bridge dynamics \label{sec:encoding}}
Our encoder is a nonlinear mapping from raw input space to latent space,  $f_{\theta}: \mathcal{X} \rightarrow \mathcal{Z}$. 
The objective for the encoder is to map high-dimensional sequential data into low-dimensional latents which follow a stochastic process of interest---in this paper, it is the Brownian bridge process. The density of a Brownian bridge process between an arbitrary start point $z_{0}$ at $t=0$ and end point $z_{T}$ at $t=T$ is,

\begin{equation}
    p(z_{t}|z_{0}, z_{T}) = \mathcal{N}\bigg( \Big(1-\frac{t}{T}\Big)z_0 + \frac{t}{T}z_T, \frac{t\big(T-t\big)}{T}\bigg). \label{eq:transition}
\end{equation}
This density is intuitive to understand: It acts like a noisy linear interpolation between the start and end point of the trajectory, where $z_t$ should be more like $z_0$ at the start and more like $z_T$ at the end of the trajectory.
Uncertainty is highest in the middle region, and low near the end points (rf. Figure~\ref{fig:model}).

Consider a set of triplet observations, $(x_1, x_2, x_3)$. The goal of our work is to ensure that $f_{\theta}(x_1), f_{\theta}(x_2), f_{\theta}(x_3)$ follow the Brownian bridge transition density in Equation~\ref{eq:transition}. We ensure this using a contrastive objective.
Formally, given multiple sequences of data points, $X = \{x_1, ..., x_N\}$, we draw batches consisting of randomly sampled positive triplets $x_{0}, x_{t}, x_{T}$ where $0 < t < T$: $\mathcal{B}= \{(x_{0}, x_{t}, x_{T})\}$.\footnote{We use indices $0, t, T$ to denote the start, middle and end point of a Brownian bridge, but these do \textit{not} correspond to strictly sampling the first, middle and last sentence of a document. $x_0, x_t, x_T$ can be \textit{any} sentence in a document as long as $x_0$ comes before $x_t$ and $x_t$ before $x_T$ in the document.} 
Our encoder is optimized by, 

\begin{align}
    \mathcal{L}_N &= \mathbb{E}_{X}\left[-\log \frac{\exp(\text{d}(x_{0}, x_{t}, x_{T}; f_{\theta}))}{\sum\limits_{(x_0, x_{t'}, x_T) \in \mathcal{B}^{}} \exp(\text{d}(x_{0}, x_{t'}, x_{T}; f_{\theta}))}\right], \text{where} \label{eq:loss_objective} \\ 
    \quad \text{d}(x_{0}, x_{t}, x_{T}; f_{\theta}) &= -\frac{1}{2\sigma^2}\bigg\|\underbrace{f_{\theta}(x_t)}_{z_t}- \underbrace{\left(1 - \frac{t}{T}\right)f_{\theta}(x_0) -  \frac{t}{T}f_{\theta}(x_T)}_{ \text{mean in Equation~\ref{eq:transition}}} \bigg\|^2_2 \label{eq:distance_function}
\end{align}

$\sigma^2$ is the variance in Equation~\ref{eq:transition}: $\frac{t(T-t)}{T}$. 
Note that Equation~\ref{eq:loss_objective} sums over negative middle contrasts, $x_{t'}$.
This objective can be viewed as maximizing the extent to which true triplets from the data follow the Brownian bridge process while minimizing the extent to which an alternative mid-point sampled from another sequence does so. \footnote{Empirically, we found Brownian bridge dynamics easier to recover with triplets rather than pairs of contrasts. Appendix~\ref{sec:exploratory_t} discusses some of the pair-wise contrast results.}

Figure~\ref{fig:model} illustrates how the objective translates into the language setting for training the encoder.
The objective samples triplet sentences from a document. 
Sentences drawn from the same document make up a smooth latent trajectory; they should be close to each other and follow the conditional density in latent space.
Sentences drawn from different documents should not make up a smooth trajectory and should less likely follow bridge dynamics.

\paragraph{Connection to mutual information estimation and triplet classification}
We draw connections between our contrastive loss and the mutual information estimation setup from \citet{oord2019representation, poole2019variational} (as $|\mathcal{B}| \rightarrow \infty$) and the classification setup from \citet{liu_contrastive_2021} ($|\mathcal{B}| = 1$). 

Following \citet{oord2019representation, poole2019variational}, this objective can be seen as a lower bound on the mutual information between the two end points and the middle point: $I(X_t, \{X_0, X_T\}) \geq \log (N) - \mathcal{L}_N$. Hence, by minimizing the contrastive loss, we are maximizing the amount of information between the trajectory and the linear interpolation of its end points.

Assuming $|\mathcal{B}|=1$, we can draw a connection to the classification setup studied in  \citet{liu_contrastive_2021}. They train a classifier to distinguish in- vs. out-of-order input pairs and show that the Bayes optimal logits for pair-wise classification can be written as a function of the stochastic process transition kernel.
This is equivalent to the our loss on a single triplet $i$:  $ l_i = -\log \frac{\exp(\text{d}(x_{0}, x_{t}, x_{T}; f_{\theta}))}{\exp(\text{d}(x_{0}, x_{t}, x_{T}; f_{\theta})) + \exp(\text{d}(x_{0}, x_{t'}, x_{T}; f_{\theta}))}$. 
\citet{liu_contrastive_2021} consider pairs whereas our work considers triplets; we show in Appendix~\ref{sec:triplet_classification} the pairwise and triplet setups are equivalent.

\subsection{Training a decoder with latent plans \label{sec:decoding}}
Here we discuss how to train a language model to decode latent sequences for generation.
We first map all the sentences in the training dataset to our learned latent space using the pretrained encoder $f_{\theta}$. 
This gives us a Brownian bridge trajectory of sentence-level latent codes $(z_0, \ldots, z_t, \ldots, z_T)$ for a document in the dataset.
Then, rather than learning a decoder from scratch, we fine-tune GPT2 \citep{radford2019language} to generate text conditioned on past context and the latent plan.

We fine-tune in the following manner. 
Let $x_1 \ldots x_W$ be a document with $W$ tokens and $T$ sentences used to train the decoder. Using the encoder $f_{\theta}$, we can obtain embeddings $z_1 \ldots z_T$ for each sentence. The decoder is a standard auto-regressive language model that is modified in the following way: at time $t$, the decoder must predict $x_{t}$ using all tokens in the past $x_{<t}$, as well as the sentence embedding $z_{s_t}$, where the index $s_{t} \in [T]$ is a map which takes each token to its corresponding sentence. This is a form of a reconstruction objective, as the identity of $x_t$ is encoded in $z_{s_t}$.

\subsection{Generating text with latent plans at inference time \label{sec:generation}}

\begin{figure*}[t]
    \centering
    \small
    \newcommand{\gw}{130mm}
    \includegraphics[width=\gw]{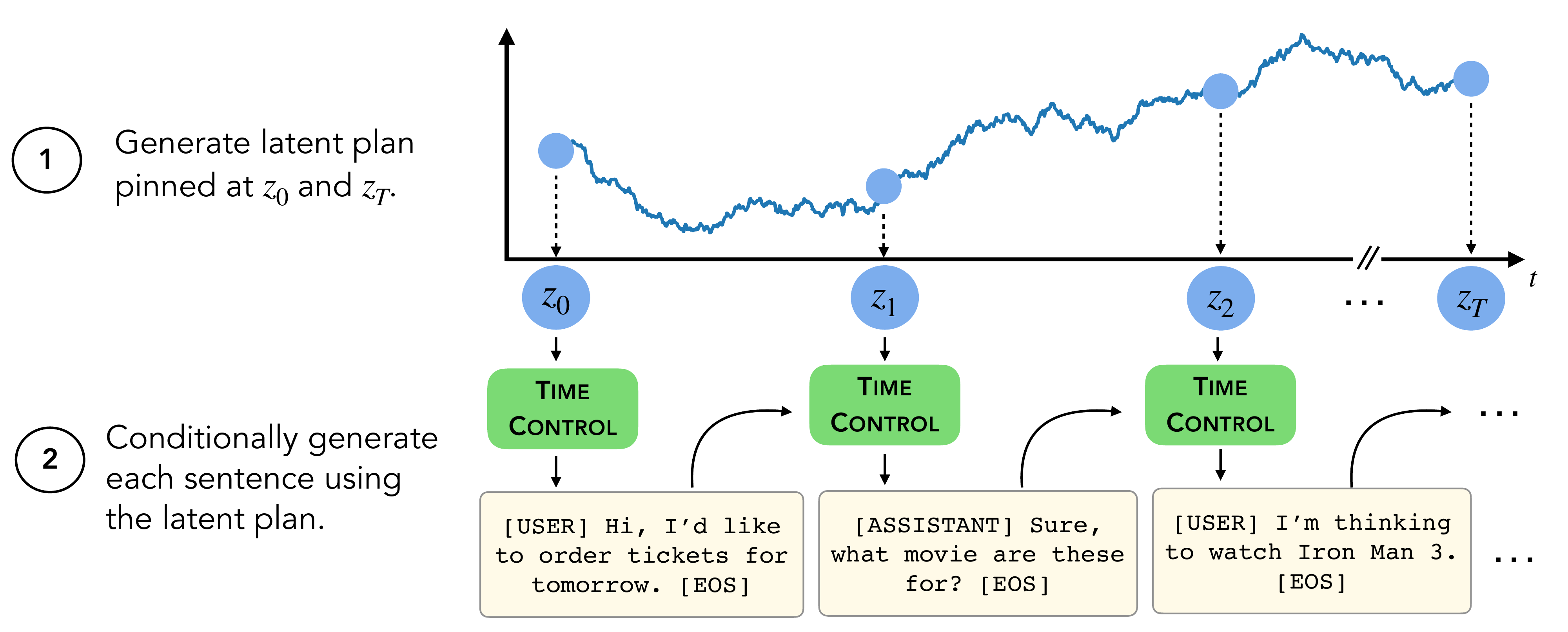}
    \caption{
    \small
    \model\ generates text conditioned on a latent plan.
    A latent plan is first generated by running Brownian bridge dynamics pinned between a sampled start  $z_0$ and goal latent variable $z_T$ forward.
    A decoder then conditionally generates from this latent plan on a sentence-level.
    }
    \label{fig:generation}
\end{figure*}

Figure~\ref{fig:generation} illustrates how the trained decoder generates text at inference time. 
Given two end points $z_0, z_T$, we sample a trajectory from a latent Brownian bridge, and then generate from the decoder conditioned on this bridge. 
In many situations, we may not know the endpoints of the Brownian bridge explicitly. 
In this case, we encode a set of sentences corresponding to start and end points (eg. the first and last sentences of our training set), and fit a Gaussian to these points to form a density estimate. 
Generating in this case involves first sampling from the Gaussian, and then generating as before from the bridge.
More details on training and generation can be found in Appendix~\ref{sec:training}.

\section{Experiments}
\label{sec:experiments}

We now evaluate the ability of a model trained with a contrastive objective to capture text dynamics. Specifically, we aim to answer the following research questions (RQ):

\begin{description}
    \item[RQ1:] \emph{Can the contrastive learning objective model local text dynamics?}
    Section~\ref{sec:model_local_text_dynamics} investigates this question using a sentence ordering prediction task: given two sentences from the same document, we evaluate whether different models can predict their original order.
    \item[RQ2:] \emph{Can the contrastive learning objective model global text dynamics?} Section~\ref{sec:model_global_text_dynamics}  investigates this question on text generation for Wikipedia city articles by examining the length of generated sections.
    \item[RQ3:] \emph{Can the contrastive learning objective help generate long coherent documents?} Section~\ref{sec:generate_global_text} investigates this question on forced long text generation: we evaluate how well models preserve global text statistics (such as typical section orders and lengths) when forced to extrapolate during generation.
\end{description}

We run our setup with different latent dimensions ($d=8,16,32$).
Our encoder architecture is a frozen, pretrained GPT2 model from Huggingface \citep{radford2019language, wolf-etal-2020-transformers} and a trainable MLP network.
We extract GPT2's last layer hidden state that corresponds to the end-of-sentence (EOS) token and train the 4-layer MLP on top of the hidden state. 
The MLP network has intermediate ReLU activations and is trained with stochastic gradient descent with a learning rate of 1e-4 and with momentum 0.9.
For more details on the machines we used for our experiments, please refer to Appendix~\ref{sec:machine_training}.
This version of the manuscript contains results after correcting for random seeding and GPU nondeterminism. Please refer to Appendix~\ref{sec:version_diffs} for details on the version differences.

\paragraph{Encoder ablations} We perform three ablations on our encoder model. 
Recall that \model\ encoder \textbf{(A) explicitly} models latent structure with \textbf{(B) Brownian bridge dynamics} using a \textbf{(C) contrastive} loss.
\textbf{(A)} replaces explicit dynamics with \infonce\ (\infonceabbr) where future latents are directly predicted with an autoregressive model \citep{oord2019representation}. 
\textbf{(B)} replaces Brownian bridge dynamics with \brownian\ (\brownianabbr): latents follow the transition density $ z_t | z_s  \sim \mathcal{N}(z_s, t-s)$ in Equation~\ref{eq:distance_function}. Note $z_t$ is centered at $z_s$ and is not conditioned on a goal end-point.
\textbf{(C)} replaces the contrastive loss with a \vae\ (\vaeabbr) and centers the priors over $z_0$ to 0 and $z_T$ to 1, as done in our setup. Appendix~\ref{sec:app_ablations} includes more detail on the ablations.

\paragraph{Decoder instantiations}
We compare two methods for decoding from contrastive representations. 
The first method decodes from the path $[z_0, z_1, \dots, z_T]$ as detailed in Section~\ref{sec:generation} by generated by sampling from Brownian bridge or Brownian motion for \brownianabbr. 
The second method---which we will call \static---decodes from repititions of the initial latent $[z_0, z_0, \dots, z_0 ]$.

\paragraph{Datasets}
We use language datasets that elicit different kinds of structure, from section structure to discourse structure to narrative structure. 
\model\ does not take in any information about the structure, treating each domain the same under its encoding objective.
More information and dataset examples are provided in Appendix~\ref{sec:dataset_meta}.
\textbf{Wikisection} \citep{arnold2019sector} includes Wikipedia articles on cities split by sections. We adapt this dataset such that each article contains four ordered sections (abstract, history, geography, demographics) marked with section id tokens: Each article is represented as, ``[ABSTRACT] text [HISTORY] text [GEOGRAPHY] text [DEMOGRAPHICS] text''.
\textbf{Wikihow} (WH) \citep{koupaee2018wikihow} contains how-to articles organized by a title, method, and steps. We mark each with its own section id tokens: Each article is represented as ``[TITLE] text [METHOD] text [STEP] 1 text [STEP] 2 text ...'' 
\textbf{Recipe NLG} \citep{bien-etal-2020-recipenlg} contains recipes, each with a title, ingredients and set of directions. A recipe is constructed as ``[TITLE] text [INGREDIENTS] text [DIRECTIONS] text''.
\textbf{Taskmaster-2} (TM-2) \citep{byrne2019taskmaster} contains conversations on finding restaurants between an assistant and a user. The assistant's turn is marked with an ``[ASSISTANT]'' tag, and the user's turn is marked with a ``[USER]'' tag.
\textbf{TicketTalk} \citep{byrne-etal-2021-tickettalk} contains conversations on booking movie tickets between an assistant and a user. The assistant's and user's turns are similarly marked as in TM-2.

\subsection{Modeling local text dynamics \label{sec:model_local_text_dynamics}}

\begin{spacing}{0.5}
\begin{table*}[t]
  \centering
  \small
  \begin{tabular}{l|cc|cc|cc}
    \toprule
    \multicolumn{1}{c}{\bf }  &\multicolumn{2}{c}{\bf Wikisection} &\multicolumn{2}{c}{\bf TM-2} &\multicolumn{2}{c}{\bf TicketTalk}
\\ 
    \multicolumn{1}{c}{\bf Method}  &\multicolumn{1}{c}{$k=5$} &\multicolumn{1}{c}{$k=10$} &\multicolumn{1}{c}{$k=5$}&\multicolumn{1}{c}{$k=10$} &\multicolumn{1}{c}{$k=5$}&\multicolumn{1}{c}{$k=10$} \\
    \midrule
    BERT & $ \graycell \mathbf{72.7 \pm 0.1}$ &  $80.4 \pm 0.2 $ & $ 75.6 \pm 0.0 $ & $ 86.7 \pm 0.0 $ & $73.1 \pm 0.2$ & $ 86.4 \pm 0.0 $ \\
    ALBERT & $ 67.7 \pm 15.1$ &  $\graycell \mathbf{ 82.3 \pm 0.3}$ & $ \graycell \mathbf{83.9 \pm 1.7 }$ & $\graycell \mathbf{ 92.2 \pm 0.9 } $ &  $ 74.5 \pm 9.8$ &  $ \graycell \mathbf{93.3 \pm 0.6 }$ \\
    S-BERT  & $ 71.8 \pm 0.1$ & $ 79.0 \pm 0.1$  & $ 75.6 \pm 0.0 $ & $ 86.7 \pm 0.0 $ & $ 76.2 \pm 0.1 $ & $ 88.8 \pm 0.0 $   \\
    Sim-CSE  & $72.6 \pm 0.1$ & $ 79.5 \pm 0.1$ 	& $ 77.8 \pm 0.0 $ & $ 88.3 \pm 0.0 $ & $ \graycell \mathbf{78.0 \pm 0.0} $ & $ 90.1 \pm 0.1 $   \\
    \midrule\midrule
    GPT2  & $ 58.4 \pm 3.0$  & $67.1 \pm 2.3$ & $ 66.8 \pm 0.9 $ & $ 75.6 \pm 0.0 $ & $55.2 \pm 1.9$  & $78.0 \pm 2.7$  \\  
    \midrule
    \vaeabbr\ (8) & $50.0 \pm 3.8$  & $50.3 \pm 0.8$ & $ 49.4 \pm 1.1 $  & $ 49.9 \pm 1.7 $ &  $ 49.0 \pm 2.6$ & $ 49.0 \pm 5.7$ \\
    \vaeabbr\ (16) & $ 51.9 \pm 1.3$ & $ 55.3 \pm 3.2$ & $ 52.9 \pm 4.2 $ & $ 56.8 \pm 6.5$ &  $ 51.9 \pm 4.2 $ &  $56.3 \pm 5.5 $   \\
    \vaeabbr\ (32) & $ 52.3 \pm 3.0 $ & $ 50.7 \pm 2.3$ &  $ 55.7 \pm 1.3  $  & $ 64.1 \pm 2.3 $ & $59.3 \pm 1.5 $ & $ 69.5 \pm 1.0 $   \\
    \midrule
    \infonceabbr\ (8) & $ 58.2 \pm 0.4$  & $ 64.7 \pm 1.6  $ & $ 62.4 \pm 0.6 $  & $ 71.0 \pm 0.5$ &  $ 54.3 \pm 1.0$ &  $ 64.7 \pm 0.5$   \\
    \infonceabbr\ (16) & $ 68.5 \pm 0.1 $ & $ 76.9 \pm 0.1 $  & $ 62.7 \pm 0.1 $ & $ 71.1 \pm 0.1 $ &  $ 53.0 \pm 0.2$ &  $ 66.2 \pm 0.1 $  \\
    \infonceabbr\ (32) & $70.1 \pm 0.1$ & $ 77.1 \pm 0.1 $ & $ 62.6 \pm 0.1 $ & $71.1 \pm 0.0$ &  $ 55.4 \pm 0.2 $ &  $ 69.4 \pm 0.1 $  \\
    \midrule
    \brownianabbr\ (8) &  $ 49.4 \pm 3.2$ & $ 56.7 \pm 6.8$ & $ 52.1 \pm 3.1 $ & $ 55.0 \pm 5.1 $ &  $ 47.8 \pm 2.8 $ &   $ 51.2 \pm 3.2 $ \\
    \brownianabbr\ (16) & $ 52.8 \pm 0.8$ & $ 56.3 \pm 0.9$ &  $ 51.5 \pm 1.1$ & $52.8 \pm 2.3 $ &  $ 55.2 \pm 5.5 $ &  $62.4 \pm 5.6$  \\
    \brownianabbr\ (32) & $ 51.8 \pm 1.9 $ & $ 55.9 \pm 4.3 $ & $ 50.4 \pm 1.0$ & $ 51.3 \pm 2.2$ &  $ 55.0 \pm 2.6 $ &  $ 69.6 \pm 2.6 $  \\
    \midrule
    \modelabbr\ (8) & $ 69.3 \pm 0.1 $ & $ 77.6 \pm 0.0$ &  $ 78.4 \pm 0.0 $ & $ 88.7 \pm 0.0$ & $\bf \underline{74.7 \pm 0.1} $ & $ 87.2 \pm 0.0 $  \\
    \modelabbr\ (16) & $\bf \underline{71.7 \pm 0.0 }$  & $ 76.8 \pm 0.0$  &  $ 78.7 \pm 0.0 $ & $ 89.0 \pm 0.0 $ & $\bf \underline{74.7 \pm 0.0} $ & $ 87.2 \pm 0.1 $ \\
    \modelabbr\ (32) & $ 71.6 \pm 0.0$  & $\bf \underline{77.3 \pm 0.0}$ & $\bf \underline{ 79.2 \pm 0.0 }$ & $\bf \underline{89.1 \pm 0.0}$  & $ 74.0 \pm 0.0 $ & $ \bf \underline{ 87.3 \pm 0.0 }$  \\
    \bottomrule
  \end{tabular}
  \vspace{5pt}
  \caption{\small  Discourse coherence accuracy measured by the test accuracy of the trained linear classifier, reporting $\mu \pm $ standard error over 3 runs. Random accuracy is 50\%. 
  \revise{
  The highest mean score for the \textit{non-GPT2}-based methods (rows above the double lines) are marked in gray cells. 
  The highest mean score for the \textit{GPT2}-based methods (rows below the double lines) are underlined. }
  When applicable, the methods are run with varying latent dimensions marked in parentheses (dimension).}
  \label{tab:discourse_results}
\end{table*}
\end{spacing}

\revise{We evaluate how well the Brownian bridge contrastive learning objective models local text dynamics (\textbf{RQ1}) on the discourse coherence setting \citep{jurafsky2000speech}.}
Discourse coherence is often measured by how well representations capture discourse structure by testing for whether a linear classifier can detect in-order and vs. out-of-order sentence pairs \citep{chen_evaluation_2019}. 
We compare \model's encoder against GPT2's last layer's hidden state corresponding to the EOS token \citep{radford2019language}, BERT \citep{devlin_bert_2019}, ALBERT \citep{lan2019albert}, Sentence BERT \citep{reimers2019sentence}, and SimCSE \citep{gao2021simcse}. 
The latter 4 methods are designed as sentence embedding models. We also compare to our ablations.

The setup is the following: The encoder takes in two sentences $x_t, x_{t+k}$ to produce their latents $z_{t}, z_{t+k} \in \mathbb{R}^d$. 
At random, the latents are fed either in- or out-of-order. A linear classifier is trained on 100 epochs with stochastic gradient descent with a learning rate of 1e-4 and with momentum 0.9.
We varied the sentence distance $k \in \{1, 5, 10\}$ and found that on some domains and for $k=1$, all methods scored near random accuracy; we have omitted those results in the main paper. 
Otherwise, the results are summarized in Table~\ref{tab:discourse_results} where we report the mean and standard error accuracy on a held-out discourse dataset of 3000 examples on 3 runs. 
Our method is able to compete with sentence embedding specific methods, like ALBERT, BERT, SimCSE.
Additionally, though GPT2 is used as a base encoder for \model, we observe that \model\ greatly improves upon GPT2 with significant gains on TM-2, TicketTalk, and Wikisection (up to $+20\%$). 
\revise{Out of all the GPT2-based methods, it performs the best.}

Neither \vaeabbr\ nor \brownianabbr\ perform better than random accuracy on \revise{most of the} domains. This suggests that the variational lower bound does not recover the latent embeddings as well as contrastive objectives and the choice of stochastic processes matter for learning informative structural embeddings.

\revise{Overall}, our model performs \revise{competitively on } both \revise{task-oriented conversations like TM-2 and TicketTalk and enumerative, factual domains like Wikisection}. This answers \textbf{RQ1} in the positive: \model\ can model local text dynamics, like in conversations and articles.

\subsection{Modeling global text dynamics} \label{sec:model_global_text_dynamics}

\begin{minipage}[ht]{\textwidth}
    \small
    \centering
    \begin{spacing}{0.7}
    \begin{minipage}[b]{\textwidth}
    \centering
  \begin{tabular}{l|c}
    \toprule
    \multicolumn{1}{c}{\bf Method}  &\multicolumn{1}{c}{\bf MM \% ($\downarrow$)}
\\ 
    \midrule\midrule
    GPT2  & \revise{$9.9 \pm 3.6$} \\ 
    \revise{SD}  & \revise{$28.3 \pm 7.9$}  \\ 
    \revise{SS}  & \revise{$223.9 \pm 18.5$}  \\ 
    \midrule
    \vaeabbr\ (8) & \revise{$ 8.1 \pm 3.0$} \\
    \vaeabbr\ (16)  & \revise{$14.1 \pm 7.8$}  \\
    \vaeabbr\ (32) & \revise{$18.5 \pm 4.5$}  \\ 
    \midrule
    \revise{\static{}  \vaeabbr\ (8)} & \revise{$\underline{8.9 \pm 2.2}$} \\
    \revise{\static{} \vaeabbr\ (16)}  & \revise{$13.1 \pm 5.8$}  \\
    \revise{\static{} \vaeabbr\ (32)} & \revise{$\bf \underline{6.7 \pm 1.9}$}  \\ 
    \midrule
    \infonceabbr\ (8) &  \revise{$13.3 \pm 3.8$} \\
    \infonceabbr\ (16)  & \revise{$61.8 \pm 5.5$ } \\
    \infonceabbr\ (32) & \revise{$84.6 \pm 11.0$}  \\ 
    \midrule
    \static{} \infonceabbr\ (8) &  \revise{$12.3 \pm 3.4$} \\
    \static{} \infonceabbr\ (16)  & \revise{$ 77.1 \pm 4.4$ } \\
    \static{} \infonceabbr\ (32) & \revise{$ 79.6 \pm 8.0 $}  \\ 
    \midrule
    \brownianabbr\ (8) &  \revise{$16.2 \pm 11.4$} \\
    \brownianabbr\ (16)  & \revise{$21.1 \pm 11.8$ } \\
    \brownianabbr\ (32) & \revise{$51.6 \pm 14.2$ } \\ 
    \midrule
    \static{} \brownianabbr\ (8) &  \revise{$15.4 \pm 9.2$} \\
    \static{} \brownianabbr\ (16)  & \revise{$11.8 \pm 5.4 $ } \\
    \static{} \brownianabbr\ (32) & \revise{$ 13.0 \pm 5.0$ } \\ 
    \midrule 
    \modelabbr\ (8) & \revise{$ 13.1 \pm 4.4$} \\
    \modelabbr\ (16) & \revise{ $15.4 \pm 10.1 $ }\\
    \modelabbr\ (32) &\revise{ $28.1 \pm 12.2$} \\
    \midrule
    \static{} \modelabbr\ (8) & \revise{$12.8 \pm 3.2$} \\
    \static{} \modelabbr\ (16) & \revise{ $13.0 \pm 7.5$ }\\
    \static{} \modelabbr\ (32) &\revise{ $13.0 \pm 7.8$} \\
    \bottomrule
  \end{tabular}
      \captionof{table}{\small Percentage of length mismatch (MM) during generation. \revise{These results are run over 3 random GPU machines, each with 3 seeds.
      The method with the highest mean is marked in bold, and methods with overlapping confidence intervals are underlined.
      } \label{tab:length_matching}}
    \end{minipage}
\end{spacing}
\end{minipage}

\revise{We evaluate how well the Brownian bridge contrastive learning objective models global text dynamics (\revise{\textbf{RQ2}}) by assessing whether the methods mimic document structure on Wikisection.}
We check whether the generated section lengths match with the average lengths in the dataset.
We focus on Wikisection because it is the only dataset with long, well-defined  sections (rf.  Appendix~\ref{sec:dataset_meta} on dataset statistics).
Each document contains an abstract, history, geography, and demographics section on a city. 

\model\ plans a latent trajectory by running the bridge process between the start and end latent, $z_0 \sim p(z_{0})$ and $z_T \sim  p(z_T)$. 
We compare against fine-tuned GPT2. We also include two oracle methods that are fine-tuned GPT2 models with additional section-embedding supervision. One is ``sec-dense GPT2'' (SD) where each token's embedding contains the current section identity; section $i$'s embedding is added onto the token's positional token embedding. The other is ``sec-sparse GPT2'' (SS) where the token embedding contains an indicator that is $1$ if the token is the start of a new section, and $0$ otherwise.
For \vaeabbr\ and \infonceabbr, we calculate the density estimate $p(z_{0})$ and $p(z_T)$ and run a linear interpolation between the start and end latents.
For \brownianabbr, we calculate the density estimate $p(z_{0})$ and run Brownian motion.
Additionally, we run \static\ for \model, \vaeabbr, \infonceabbr, \brownianabbr\ which samples $z_0 \sim p(z_{0})$ and repeatedly decodes from this latent representation. Note that the difference between these \static\ models is that the decoders are finetuned with their respective objectives (Section~\ref{sec:experiments}).

Table~\ref{tab:length_matching} reports in percentage how much the generated section lengths deviate from the average section lengths. 
\vaeabbr\ best matches the section lengths out of all the methods, despite performing at near random accuracy on discourse coherence.
We hypothesize that \modelabbr\ does not match the text lengths as well as the \vaeabbr\ method, as the \vaeabbr\ is more explicitly trained to match the text distribution.
Nonetheless, \modelabbr\ and \static{} \modelabbr\ perform competitively with other top-performing baselines like \vaeabbr\ and \static{} \vaeabbr.
Another surprising observation was that fine-tuned GPT2, SD and SS mismatch the section lengths by $10$-$224\%$. 
Upon further investigation, we noticed the models overshoot short sections and undershoots long sections; this causes the section length deviations. 
\infonceabbr\ and \static{} \infonceabbr\ performs extremely poorly in contrast to these methods; this highlights the challenge of interpolating on learned dynamics models which is exacerbated in the long text generation settings. 
The results affirm that \revise{contrastive objectives can model global text dynamics, such as in matching document structure, however probabilistic objectives like \vaeabbr\ might be a better alternative learning objective.}

\subsection{Generating globally coherent text \label{sec:generate_global_text}}

\begin{minipage}[t]{\textwidth}
    \small
    \centering
    \begin{spacing}{0.7}
    \vspace{1em}
  \begin{minipage}[b]{\textwidth}
    \centering
  \begin{tabular}{lc|c|c}
    \toprule
    \multicolumn{1}{c}{\bf Method}  &\multicolumn{1}{c}{WH } &\multicolumn{1}{c}{TM-2} &\multicolumn{1}{c}{ TT}\\
    \midrule\midrule
    GPT2    & \revise{$\underline{10.6 \pm 4.3} $}  & \revise{$81.7 \pm 10.7$}  & \revise{$108.5 \pm 8.6$}    \\
    \midrule
    \vaeabbr\ (8) &  \revise{$\underline{10.8 \pm 6.3}$}  & \revise{ $120.4 \pm 36.3$}  &  \revise{$80.9 \pm 7.1$ }   \\
    \revise{  \vaeabbr\ (16)} & \revise{ $13.2 \pm 6.4$}   & \revise{ $109.4 \pm 51.2$}  &  \revise{$75.0 \pm 4.5$ }   \\
    \vaeabbr\ (32) & \revise{ $\underline{11.0 \pm 1.9}$}  & \revise{ $66.9 \pm 10.8$} & \revise{ $77.5 \pm 25.0$}   \\
    \midrule
    \revise{\static{} \vaeabbr\ (8)} &  \revise{$8.6 \pm 4.4$}  & \revise{ $123.8 \pm 35.3$}  &  \revise{$81.4 \pm 5.7$ }   \\
    \revise{ \static{}   \vaeabbr\ (16)} & \revise{ $12.9 \pm 5.3$}   & \revise{ $115.5 \pm 60.6$}  &  \revise{$75.2 \pm 3.2$ }   \\
    \revise{ \static{}   \vaeabbr\ (32)} & \revise{ $12.8 \pm 5.3$}  & \revise{ $73.9 \pm 10.9$} & \revise{ $80.2 \pm 23.9$}   \\
    \midrule
    \infonceabbr\ (8) & \revise{$14.9 \pm 6.5$}  & \revise{ $200.3 \pm 27.7$ } & \revise{ $108.8 \pm 9.5$ }  \\
    \infonceabbr\ (16) & \revise{ $ 37.7 \pm 4.0$}   &  \revise{ $198.2 \pm 19.2$}  & \revise{ $68.7 \pm 20.3$  }  \\
    \infonceabbr\ (32) &  \revise{$35.9 \pm 8.5$}  &  \revise{$231.5 \pm 97.1$ }&  \revise{$78.6 \pm 22.5$ }  \\
    \midrule
    \revise{\static{}   \infonceabbr\ (8)} & \revise{$ 13.5 \pm 5.6$}  & \revise{ $120.2 \pm 37.4$ } & \revise{ $ 64.1 \pm 6.0$ }  \\
    \revise{\static{}  \infonceabbr\ (16)} & \revise{ $37.2 \pm 2.7$}   &  \revise{ $93.9 \pm 32.3$}  & \revise{ $66.2 \pm 20.5$  }  \\
    \revise{\static{}  \infonceabbr\ (32)} &  \revise{$25.8 \pm 4.4$}  &  \revise{$124.4 \pm 50.4$ }&  \revise{$ 58.9 \pm 16.9$ }  \\
    \midrule
    \brownianabbr\ (8) & \revise{ $11.8 \pm 5.1$} & \revise{ $93.4 \pm 20.4$ } & \revise{ $68.2 \pm 28.9$    }  \\
    \brownianabbr\ (16) &  \revise{ $20.8 \pm 11.5$} &  \revise{$106.8 \pm 17.6$}   & \revise{ $111.3 \pm 45.6$ } \\
    \brownianabbr\ (32) &  \revise{$30.2 \pm 13.4$}  & \revise{ $118.6 \pm 17.7$ } & \revise{ $170.5 \pm 36.2$  } \\
    \midrule
    \static{} \brownianabbr\ (8) & \revise{ $\bf \underline{7.8 \pm 3.2}$} & \revise{ $102.5 \pm 14.6$ } & \revise{ $66.3 \pm 26.2$    }  \\
    \static{} \brownianabbr\ (16) &  \revise{ $\underline{10.7 \pm 6.6}$} &  \revise{$93.7 \pm 11.7$}   & \revise{ $66.2 \pm 34.7$ } \\
    \static{} \brownianabbr\ (32) &  \revise{$14.1 \pm 9.7$}  & \revise{ $111.5 \pm 17.2$ } & \revise{ $82.7 \pm 25.9$  } \\
    \midrule
    \modelabbr\ (8) & \revise{ $13.5 \pm 6.1$}  &  \revise{$41.9 \pm 9.5$}  &  \revise{$69.8 \pm 6.2$ }     \\
    \modelabbr\ (16) & \revise{$ 8.5 \pm 4.4$} & \revise{$21.5 \pm 4.7$}  &\revise{ $60.8 \pm 16.6$ }   \\
    \modelabbr\ (32) &\revise{ $13.3 \pm 3.6$}   &  \revise{$33.2 \pm 3.3$} &\revise{ $53.5 \pm 27.4$}    \\
    \midrule
    \static{} \modelabbr\ (8) & \revise{ $13.4 \pm 5.5$}  &  \revise{$15.6 \pm 2.7$}  &  \revise{$\underline{36.4 \pm 11.4}$ }     \\
    \static{} \modelabbr\ (16) & \revise{$\underline{10.8 \pm 5.4}$} & \revise{$\underline{13.7 \pm 3.4}$}  &\revise{ $\underline{37.4 \pm 12.9}$ }   \\
    \static{} \modelabbr\ (32) &\revise{ $\underline{10.3 \pm 7.1}$}   &  \revise{$\bf \underline{8.6 \pm 5.3}$} &\revise{ $\bf \underline{30.2 \pm 17.7}$}    \\
    \bottomrule
  \end{tabular}
      \captionof{table}{\small Section lengths deviating from expected length in forced long text generation reported in \% ($\downarrow$).  \revise{These results are run over 3 random GPU machines, each with 3 seeds.
      The method with the highest mean is marked in bold, and methods with overlapping confidence intervals are marked in yellow cells.
      } \label{tab:forced_long_generation_length}}
    \end{minipage}
\end{spacing}
\end{minipage}

\begin{minipage}[t]{\textwidth}
    \small
    \footnotesize
    \begin{spacing}{0.7}
    \begin{minipage}[b]{\textwidth}
    \centering
  \begin{tabular}{lc|c|c|c}
    \toprule
    \multicolumn{1}{c}{\bf Method} &\multicolumn{1}{c}{Wikisection} &\multicolumn{1}{c}{Wikihow} &\multicolumn{1}{c}{ TicketTalk}&\multicolumn{1}{c}{Recipe}\\
    \midrule\midrule
    GPT2   & \revise{$47.4 \pm 5.4$} & \revise{$\underline{61.3 \pm 9.1}$} &  \revise{$19.7 \pm 2.3$} & \revise{$71.9 \pm 2.5$}   \\
    \midrule
    \vaeabbr\ (8)    & \revise{$49.7 \pm 4.3$}  &\revise{ $56.8 \pm 10.3$}  & \revise{$ 44.0 \pm 10.5$}  & \revise{$61.8 \pm 16.8$}   \\
    \vaeabbr\ (16)    & \revise{$49.3 \pm 5.3$ } & \revise{$\underline{67.7 \pm 11.2}$}  & \revise{$50.9 \pm 17.4$}  &\revise{ $54.0 \pm 25.7$}   \\
    \vaeabbr\ (32)   & \revise{ $41.3 \pm 5.6$}  & \revise{$\underline{ 62.6 \pm 5.9}$}  &\revise{ $23.1 \pm 15.5$}  & \revise{$\bf \underline{85.2 \pm 3.7}$ }  \\
    \midrule
    \revise{\static{} \vaeabbr\ (8)}    & \revise{$51.0 \pm 4.3$}  &\revise{ $53.9 \pm 11.6$}  & \revise{$45.5 \pm 11.3$}  & \revise{$70.6 \pm 13.9$}   \\
    \revise{\static{} \vaeabbr\ (16)}    & \revise{$49.8 \pm 6.1$ } & \revise{$\underline{67.9 \pm 8.4}$}  & \revise{$52.0 \pm 18.6$}  &\revise{ $ 64.8 \pm 14.7$}   \\
    \revise{\static{} \vaeabbr\ (32) }   & \revise{ $45.9 \pm 6.1$}  & \revise{$\underline{61.1 \pm 6.0}$}  &\revise{ $21.9 \pm 14.8$}  & \revise{$78.8 \pm 3.9$ }  \\
    \midrule
    \infonceabbr\ (8)   & \revise{ $50.8 \pm 3.8$ } &\revise{ $60.5 \pm 11.9$}  & \revise{ $50.0 \pm 3.4$}  & \revise{$65.6 \pm 15.0$ }   \\
    \revise{  \infonceabbr\ (16) }   & \revise{ $\underline{62.3 \pm 6.7}$} & \revise{$27.8 \pm 20.1$ } & \revise{$55.4 \pm 19.7$} & \revise{$55.7 \pm 8.5$}    \\
    \revise{\infonceabbr\ (32) }   &\revise{ $\bf \underline{64.0 \pm 3.2}$ } & \revise{$44.1 \pm 7.9$ } & \revise{$64.8 \pm 6.2$ }& \revise{$\underline{81.5 \pm 5.2}$ }   \\
    \midrule
    \revise{\static{} \infonceabbr\ (8) }   & \revise{ $51.0 \pm 5.0$ } &\revise{ $ \underline{8.5 \pm 10.4}$}  & \revise{ $49.2 \pm 13.4$}  & \revise{$75.9 \pm 7.1$ }   \\
    \revise{\static{} \infonceabbr\ (16) }   & \revise{ $21.2 \pm 4.1$} & \revise{$28.3 \pm 20.9$ } & \revise{$54.3 \pm 21.3$} & \revise{$56.6 \pm 15.7$}    \\
    \revise{\static{} \infonceabbr\ (32) }   &\revise{ $35.4 \pm 7.0$ } & \revise{$48.5 \pm 8.9$ } & \revise{$\bf \underline{67.8 \pm 10.5}$ }& \revise{$77.0 \pm 6.4$ }   \\
    \midrule
    \brownianabbr\ (8)   & \revise{$50.1 \pm 3.4$ } & \revise{$\underline{61.8 \pm 8.9}$ } & \revise{$53.8 \pm 17.3$ } & \revise{ $71.0 \pm 6.9$ }   \\
    \revise{ \brownianabbr\ (16)}   & \revise{$40.7 \pm 6.1$ } & \revise{$51.3 \pm 20.5$ } & \revise{$37.7 \pm 15.4$ } & \revise{$37.3 \pm 16.6$   } \\
    \brownianabbr\ (32)    & \revise{$43.4 \pm 7.4$}  & \revise{$41.9 \pm 15.1$ } & \revise{$ 49.2 \pm 3.7$  } & \revise{$53.2 \pm 7.6$ }   \\
    \midrule
    \revise{\static{} \brownianabbr\ (8) }   & \revise{$51.5 \pm 6.6$ } & \revise{$\underline{65.2 \pm 10.1}$ } & \revise{$54.2 \pm 19.3$ } & \revise{ $54.9 \pm 17.9$ }   \\
    \revise{\static{} \brownianabbr\ (16) }   & \revise{$42.8 \pm 5.3$ } & \revise{$\underline{66.5 \pm 15.1}$ } & \revise{$22.8 \pm 19.1$ } & \revise{$45.0 \pm 14.0$   } \\
    \static{} \brownianabbr\ (32)    & \revise{$46.3 \pm 4.2$}  & \revise{$\underline{ 66.7 \pm 10.3}$ } & \revise{$23.4 \pm 12.6$  } & \revise{$30.1 \pm 12.4$ }   \\
    \midrule
    \modelabbr\ (8) & \revise{$49.2 \pm 10.5$ }& \revise{$\underline{64.3 \pm 14.7}$ } & \revise{$48.3 \pm 23.8$ }&	\revise{$56.0 \pm 10.4$ }  \\
    \modelabbr\ (16) & \revise{$42.5 \pm 7.0$} &  \revise{$\underline{66.2 \pm 17.7}$ } & \revise{$\underline{  58.8 \pm 11.7}$} &	\revise{$66.6 \pm 10.0$ }  \\
    \revise{ \modelabbr\ (32)} &\revise{ $36.9 \pm 7.7$ } &\revise{ $\underline{69.7 \pm 12.3}$ } & \revise{$35.4 \pm 16.5 $} &	\revise{$77.0 \pm 3.5$}   \\
    \midrule
    \static{} \modelabbr\ (8) & \revise{$ 50.5 \pm 11.2$} & \revise{$\underline{69.7 \pm 12.3}$ } & \revise{$\underline{58.1 \pm 19.6} $ }&	\revise{$55.7 \pm 11.3$ }  \\
    \static{} \modelabbr\ (16) & \revise{$ 48.5 \pm 5.5$} &  \revise{$\underline{67.9 \pm 19.0}$ } & \revise{$\underline{ 67.3 \pm 8.9}$} &	\revise{$70.3 \pm 7.5$ }  \\
    \static{} \modelabbr\ (32) &\revise{ $52.1 \pm 6.4 $ } &\revise{ $\bf \underline{71.5 \pm 10.5}$ } & \revise{$ 49.1 \pm 18.0$} &	\revise{$\underline{82.4 \pm 3.3}$}   \\
    \bottomrule
  \end{tabular}
  \vspace{5pt}
  \captionof{table}{\small Ordering in forced long text generation. ROC Stories and TM-2 omitted because they are not applicable.  These results are run over 3 random GPU machines, each with 3 seeds.
  The method with the highest mean is marked in bold, and methods with overlapping confidence intervals are underlined.
  \label{tab:ordering}
  }
  
    \end{minipage}
    \hspace{1em}
    \end{spacing}
\end{minipage}

We evaluate how well the Brownian bridge contrastive learning objective helps generate globally coherent text (\textbf{RQ3}) where the EOS token is omitted.
We refer to this as the forced long text generation setup because the model must extrapolate beyond its natural end point in generation.
Appendix~\ref{sec:long_text_generation_setup} details how the latent plan is generated.
For reference, 1000 tokens is about $50\%$ longer than the average Wikisection document (the longest text domain). 
This setting is intrinsically difficult for auto-regressive models which do not have the ability to ``look ahead'' during generation  \citep{lin-etal-2021-limitations}. 

We evaluate model extrapolation on two metrics. First is ordering: how well do the models maintain the structure of the document (eg. turn-ordering in TicketTalk)? 
Second is length consistency. 
Length consistency captures common failure modes such as a model which stops modeling a conversation between two agents, and instead outputs a long monologue from one agent.

The results are summarized in Table~\ref{tab:forced_long_generation_length} for length consistency and Table~\ref{tab:ordering} for ordering. 
According to the length consistency metric in Table~\ref{tab:forced_long_generation_length}, \static{} \modelabbr\ maintains text flow the best.
Surprisingly, \static{} \modelabbr\ performs better than \modelabbr, which decodes from an actual Brownian bridge sequence.
We hypothesize that training the decoder with the contrastive encoder regularizes the model to generate better forced long texts.
\vaeabbr, \brownianabbr, \infonceabbr, and their \static{} variants do particularly worse than \static{} \modelabbr\  on length consistency on task-oriented and turn-taking domains like TM-2 and TT, suggesting that the Brownian Bridge objective can model these domains more faithfully. 
This suggests the importance of good latent dynamics for long text generation. 
\revise{Here, a common failure mode of the nominal GPT2 model} is that it produces nonsensical text where it would naturally end generation; this is particularly noticeable on TM-2 and TicketTalk. 
Figure~\ref{table:text_examples} shows an example of our model's behavior versus GPT2's behavior on generating long texts in TicketTalk. 
The example illustrates how \model\ continues the dialog whereas GPT2 utters nonsensical text.

According to the ordering metric in Table~\ref{tab:ordering}, none of the methods surpass the other consistently.
These results suggest that that even though \static{} \modelabbr\ is better at matching text lengths on the forced long text setting, decoding from latent variables to preserve semantically meaningful metrics like section ordering still remains a challenge.

\begin{table*}[t]
\centering
\small
\begin{spacing}{0.7}
\begin{tabular}{ p{13cm} } 
\toprule
\textbf{\model\ :}  [ USER ] I'm looking for movie tickets please. [ ASSISTANT ] OK, where would you like to see the movie? [ USER ] I'd like to see it at Creek's End, Oregon please. \textbf{\color{ForestGreen}[...]} [ ASSISTANT ] Is it OK to go ahead and purchase these tickets? [ USER ] Yeah, that would be great. [ ASSISTANT ] OK. I understand that you are interested in tickets for Emma at AMC Mercado 24 tonight at 7:10pm. Is that OK? [ USER ] Yes, please do that. [ ASSISTANT ] OK\\
\midrule
\textbf{GPT2:} [ USER ] Hi! Tonight, I'd like to go to the movies. [ ASSISTANT ] Okay. What theater would you like to go to? [ USER ] Center City. \textbf{\color{ForestGreen}[...]} [ ASSISTANT ] That will be all for now. Thank you for all your help. N/A [ USER ] Bye Bye. [ ASSISTANT ] N/A [ ASSISTANT ] N/A N/A N/A N/A N/A N/A N/A N/A [ USER ] N/A [ ASSISTANT ] N/A N/A N/A [ USER ] N/A [ ASSISTANT ] N/A N/A N/A [ USER ] N/A N/A N/A N/A N/A N/A N/A N/A N/A N/A [ USER ] N/A N/A N/A N/A \textbf{\color{ForestGreen}[...]} 
\\
\bottomrule
\end{tabular}
  \caption{\small Example of forced long text generation on TicketTalk with \model\ vs. fine-tuned GPT2. 
  Both models are forced to extrapolate when generating long texts.
  They start coherently, but only \model\ extrapolates coherently. 
  For space reasons, some of the text has been removed, marked with  \textbf{{\color{ForestGreen}{[...]}}}.}
\label{table:text_examples}
\end{spacing}
\end{table*}

\section{Discussion on \static{} and non-\static{} methods}

In an earlier version of this paper, we evaluated the \static{} methods. 
Surprisingly, the \static{} methods perform well despite not leveraging the dynamics at test time.
This raises an interesting question of whether there are better ways to leverage contrastive representations at decoding time.

For more details on the differences between the original manuscript and the current manuscript, please refer to Appendix~\ref{sec:version_diffs}.

\section{Conclusion}

Our work explores using representations learned via contrastive learning for generative and discriminative tasks. 
We propose one way of learning effective language representations which follow Brownian bridge dynamics.

Empirically, we demonstrate this leads to benefits such as simultaneously yielding good representations for discriminative tasks like discourse coherence and generating long texts that preserve section lengths.
Future work includes understanding why decoding from a static latent sequence is just as effective---if not more effective---as a dynamic sequence.
Additionally, decoding from a latent sequence while preserving other semantically meaningful measures beyond text length still remains a challenge, as suggested by the ordering metrics in the forced long text generation setting.
Although our work focuses on benefits of learning stochastic process latents for language generation, our approach can be extended to other domains with sequential data like videos or audio such as in \citet{shin2022clap}, or extended to handle arbitrary bridge processes without known fixed start and end points.

\newpage

\section{Reproducibility Statement}
In the supplemental, we include a zip file containing our code and processed datasets. 
We've also included in the appendix how we processed the datasets.
Our results on showing equivalence between the triplet classification setting and pairwise classification setting are also included in the appendix.

\subsection*{Acknowledgments}
REW is supported by the National Science Foundation Graduate Research Fellowship. 
The authors would give special thanks to CoColab members Mike Wu, Gabriel Poesia, Ali Malik and Alex Tamkin for their support and incredibly helpful discussions.
The authors would like to thank the rest of CoCoLab and Rishi Bommasani for reviewing the paper draft. 
The authors also thank Chris Donahue for helpful pointers on using ILM.

\bibliography{main}
\bibliographystyle{iclr2022_conference}

\appendix

\section{Showing the equivalence of the pairwise and triplet classification task} \label{sec:triplet_classification}

In this section, we focus on the classification setup where the goal is to train a classifier to detect in- vs. out-of-order groups of inputs.
Specifically, we want to show that the classification setup from \citet{liu_contrastive_2021} where the inputs come in pairs  is equivalent to the setup where the inputs come in triplets.
The triplet inputs are what we assume for \model.

\paragraph{Notation} We denote latent space as  $\mathcal{Z}$; we typically do not observe this space directly. We denote the observation space as $\mathcal{X}$; this is the space we observe directly.

We define the Brownian bridge for $t \in [0, T]$ as 
\begin{equation}
    B_t = B(t) = W(t) - \frac{t}{T} W(T),
\end{equation}
where $W(t)$ is a standard Wiener process, $\mathcal{N}(0, t)$.

We state the assumptions:
\begin{itemize}
    \item We assume latents $\{z_t\}_{t \geq 0} \in \mathbb{R}^d$ are drawn from the Brownian bridge process defined by the stochastic differential equation, 
    \begin{equation}
        dz_t = \frac{z_T-z_t}{1-\frac{t}{T}} dt + dB_t
    \end{equation}
    The intervals at which latents are sampled are every $\Delta t$ step: $z' = z_{t + \Delta t}$. 
    \item We denote the transition probabilities as 
    \begin{equation}
        p_*(z_{t}|z_0, z_T) := \mathcal{N}\left((1-\frac{t}{T})z_0 + \frac{t}{T}z_T, \frac{t(T-t)}{t}\right).
    \end{equation}
    We denote the proposal distribution over possible intermediate triplets $q(z_0', z_t', z_T')$. 
\end{itemize}

\citet{liu_contrastive_2021} characterize properties of an optimal classifier $h^*(z, z')$ which observes pairs of latents $(z, z')$ and it outputs in $[0, 1]$, a probability indicating whether the pair comes \textit{in order} (ie. $z' = z_t + \Delta t \cdot dz_t$) or \textit{not in order} (ie. a randomly sampled latent). 
They train $h^*$ using an L2 loss, $\mathcal{L}(h, \{(z, z'), y\})$. 

Lemma 1 of their work states the following : The optimum of the contrastive learning objective $\arg\min_{h} \mathbb{E}_{((z, z'), y)}[\mathcal{L}(h, \{(z, z'), y\})]$ satisfies 
\begin{equation}
    h^{*}(z, z') = \frac{p^{\Delta t}(z, z')}{q(z') + p^{\Delta t}(z, z')}. 
\end{equation}

Manipulating this equality, we observe that the transition kernel has the following relation to the classifier which takes in pairs of observations, 

\begin{align}
   p^{\Delta t}(z, z') &= \frac{q(z')h^{*}(z, z')}{1-h^{*}(z, z')} \\
   \log p^{\Delta t}(z, z') &= \log q(z') + \log h^{*}(z, z') - \log (1-h^{*}(z, z')). 
\end{align}

Our setting however assumes that the algorithm receives triplets of data points, $z_{t_1}, z_{t_2}, z_{t_3}$. 
We want to show below that minimizing $\mathcal{L}$ with triplet contrasts in the classification setting still approximates the transition kernel. 
In particular, we're interested in transitions of a Brownian bridge pinned at $z_0, z_T$: $p^{\Delta t}(z_0, z_t, z_T) = Pr(z_{t} | z_0, z_T)$. 

Let's say we have two positive triplet samples, $(z^i_{t_1}, z^i_{t_2}, z^i_{t_3})$ and $(z^j_{t_1}, z^j_{t_2}, z^j_{t_3})$, where $t_1 < t_2 < t_3$. 
Following \citet{liu_contrastive_2021}, minimizing $\mathcal{L}$ yields the following on each triplet: 

\begin{align}
    \log p^{\Delta t}(z^i_{t_1}, z^i_{t_2}, z^i_{t_3}) &= \log q(z') + \log h^{*}(z^i_{t_1}, z^i_{t_2}, z^i_{t_3}) - \log (1-h^{*}(z^i_{t_1}, z^i_{t_2}, z^i_{t_3})) \label{eq:triplet_probs_i} \\
    \log p^{\Delta t}(z^j_{t_1}, z^j_{t_2}, z^j_{t_3}) &= \log q(z') + \log h^{*}(z^j_{t_1}, z^j_{t_2}, z^j_{t_3}) - \log (1-h^{*}(z^j_{t_1}, z^j_{t_2}, z^j_{t_3})) \label{eq:triplet_probs_j}. 
\end{align}

Taking the difference in log probabilities between Equations~\ref{eq:triplet_probs_i}-\ref{eq:triplet_probs_j} results in

\begin{equation}
    \begin{split}
        \log p^{\Delta t}(z^i_{t_1}, z^i_{t_2}, z^i_{t_3}) - \log p^{\Delta t}(z^j_{t_1}, z^j_{t_2}, z^j_{t_3}) &= \left[\log h^{*}(z^i_{t_1}, z^i_{t_2}, z^i_{t_3}) - \log (1-h^{*}(z^i_{t_1}, z^i_{t_2}, z^i_{t_3}))\right] \\ &\quad - \left[h^{*}(z^j_{t_1}, z^j_{t_2}, z^j_{t_3}) - \log (1-h^{*}(z^j_{t_1}, z^j_{t_2}, z^j_{t_3}))\right].
    \end{split}
\end{equation}

Similar to the pair-wise classification setting, we've shown that minimizing $\mathcal{L}$ in the triplet classification setting results in approximating the transition kernel of the Brownian bridge process.

\section{\revise{Machines used for training}  \label{sec:machine_training}}

\revise{
We found that our decoding results are sensitive to the GPU machine type. 
For example, in the length mismatch results in Table~\ref{tab:length_matching}, \modelabbr\ (8) had a deviation of $7.7 \pm 1.2$ in one set of three seeds and a deviation of $17.3\pm5.2$ in another set of three seeds.
In light of this, we randomized over the choice of machines.
These machines include: RTX A6000, RTX A5000, GeFORCE RTX 3090, TITAN RTX,  TITAN Xp, TITAN V, and GeForce GTX TITAN X machines. 
}

\section{Training details  \label{sec:training}}

Here we describe the training procedure for the encoder and the fine-tuning procedure for decoding. 

\paragraph{Brownian bridge encoder}
The encoder architecture is a frozen, pretrained GPT2 model from Huggingface \citep{radford2019language, wolf-etal-2020-transformers} and a trainable MLP network.
We extract the GPT2's last layer hidden state that corresponds to the end-of-sentence (EOS) token and train the 4-layer MLP on top of the hidden state. 
The MLP network has intermediate ReLU activations and is trained with stochastic gradient descent with a learning rate of 1e-4 and with momentum 0.9. We train the encoder for 100 epochs on each of the datasets.

The text fed into GPT2 are fed in on a sentence level. This means that the input $x_t$ refers to the $t$'th sentence of a document. 
The sentences are separated from each other in the main text as `` . '' which is added to the tokenizer as a separate token for indexing convenience.

\paragraph{Fine-tuning GPT2 with latent embeddings}
After training the encoder, we run it on the training dataset to collect an accompanying latent trajectory for each text.
The encoder is run on the dataset at a sentence level: we separate the text by sentences and pass the sentences through the encoder. 

The sentence latent embeddings are aligned with the tokens of that sentence and offset by one token before the start of that sentence token. 
Let's illustrate by an example. We denote [SOS] as the start of the document token, [s1] as sentence 1 tokens and [s2] as sentence 2 tokens. [ . ] is the period token which we've added into the tokenizer. $z_i$ denote the latent variable corresponding to the $i$'th sentence. 

Let's say that the sequence fed into GPT2 is ``[SOS] [s1] [s1] [s1] [ . ] [s2] [s2] [s2]''. Then the corresponding latent trajectory is ``$z_1,z_1,z_1,z_1,z_2,z_2,z_2,z_2$''.
The latent variables are added onto the positional embeddings. We then fine-tune GPT2 as normal.

\section{Generation with embeddings\label{sec:app_generation}}

\subsection{Normal text generation}
We first sample a start and end latent, $z_{0} \sim p(z_0), z_{T} \sim p(z_T)$ where $p(z_0), p(z_T)$ are calculated as the density estimates over the training dataset. 
We pin our trajectory to the start and end latent, and run the Brownian bridge using a first-order approximation.
For normal long text generation, we set $T$ to be the average number of sentences in each of the dataset. 
For forced long text generation, we set $T$ to be proportional to the number of sentences needed in order to generate 1000 tokens.
By the end, we have a trajectory $z_0, z_1, ..., z_T$. 

Generation starts with feeding the SOS token and the first latent $z_0$. 
Once GPT2 emits a [ . ] token and terminates sentence $t$, we transition to the next latent $z_{t+1}$. This process continues until GPT2 is finished with generation. 
If GPT2 generates more sentences than there are latents in the trajectory, the last latent $z_T$ is used until the end of generation.

\subsection{Forced long text generation \label{sec:long_text_generation_setup}}

Let $S_{\text{avg}(\mathcal{D})}$ denote the average number of sentences in a document and $T_{\text{avg}(\mathcal{D})}$ denote the average number of tokens in a document.
Rather than planning a trajectory of length $S_{\text{avg}(\mathcal{D})}$ (the average number of sentences in a document) which is what is done in Section 4.3 for normal text generation, we scale the trajectory length to $c \cdot S_{\text{avg}(\mathcal{D})}$. $c$ is determined by how many more tokens we need in order to fill up to GPT-2 maximum context length of 1024: $c = \frac{1024 - T_{\text{avg}(\mathcal{D})}}{T_{\text{avg}(\mathcal{D})}}$. 

\section{Ablations} \label{sec:app_ablations}

The following methods ablate the encoder model in \model; in other words, the ablations dissect the assumptions we make in the first step of \model\ described in Section~\ref{sec:encoding}.
Recall that \model\ \textbf{(A) explicitly} models latent structure with \textbf{(B) Brownian bridge dynamics} using a \textbf{(C) contrastive} loss.
\textbf{(A)} replaces explicit dynamics with \infonce\ (\infonceabbr) where future latents are directly predicted with an autoregressive model \citep{oord2019representation}. 
\textbf{(B)} replaces Brownian bridge dynamics with \brownian\ (\brownianabbr) which doesn't rely on pinning trajectories. Latents follow the transition density $ z_t | z_s  \sim \mathcal{N}(z_s, t-s)$. 
\textbf{(C)} replaces the contrastive loss with the \vae\ (\vaeabbr).
Below we detail how these ablations are implemented.

\subsection{\infonce\ (\infonceabbr)}
The \infonce\ ablation is where we compare our \textit{explicit} dynamics objective to \citet{oord2019representation} which suggests an \textit{implicit} latent dynamics objective. 
This ablation thus changes the learned latent dynamics. 
In \citet{oord2019representation}, they train two models. 
One is a non-linear encoder $g_{\text{enc}}(x_t) = z_t$ which takes observation $x_t$ (eg. a sentence) and maps it to a latent representation $z_t$.
Two is an autoregressive context model $g_{\text{ar}}(z_{\leq t})=c_t$ which summarizes a sequence of latent variables $z_{\leq t}$ into a context latent representation $c_t$.
Rather than directly predicting a future observation $x_{t+k}$ ($k$ steps from the current timestep $t$), they model the density ratio that preserves the mutual information between $x_{t+k}$ and the context variable $c_t$:

\begin{equation}
    f_{k}(x_{t+k}) \propto \frac{p(x_{t+k}|c_{t})}{p(x_{t+k})}
\end{equation}

They model this with a log-bilinear model $f_{k}(x_{t+k}) = \exp(z_{t+k}^T W_k c_t)$ which applies a linear transformation $W_k$ for modelling latents k-steps away from timestep $t$.
This way, they avoid directly learning the generative model $p(x_{t+k}|c_t)$.

They train both models jointly via a contrastive InfoNCE loss. Given a set $X=\{x_1, \dots, x_N\}$ of $N$ random samples (eg. sentences from different documents) containing one positive sample from $p(x_{t+k}|c_t)$ and $N-1$ negative samples from a proposal distribution $p(x_{t+k})$, they optimize:

\begin{equation}
    \mathcal{L} = - \mathbb{E}_{X}\left[\log \frac{f_k(x_{t+k}, c_t)}{\sum_{x_j \in X} f_{k}(x_j, c_t)} \right].
\end{equation}

The encoder $g_{\text{enc}}$ for \infonce\ has the same architecture as \model. The context encoder $g_{\text{ar}}$ using a 2400-hidden-unit GRU, as done in \citet{oord2019representation}.
We then train both encoders using the InfoNCE loss, as done in prior work.
Since $g_{\text{enc}}$ and $g_{\text{ar}}$ are trained to align latents up to a linear rotation, we use $g_{\text{enc}}$ for extracting the sentence embeddings. 

\subsection{\brownian\ (\brownianabbr)}
The \brownian\ ablation is where we remove goal-conditioning  in the dynamics. 
The main change is in distance function in Equation~\ref{eq:distance_function}. 
\brownianabbr\ instead optimizes the following equation: 

\begin{equation}
    \text{d}(x_{t}, x_{t'}; f_{\theta}) = -\frac{1}{2\sigma^2}\| \underbrace{f_{\theta}(x_{t'})}_{z_{t'}} - \underbrace{f_{\theta}(x_t)}_{z_t} \|^2_2    
\end{equation}
where $\sigma^2$ is the variance in of the Wiener process, $\sigma^2 = t' - t$.

\subsection{\vae\ (\vaeabbr)}
The \vae\ ablation is where we compare our contrastive objective with a variational one. 
This ablation changes the encoder objective from Section~\ref{sec:encoding} from Equation~\ref{eq:loss_objective} to the ELBO objective.
Below we derive the ELBO objective.
Similar to the contrastive objective, we're deriving the ELBO over the triplet dynamics in the Brownian bridge.

\begin{align*}
    \log p(\mathbf{x}) &\geq \mathbb{E}_{q_{\phi}(\mathbf{z}|\mathbf{x})}[\log \frac{p(\mathbf{x}, \mathbf{z})}{q(\mathbf{z}|\mathbf{x})}] \\
    &= \mathbb{E}_{q_{\phi}(\mathbf{z}|\mathbf{x})}[\log \frac{p(x_0, x_t, x_T, z_0, z_t, z_T)}{q(z_0, z_t, z_T|x_0, x_t, x_T)}] \\
    &= \mathbb{E}_{q_{\phi}(\mathbf{z}|\mathbf{x})}[\log \frac{p(x_0|z_0)p(x_t|z_t)p(x_T|z_T)p(z_t|z_0, z_T)p(z_0)p(z_T)}{q(z_t|z_0, z_T, x_t)q(z_0|x_0)q(z_T|x_T)}] \\
    &= \mathbb{E}_{q_{\phi}(\mathbf{z}|\mathbf{x})}[p(x_0|z_0)p(x_t|z_t)p(x_T|z_T)] + \mathbb{E}_{q_{\phi}(\mathbf{z}|\mathbf{x})}[\log \frac{p(z_t|z_0, z_T)}{q(z_t|z_0, z_T, x_t)}] \\ & \quad \quad  + \mathbb{E}_{q_{\phi}(\mathbf{z}|\mathbf{x})}[\log \frac{p(z_0)}{q(z_0|x_0)}]+ \mathbb{E}_{q_{\phi}(\mathbf{z}|\mathbf{x})}[\log \frac{p(z_T)}{q(z_T|x_T)}] \\
    &= \mathbb{E}_{q_{\phi}(\mathbf{z}|\mathbf{x})}[\log p(x_0|z_0)p(x_t|z_t)p(x_T|z_T)] \\ & \quad \quad  - D_{\text{KL}}(q(z_t|z_0, z_T, x_t) \| p(z_t|z_0, z_T)) - D_{\text{KL}}(q(z_0|x_0) \| p(z_0)) - D_{\text{KL}}(q(z_T|x_T) \| p(z_T))
\end{align*}

We assume the priors over $z_0$ is 0-centered and $z_T$ is 1-centered, which is similar to our Brownian Bridge setup. 
The encoder $q_{\phi}(z|x)$ is parameterized with the same architecture as our encoder. 
The decoder $p(x|z)$ is a fine-tuned GPT2 model. 

\subsection{\static\ decoding}

\revise{\static{} impacts how the latent plans are generated for the methods at inference time. Instead of generating the latent plan $[z_0, z_1, \dots, z_T]$ by running the Brownian bridge dynamics (applicable to \vaeabbr, \infonceabbr, and \modelabbr) or Brownian motion (applicable to \brownianabbr), the method generates the plan $[z_0, z_0, \dots, z_0]$ of the same $T$ length and decodes from this plan.}

\section{Dataset information \label{sec:dataset_meta}}

For each dataset, text examples were filtered out if they did not fit within GPT2's context length of 1024 tokens. 
We also added the token `` . '' for each setting to mark the end of a sentence. This was done for indexing purposes, eg. when aligning the latent embeddings.

\paragraph{Wikisection} \citep{arnold2019sector} includes Wikipedia articles on cities split by sections. 
We adapt this dataset such that each article contains four ordered sections (abstract, history, geography, demographics) marked with section id tokens: Each article is represented as, ``[ABSTRACT] text [HISTORY] text [GEOGRAPHY] text [DEMOGRAPHICS] text''. These section id tokens are added to the tokenizer.

The training dataset contains 1420 articles. The section lengths have the following breakdown measured in the number of BPE tokens (GPT2 tokenizer):
\begin{itemize}
    \item Abstract: $75.8 \pm 1.4$
    \item History: $191.5\pm3.7$
    \item Geography: $83.9\pm 1.5$
    \item Demographics: $342.6\pm 4.6$
\end{itemize}

The test dataset contains 431 articles. The section lengths have a similar breakdown: 
\begin{itemize}
    \item Abstract: $73.5 \pm 2.6$
    \item History: $180.2 \pm 6.2$
    \item Geography: $85.2 \pm 2.7$
    \item Demographics: $332.5 \pm 8.6$
\end{itemize}

The ordering metric used in Table~\ref{tab:ordering} is $1$ if all four section ids occur exactly once and come in the order as they are listed above. 

The length mismatch in $\%$ used in Table~\ref{tab:length_matching} is calculated with respect to the training set lengths. 

\paragraph{Wikihow} \citep{koupaee2018wikihow} contains how-to articles organized by a title, method, and steps. Each article includes multiple methods for completing a multi-step procedural task such as \textit{``How to Register to Vote''}. We scraped all the available English articles covering a wide range of topics following \citet{koupaee2018wikihow}. 
We mark each with its own section id tokens: Each article is represented as ``[TITLE] text [METHOD] text [STEP] 1 text [STEP] 2 text ...'' 

The training dataset contains 1566 articles. The section lengths are, 
\begin{itemize}
    \item Title: $10.4\pm 2.2$
    \item Method: $8.7 \pm 2.2$
    \item Steps (total step length): $480.2\pm 231.5$
\end{itemize}

The test dataset contains 243 articles. The section lengths are, 
\begin{itemize}
    \item Title: $10.7\pm 2.2$
    \item Method: $8.6 \pm 2.1$
    \item Steps (total step length): $480.1\pm 224.0$
\end{itemize}

The ordering metric used in Table~\ref{tab:ordering} is $1$ if all the section ids appear in order, the TITLE and METHOD section ids are not repeated, and the step numbers come in order. It's $0$ otherwise.

The deviation in section length measured in Table~\ref{tab:forced_long_generation_length} is calculated with respect to the training set lengths. 
We check for whether the models are able to maintain the section lengths when it has to extrapolate. 
The most common failure mode is that the model generates incoherent text and allocates this text to the last section of what it's generated thus far, resulting in the deviation in section lengths.

\paragraph{Recipe NLG} \citep{bien-etal-2020-recipenlg} contains recipes, each with a title, ingredients and set of instructions. A recipe is constructed as ``[TITLE] text [INGREDIENTS] text [DIRECTIONS] text''.

The training dataset contains 4000 recipes. The section lengths are, 
\begin{itemize}
    \item Title: $9.7\pm 3.4$
    \item Ingredients: $23.8 \pm 4.5$
    \item Directions (total step length): $62.0\pm 14.5$
\end{itemize}

The test dataset contains 1000 recipes. The section lengths are, 
\begin{itemize}
    \item Title: $9.4\pm 3.0$
    \item Ingredients: $24.1 \pm 4.5$
    \item Directions (total step length): $63.3\pm 13.7$
\end{itemize}

The ordering metric used in Table~\ref{tab:ordering} is $1$ if all the section ids appear exactly once and in order. It's $0$ otherwise.

\paragraph{Taskmaster-2} (TM-2) \citep{byrne2019taskmaster} which contains conversations on finding restaurants between an assistant and a user. The assistant's turn is marked with an ``[ASSISTANT]'' tag, and the user's turn is marked with a ``[USER]'' tag.

The training dataset contains 2000 conversations. The section lengths are, 
\begin{itemize}
    \item User: $11.8 \pm 5.6$
    \item Assistant: $18.0 \pm 3.7$
\end{itemize}

The test dataset contains 1276 conversations. The section lengths are, 
\begin{itemize}
    \item User: $11.7 \pm 5.6$
    \item Assistant: $19.5 \pm 3.0$
\end{itemize}

The ordering metric used in Table~\ref{tab:ordering} does not apply here because the user and assistant don't take turns in the dialog. 

The deviation in section length measured in Table~\ref{tab:forced_long_generation_length} is calculated with respect to the training set lengths. 
We check for whether the models are able to maintain the utterance lengths between the user and assistant when it has to extrapolate. 
The most common failure mode is that the model generates incoherent text and allocates this text to the last section of what it's generated thus far, resulting in the deviation in section lengths.

\paragraph{TicketTalk} \citep{byrne-etal-2021-tickettalk} which contains conversations on booking movie tickets between an assistant and a user. The assistant's and user's turned are similarly marked as in TM-2.

The training dataset contains 2000 conversations. The section lengths are, 
\begin{itemize}
    \item User: $11.8 \pm 5.6$
    \item Assistant: $18.0 \pm 3.7$
\end{itemize}

The test dataset contains 1276 conversations. The section lengths are, 
\begin{itemize}
    \item User: $11.7 \pm 5.6$
    \item Assistant: $19.5 \pm 3.0$
\end{itemize}

The deviation in section length measured in Table~\ref{tab:forced_long_generation_length} is calculated similarly to TM-2.

The ordering metric used in Table~\ref{tab:ordering} applies because the user and assistant take turns in the dialog. The ordering metric is $1$ if the user and assistant take turns in the conversation. It's $0$ otherwise.

\paragraph{ROC Stories} \citep{mostafazadeh-etal-2016-corpus} is a short stories dataset. Each story contains 5 sentences. No additional tokens are added in this dataset. The training dataset contains 2000 stories, and the test dataset contains 1000 stories.
\revise{This dataset was used in the original text infilling experiments we had. Those previous results are in Table~\ref{tab:bertscore}, Table~\ref{tab:bleu_scores} and Table~\ref{tab:human_eval_text_infilling}.}

\section{Perplexity after fine-tuning \label{sec:ppl_fine-tuning}}

\begin{table}[]
    \centering
  \begin{tabular}{lc|c|c|c|c}
    \toprule
    \multicolumn{1}{c}{\bf Method} &\multicolumn{1}{c}{Wikisection} &\multicolumn{1}{c}{Wikihow}&\multicolumn{1}{c}{TM-2} &\multicolumn{1}{c}{ TicketTalk}&\multicolumn{1}{c}{Recipe}\\
    \midrule\midrule
    GPT2   & $ 5.9 $ & $  15.3 $ & $  4.5 $ & $   4.4 $ & $ 7.5  $  \\
    sec-dense   & $ 5.9 $ & $-$  & $-$  & $-$  & $-$    \\
    sec-sparse   & $ 5.9 $ & $-$   & $-$  & $-$  & $-$    \\
    \midrule
    \vaeabbr\ (8)    & $  5.5 $ & $ 15.3  $ & $  4.5 $ & $ 4.1  $ & $ 7.3  $  \\
    \vaeabbr\ (16)    & $  5.5 $ & $ 15.3  $ & $ 4.5  $ & $  4.1 $ & $  7.3 $  \\
    \vaeabbr\ (32)    & $  5.5 $ & $  15.3 $ & $  4.5 $ & $ 4.1  $ & $ 7.3  $  \\
    \midrule
    \infonceabbr\ (8)   & $  5.5 $ & $  14.9 $ & $  4.5 $ & $ 4.0  $ & $ 7.3  $  \\
    \infonceabbr\ (16)   & $  5.4 $ & $  14.9 $ & $  4.3 $ & $ 3.9  $ & $ 7.0  $  \\
    \infonceabbr\ (32)    & $  5.3 $ & $  14.9 $ & $ 4.2  $ & $ 3.9  $ & $  6.7 $  \\
    \midrule
    \brownianabbr\ (8)    & $ 5.5 $ & $  15.2 $ & $ 4.5  $ & $ 4.1  $ & $ 7.3  $  \\
    \brownianabbr\ (16)    & $ 5.5 $ & $  15.3 $ & $ 4.5  $ & $ 4.1  $ & $    7.3 $  \\
    \brownianabbr\ (32)    & $ 5.5 $ & $ 15.2  $ & $ 4.5  $ & $ 4.1  $ & $   7.3  $  \\
    \midrule
    \modelabbr\ (8) & $  5.5 $ & $  15.2 $ & $ 4.3  $ & $  4.0 $ & $  6.9 $  \\
    \modelabbr\ (16) & $  5.5 $ & $  15.3 $ & $ 4.3  $ & $ 4.0  $ & $ 6.9  $  \\
    \modelabbr\ (32) & $  5.5 $ & $ 15.2  $ & $  4.3 $ & $  4.0 $ & $ 7.0  $  \\
    \bottomrule
  \end{tabular}
  \vspace{5pt}
  \caption{\small Perplexity after fine-tuning.\label{tab:app_ppl}}
\end{table}

Table~\ref{tab:app_ppl} reports the final perplexity scores after fine-tuning GPT2 on the different domains with the methods. 
We fine-tune for 10 epochs and checkpoint the models every 1000 steps; we keep the model checkpoint that scores the lowest PPL on a held-out validation set.

\section{Differences in the corrected manuscript \label{sec:version_diffs}}

\revise{
This section discusses updates and differences from an earlier version of this paper due to correcting bugs during a post-publication code review and re-running experiments.
Sources of randomness, such as the behavior of the torch dataloader, GPU nondeterminism, and Weights and Biases logging have caused some differences in the experimental outcomes even in cases where there were no bugs. To make the differences clear, we highlight both original and current results, as well as a short explanation of any differences.
The numbers reported in the main text correspond to the current version of the codebase available at \url{https://github.com/rosewang2008/language_modeling_via_stochastic_processes} with version hash: \textbf{95ded1e82f904e1c2dbce5701f54ab880a1a1e62}.
}

\subsection{Discourse coherence}

\revise{For reference, Table~\ref{tab:full_discourse_results} shows both the original and new results. The original results are marked with \original.}

\paragraph{Differences} 
\revise{In the previous version, ALBERT, SimCSE and TC were competitive with each other. In the updated version,  BERT is competitive with these methods as well. TC is the most competitive out of all the GPT2-based models.}

\paragraph{Cause} 
\revise{This discrepancy is due to a wandb step logging issue where wandb previously only reported on early training epochs. Therefore, the previous discourse coherence results generally under-report the actual performance–this particularly impacts Wikisection. In general, all methods improve and do better than random accuracy after resolving this issue.}

\begin{spacing}{0.5}
\begin{table*}[t]
  \centering
  \small
  \begin{tabular}{l|cc|cc|cc}
    \toprule
    \multicolumn{1}{c}{\bf }  &\multicolumn{2}{c}{\bf Wikisection} &\multicolumn{2}{c}{\bf TM-2} &\multicolumn{2}{c}{\bf TicketTalk}
\\ 
    \multicolumn{1}{c}{\bf Method}  &\multicolumn{1}{c}{$k=5$} &\multicolumn{1}{c}{$k=10$} &\multicolumn{1}{c}{$k=5$}&\multicolumn{1}{c}{$k=10$} &\multicolumn{1}{c}{$k=5$}&\multicolumn{1}{c}{$k=10$} \\
    \midrule\midrule 
    \original{} BERT & $50.9 \pm 4.9$ &  $ 47.8 \pm 9.0 $ & $68.8 \pm 3.5$ & $80.7 \pm 3.8$ & $68.4 \pm 5.1$ & $ 80.4 \pm 6.3$ \\
    \original{} ALBERT & $49.9 \pm 12.1$ &  $ 49.6 \pm 18.0$ &$ 81.6 \pm 4.0$ & $86.1 \pm 7.3$ &$ 78.4 \pm 6.7$ &$89.4 \pm 3.1$ \\
    \original{} S-BERT  & $ 50.8 \pm 6.0$ & $48.0 \pm 9.1$  & $73.4 \pm 3.5$ & $ 83.3 \pm 4.3$ & $ 72.1 \pm 5.3$ & $84.2 \pm 5.2$   \\
    \original{} Sim-CSE  & $49.1 \pm 6.4$ & $48.1 \pm 8.5$ 	& $75.4 \pm 3.8$ & $ 86.2 \pm 3.9$ & $75.1 \pm 5.9$ & $ 85.2 \pm 3.1$   \\
    BERT & $ 72.7 \pm 0.1$ &  $80.4 \pm 0.2 $ & $ 75.6 \pm 0.0 $ & $ 86.7 \pm 0.0 $ & $73.1 \pm 0.2$ & $ 86.4 \pm 0.0 $ \\
    ALBERT & $ 67.7 \pm 15.1$ &  $ 82.3 \pm 0.3$ & $ 83.9 \pm 1.7 $ & $ 92.2 \pm 0.9 $ &  $ 74.5 \pm 9.8$ &  $ 93.3 \pm 0.6$ \\
    S-BERT  & $ 71.8 \pm 0.1$ & $ 79.0 \pm 0.1$  & $ 75.6 \pm 0.0 $ & $ 86.7 \pm 0.0 $ & $ 76.2 \pm 0.1 $ & $ 88.8 \pm 0.0 $   \\
    Sim-CSE  & $72.6 \pm 0.1$ & $ 79.5 \pm 0.1$ 	& $ 77.8 \pm 0.0 $ & $ 88.3 \pm 0.0 $ & $ 78.0 \pm 0.0 $ & $ 90.1 \pm 0.1 $   \\
    \midrule 
    \original{} GPT2  & $50.3 \pm 5.8$  & $50.2 \pm 6.3$ & $55.7 \pm 5.3$ & $63.6 \pm 7.3$ & $54.7 \pm 6.1$  & $65.0 \pm 8.1$  \\  
    GPT2  & $ 58.4 \pm 3.0$  & $67.1 \pm 2.3$ & $ 66.8 \pm 0.9 $ & $ 75.6 \pm 0.0 $ & $55.2 \pm 1.9$  & $78.0 \pm 2.7$  \\   
    \midrule
    \original{} \vaeabbr\ (8) & $49.5 \pm 5.5$  & $50.5 \pm 5.1$ & $50.5 \pm 4.4$  & $51.5 \pm 6.0$ &  $49.9 \pm 1.0$ & $51.2 \pm 1.0$ \\
    \original{} \vaeabbr\ (16) & $50.1 \pm 5.8$ & $51.3 \pm 4.7$ & $48.8 \pm 4.8$ & $50.8 \pm 4.9$ &  $50.1 \pm 1.0$ &  $49.5 \pm 1.0 $   \\
    \original{} \vaeabbr\ (32) & $50.5 \pm 5.1$ & $50.0 \pm  6.0$ &  $48.0 \pm 5.1$  & $47.3 \pm 5.9$ & $50.0 \pm 1.0$ & $49.3\pm 1.0$   \\
    \vaeabbr\ (8) & $50.0 \pm 3.8$  & $50.3 \pm 0.8$ & $ 49.4 \pm 1.1 $  & $ 49.9 \pm 1.7 $ &  $ 49.0 \pm 2.6$ & $ 49.0 \pm 5.7$ \\
    \vaeabbr\ (16) & $ 51.9 \pm 1.3$ & $ 55.3 \pm 3.2$ & $ 52.9 \pm 4.2 $ & $ 56.8 \pm 6.5$ &  $ 51.9 \pm 4.2 $ &  $56.3 \pm 5.5 $   \\
    \vaeabbr\ (32) & $ 52.3 \pm 3.0 $ & $ 50.7 \pm 2.3$ &  $ 55.7 \pm 1.3  $  & $ 64.1 \pm 2.3 $ & $59.3 \pm 1.5 $ & $ 69.5 \pm 1.0 $   \\
    \midrule
    \original{} \infonceabbr\ (8) & $ 49.8 \pm 5.9 $  & $ 50.1 \pm 5.0 $ & $60.3 \pm 5.2 $  & $65.2 \pm 6.8$ &  $59.2\pm 1.9$ &  $66.5 \pm 1.1$   \\
    \original{} \original{} \infonceabbr\ (16) & $ 53.3 \pm 5.4$ & $55.8 \pm 6.2 $  & $60.5 \pm 5.0$ & $67.7 \pm 6.8$ &  $60.3 \pm 1.0$ &  $68.4 \pm 6.4 $  \\
    \original{} \infonceabbr\ (32) & $50.0 \pm 5.0 $ & $ 50.1 \pm 5.0 $ & $60.4 \pm 5.3$ & $67.6 \pm 7.1$ &  $61.0 \pm 1.0$ &  $67.9 \pm 6.5 $  \\
    \infonceabbr\ (8) & $ 58.2 \pm 0.4$  & $ 64.7 \pm 1.6  $ & $ 62.4 \pm 0.6 $  & $ 71.0 \pm 0.5$ &  $ 54.3 \pm 1.0$ &  $ 64.7 \pm 0.5$   \\
    \infonceabbr\ (16) & $ 68.5 \pm 0.1 $ & $ 76.9 \pm 0.1 $  & $ 62.7 \pm 0.1 $ & $ 71.1 \pm 0.1 $ &  $ 53.0 \pm 0.2$ &  $ 66.2 \pm 0.1 $  \\
    \infonceabbr\ (32) & $70.1 \pm 0.1$ & $ 77.1 \pm 0.1 $ & $ 62.6 \pm 0.1 $ & $71.1 \pm 0.0$ &  $ 55.4 \pm 0.2 $ &  $ 69.4 \pm 0.1 $  \\
    \midrule
    \original{} \brownianabbr\ (8) & $ $ $49.8 \pm 5.4$ & $50.0 \pm 5.4$ & $ 49.8 \pm 5.4$ & $49.9 \pm 5.2$ &  $49.7 \pm 5.0$ &   $50.6 \pm 5.8$ \\
    \original{} \brownianabbr\ (16) & $50.3 \pm 5.5$ & $50.5 \pm 5.2$ &  $49.9 \pm 4.3$ & $51.1 \pm 6.0$ &  $50.3 \pm 4.6$ &  $50.8 \pm 5.5$  \\
    \original{} \brownianabbr\ (32) & $49.3 \pm 5.6$ & $48.8 \pm 5.8$ & $49.5 \pm 4.7$ & $49.6 \pm 5.2$ &  $49.5 \pm 5.6$ &  $49.1 \pm 6.1 $  \\
    \brownianabbr\ (8) &  $ 49.4 \pm 3.2$ & $ 56.7 \pm 6.8$ & $ 52.1 \pm 3.1 $ & $ 55.0 \pm 5.1 $ &  $ 47.8 \pm 2.8 $ &   $ 51.2 \pm 3.2 $ \\
    \brownianabbr\ (16) & $ 52.8 \pm 0.8$ & $ 56.3 \pm 0.9$ &  $ 51.5 \pm 1.1$ & $52.8 \pm 2.3 $ &  $ 55.2 \pm 5.5 $ &  $62.4 \pm 5.6$  \\
    \brownianabbr\ (32) & $ 51.8 \pm 1.9 $ & $ 55.9 \pm 4.3 $ & $ 50.4 \pm 1.0$ & $ 51.3 \pm 2.2$ &  $ 55.0 \pm 2.6 $ &  $ 69.6 \pm 2.6 $  \\
    \midrule
    \original{} \modelabbr\ (8) & $49.23 \pm 5.72$ & $48.3 \pm 6.8$ &  $  77.6 \pm 7.8$ & $ 87.7 \pm 6.9$ & $ 71.6 \pm 2.9$ & $82.9 \pm 4.1$  \\
    \original{} \modelabbr\ (16) &  ${57.25 \pm 5.30}$  &  ${65.8 \pm 5.4}$  &  ${78.2 \pm 8.1}$ &  ${88.0 \pm 7.1}$ & $ 71.3 \pm 3.3$ & $82.9 \pm 4.1$ \\
    \original{} \modelabbr\ (32) & $50.1 \pm 4.8$  & $49.8 \pm 5.8$ & $ 77.9 \pm 7.9 $ & $ 87.9 \pm 7.4$  & $72.0 \pm 3.9$ & $ 84.4 \pm 3.9$  \\
    \modelabbr\ (8) & $ 69.3 \pm 0.1 $ & $ 77.6 \pm 0.0$ &  $ 78.4 \pm 0.0 $ & $ 88.7 \pm 0.0$ & $74.7 \pm 0.1 $ & $ 87.2 \pm 0.0 $  \\
    \modelabbr\ (16) & $71.7 \pm 0.0 $  & $ 76.8 \pm 0.0$  &  $ 78.7 \pm 0.0 $ & $ 89.0 \pm 0.0 $ & $ 74.7 \pm 0.0 $ & $ 87.2 \pm 0.1 $ \\
    \modelabbr\ (32) & $ 71.6 \pm 0.0$  & $ 77.3 \pm 0.0$ & $  79.2 \pm 0.0 $ & $ 89.1 \pm 0.0$  & $ 74.0 \pm 0.0 $ & $ 87.3 \pm 0.0 $  \\
    \bottomrule
  \end{tabular}
  \vspace{5pt}
  \caption{\small  Discourse coherence accuracy measured by the test accuracy of the trained linear classifier, reporting $\mu \pm $ standard error over 3 runs. Random accuracy is 50\%. When applicable, the methods are run with varying latent dimensions marked in parentheses (dim).
  \revise{The label \original{} are the experiments from the original camera-ready version.}
  }
  \label{tab:full_discourse_results}
\end{table*}
\end{spacing}

\subsection{Length mismatch}

\revise{For reference, Table~\ref{tab:full_length_matching} shows both the original and new results. The original results are marked with \original.}

\paragraph{Differences} 
In the previous version, TC performed the best. In the updated version, \static{} \vaeabbr\ performs the best, followed by \static{} \modelabbr.

\paragraph{Causes} 
The variability is caused by the dataset loader being seeded separately from usual seeding sources, GPU nondeterminism combined with sensitivity of the evaluation to this nondeterminism.
The experiments reported in the paper are all seeded.
If the experiments are run on the same machine, they replicate exactly. 
We observe different experimental results when run on \textit{different} machines. 
For example, in our experiments, \modelabbr\ (8) on the history section mismatches the original section by about $\sim 45$ tokens on one machine, and $\sim 21$ tokens on another machine.

\subsection{Section length deviation}

\revise{
For reference, Table~\ref{tab:full_forced_long_generation_length} shows both the original and new results. The original results are marked with \original. 
}

\paragraph{Differences} 
\revise{
In the previous version, TC performed the best in not deviating from the expected section lengths (Table 4). In the updated version, this is still the case with \static{} \modelabbr. 
Surprisingly, we found that \static{} \brownianabbr\ and \vaeabbr\ performs competitively in one of the domains as well.
}

\paragraph{Causes} 
Besides the variability causes mentioned in Table 3 section, there was an improper implementation of forced generation related to how we specify the min and max generation length. 
This affected all methods equally by cutting decoding short.

\begin{spacing}{0.5}
\begin{table*}
\begin{tabular}{l|c}
    \toprule
    \multicolumn{1}{c}{\bf Method}  &\multicolumn{1}{c}{\bf MM \% ($\downarrow$)}
\\ 
    \midrule\midrule
    \original{} GPT2  & $17.5 \pm 0.1$ \\ 
    \original{} SD  & $10.0 \pm 0.1 $  \\ 
    \original{} SS  & $10.6 \pm 0.1$  \\ 
    GPT2  & \revise{$9.9 \pm 3.6$} \\ 
    \revise{SD}  & \revise{$28.3 \pm 7.9$}  \\ 
    \revise{SS}  & \revise{$223.9 \pm 18.5$}  \\ 
    \midrule
     \original{} \vaeabbr\ (8) & $ 10.8 \pm 0.1 $ \\
    \original{} \vaeabbr\ (16)  & $ 9.6 \pm 0.1 $  \\\
    \original{} \vaeabbr\ (32) & $ 8.7 \pm 0.1 $  \\ 
    \vaeabbr\ (8) & \revise{$ 8.1 \pm 3.0$} \\
    \vaeabbr\ (16)  & \revise{$14.1 \pm 7.8$}  \\
    \vaeabbr\ (32) & \revise{$18.5 \pm 4.5$}  \\ 
    \midrule
    \revise{\static{}  \vaeabbr\ (8)} & \revise{$ 8.9 \pm 2.2$} \\
    \revise{\static{} \vaeabbr\ (16)}  & \revise{$13.1 \pm 5.8$}  \\
    \revise{\static{} \vaeabbr\ (32)} & \revise{$6.7 \pm 1.9$}  \\ 
    \midrule
    \original{} \infonceabbr\ (8) &  $ 10.8 \pm 0.1$ \\
    \original{} \infonceabbr\ (16)  & $ 154.8 \pm 0.1  $  \\
    \original{} \infonceabbr\ (32) & $ 138.6 \pm 0.1 $  \\ 
    \infonceabbr\ (8) &  \revise{$13.3 \pm 3.8$} \\
    \infonceabbr\ (16)  & \revise{$61.8 \pm 5.5$ } \\
    \infonceabbr\ (32) & \revise{$84.6 \pm 11.0$}  \\ 
    \midrule
    \static{} \infonceabbr\ (8) &  \revise{$12.3 \pm 3.4$} \\
    \static{} \infonceabbr\ (16)  & \revise{$ 77.1 \pm 4.4$ } \\
    \static{} \infonceabbr\ (32) & \revise{$ 79.6 \pm 8.0 $}  \\ 
    \midrule
    \original{} \brownianabbr\ (8) &  $ 9.2 \pm 0.1 $ \\
    \original{} \brownianabbr\ (16)  & $ 17.8 \pm 0.1 $  \\
    \original{} \brownianabbr\ (32) & $ 10.8 \pm 0.1 $  \\ 
    \brownianabbr\ (8) &  \revise{$16.2 \pm 11.4$} \\
    \brownianabbr\ (16)  & \revise{$21.1 \pm 11.8$ } \\
    \brownianabbr\ (32) & \revise{$51.6 \pm 14.2$ } \\ 
    \midrule
    \static{} \brownianabbr\ (8) &  \revise{$15.4 \pm 9.2$} \\
    \static{} \brownianabbr\ (16)  & \revise{$11.8 \pm 5.4 $ } \\
    \static{} \brownianabbr\ (32) & \revise{$ 13.0 \pm 5.0$ } \\ 
    \midrule 
    \original{} \modelabbr\ (8) & $16.8\pm 0.2$ \\
    \original{} \modelabbr\ (16) &  $ {7.9 \pm 0.1}$ \\
    \original{} \modelabbr\ (32) & $9.3 \pm 0.1$ \\
    \modelabbr\ (8) & \revise{$ 13.1 \pm 4.4$} \\
    \modelabbr\ (16) & \revise{ $15.4 \pm 10.1 $ }\\
    \modelabbr\ (32) &\revise{ $28.1 \pm 12.2$} \\
    \midrule
    \static{} \modelabbr\ (8) & \revise{$12.8 \pm 3.2$} \\
    \static{} \modelabbr\ (16) & \revise{ $13.0 \pm 7.5$ }\\
    \static{} \modelabbr\ (32) &\revise{ $13.0 \pm 7.8$} \\
    \bottomrule
    \caption{\small Percentage of length mismatch (MM) during generation.\revise{The label \original{} are the experiments from the original camera-ready version.} \label{tab:full_length_matching}}
\end{tabular}
\end{table*}
\end{spacing}

\begin{table*}[t]
    \small
    \centering
    \begin{spacing}{0.7}
    \centering
  \begin{tabular}{lc|c|c}
    \toprule
    \multicolumn{1}{c}{\bf Method}  &\multicolumn{1}{c}{WH } &\multicolumn{1}{c}{TM-2} &\multicolumn{1}{c}{ TT}\\
    \midrule\midrule
    \original{} GPT2    & $ 10.7 $  & $86.8$  & $22.0$    \\
    GPT2    & \revise{$10.6 \pm 4.3 $}  & \revise{$81.7 \pm 10.7$}  & \revise{$108.5 \pm 8.6$}    \\
    \midrule
    \original{} \vaeabbr\ (8) & $10.6$  & $83.4$  & $46.2 $    \\
    \original{} \vaeabbr\ (16) & $11.6$   & $73.5$  & $ 35.1$    \\
    \original{} \vaeabbr\ (32) & $15.5$  & $90.2$ & $ 54.5$   \\
    \vaeabbr\ (8) &  \revise{$10.8 \pm 6.3$}  & \revise{ $120.4 \pm 36.3$}  &  \revise{$80.9 \pm 7.1$ }   \\
    \revise{  \vaeabbr\ (16)} & \revise{ $13.2 \pm 6.4$}   & \revise{ $109.4 \pm 51.2$}  &  \revise{$75.0 \pm 4.5$ }   \\
    \vaeabbr\ (32) & \revise{ $11.0 \pm 1.9$}  & \revise{ $66.9 \pm 10.8$} & \revise{ $77.5 \pm 25.0$}   \\
    \midrule
    \revise{\static{} \vaeabbr\ (8)} &  \revise{$8.6 \pm 4.4$}  & \revise{ $123.8 \pm 35.3$}  &  \revise{$81.4 \pm 5.7$ }   \\
    \revise{ \static{}   \vaeabbr\ (16)} & \revise{ $12.9 \pm 5.3$}   & \revise{ $115.5 \pm 60.6$}  &  \revise{$75.2 \pm 3.2$ }   \\
    \revise{ \static{}   \vaeabbr\ (32)} & \revise{ $12.8 \pm 5.3$}  & \revise{ $73.9 \pm 10.9$} & \revise{ $80.2 \pm 23.9$}   \\
    \midrule
    \original{} \infonceabbr\ (8) & $23.1$  & $119.1$  & $111.1 $   \\
    \original{} \infonceabbr\ (16) & $ 38.1$   & $87.9$  & $55.4$    \\
    \original{} \infonceabbr\ (32) & $30.1$  & $113.3$ & $78.5$   \\
    \infonceabbr\ (8) & \revise{$14.9 \pm 6.5$}  & \revise{ $200.3 \pm 27.7$ } & \revise{ $108.8 \pm 9.5$ }  \\
    \infonceabbr\ (16) & \revise{ $ 37.7 \pm 4.0$}   &  \revise{ $198.2 \pm 19.2$}  & \revise{ $68.7 \pm 20.3$  }  \\
    \infonceabbr\ (32) &  \revise{$35.9 \pm 8.5$}  &  \revise{$231.5 \pm 97.1$ }&  \revise{$78.6 \pm 22.5$ }  \\
    \midrule
    \revise{\static{}   \infonceabbr\ (8)} & \revise{$ 13.5 \pm 5.6$}  & \revise{ $120.2 \pm 37.4$ } & \revise{ $ 64.1 \pm 6.0$ }  \\
    \revise{\static{}  \infonceabbr\ (16)} & \revise{ $37.2 \pm 2.7$}   &  \revise{ $93.9 \pm 32.3$}  & \revise{ $66.2 \pm 20.5$  }  \\
    \revise{\static{}  \infonceabbr\ (32)} &  \revise{$25.8 \pm 4.4$}  &  \revise{$124.4 \pm 50.4$ }&  \revise{$ 58.9 \pm 16.9$ }  \\
    \midrule
    \original{} \brownianabbr\ (8) &  $18.1$ & $52.0$  & $34.9$      \\
    \original{} \brownianabbr\ (16) &  $12.7$ & $ 44.9$   & $75.8$  \\
    \original{} \brownianabbr\ (32) & $15.5$  & $47.9$ & $78.5$   \\
    \brownianabbr\ (8) & \revise{ $11.8 \pm 5.1$} & \revise{ $93.4 \pm 20.4$ } & \revise{ $68.2 \pm 28.9$    }  \\
    \brownianabbr\ (16) &  \revise{ $20.8 \pm 11.5$} &  \revise{$106.8 \pm 17.6$}   & \revise{ $111.3 \pm 45.6$ } \\
    \brownianabbr\ (32) &  \revise{$30.2 \pm 13.4$}  & \revise{ $118.6 \pm 17.7$ } & \revise{ $170.5 \pm 36.2$  } \\
    \midrule
    \static{} \brownianabbr\ (8) & \revise{ $ 7.8 \pm 3.2$} & \revise{ $102.5 \pm 14.6$ } & \revise{ $66.3 \pm 26.2$    }  \\
    \static{} \brownianabbr\ (16) &  \revise{ $10.7 \pm 6.6$} &  \revise{$93.7 \pm 11.7$}   & \revise{ $66.2 \pm 34.7$ } \\
    \static{} \brownianabbr\ (32) &  \revise{$14.1 \pm 9.7$}  & \revise{ $111.5 \pm 17.2$ } & \revise{ $82.7 \pm 25.9$  } \\
    \midrule
    \original{} \modelabbr\ (8) & $ {9.6}$  & $31.1 $  & $ 8.0$      \\
    \original{} \modelabbr\ (16) & $15.0$ & $9.3 $  & $ {5.5}$    \\
    \original{} \modelabbr\ (32) & $15.8 $   & $ {5.2}$ & $12.0$    \\
    \modelabbr\ (8) & \revise{ $13.5 \pm 6.1$}  &  \revise{$41.9 \pm 9.5$}  &  \revise{$69.8 \pm 6.2$ }     \\
    \modelabbr\ (16) & \revise{$ 8.5 \pm 4.4$} & \revise{$21.5 \pm 4.7$}  &\revise{ $60.8 \pm 16.6$ }   \\
    \modelabbr\ (32) &\revise{ $13.3 \pm 3.6$}   &  \revise{$33.2 \pm 3.3$} &\revise{ $53.5 \pm 27.4$}    \\
    \midrule
    \static{} \modelabbr\ (8) & \revise{ $13.4 \pm 5.5$}  &  \revise{$15.6 \pm 2.7$}  &  \revise{$36.4 \pm 11.4$ }     \\
    \static{} \modelabbr\ (16) & \revise{$10.8 \pm 5.4$} & \revise{$13.7 \pm 3.4$}  &\revise{ $37.4 \pm 12.9$ }   \\
    \static{} \modelabbr\ (32) &\revise{ $ 10.3 \pm 7.1$}   &  \revise{$ 8.6 \pm 5.3$} &\revise{ $ 30.2 \pm 17.7$}    \\
    \bottomrule
  \end{tabular}
      \captionof{table}{\small Section lengths deviating from expected length in forced long text generation reported in \% ($\downarrow$). 
  \revise{The label \original{} are the experiments from the original camera-ready version.}
  \label{tab:full_forced_long_generation_length}}
\end{spacing}
\end{table*}

\subsection{Ordering}

\revise{
For reference, Table~\ref{tab:full_ordering} shows both the original and new results. The original results are marked with \original. 
}

\paragraph{Differences} 
\revise{
In the previous version, TC performed the best along with the VAE baseline in correct ordering (Table 5). 
In the updated version, there's a lot more overlap between methods. 
}

\paragraph{Causes} 
\revise{The causes are the same as for Table 4. }

\begin{table*}[t]
    \small
    \footnotesize
    \begin{spacing}{0.7}
    \centering
  \begin{tabular}{lc|c|c|c}
    \toprule
    \multicolumn{1}{c}{\bf Method} &\multicolumn{1}{c}{Wikisection} &\multicolumn{1}{c}{Wikihow} &\multicolumn{1}{c}{ TicketTalk}&\multicolumn{1}{c}{Recipe}\\
    \midrule\midrule
    \original{} GPT2   & $50.4 \pm 1.0$ & $61.5 \pm 3.5$ & $75.8 \pm 1.6$ & $36.8 \pm 3.7$   \\
    GPT2   & \revise{$47.4 \pm 5.4$} & \revise{$61.3 \pm 9.1$} &  \revise{$19.7 \pm 2.3$} & \revise{$71.9 \pm 2.5$}   \\
    \midrule
    \original{} \vaeabbr\ (8)    & $57.1 \pm 3.5$  & $66.3 \pm 1.7$  & $66.1 \pm 4.1$  & $71.4 \pm 0.6$   \\
    \original{} \vaeabbr\ (16)    & $47.3 \pm 3.3 $  & $60.5 \pm 2.9$  & $0.8 \pm 0.0$  & $45.6 \pm 0.5$   \\
    \original{} \vaeabbr\ (32)    &  $  58.3 \pm 0.3$  & $60.9 \pm 5.6$  & $38.9\pm 1.7$  & $   87.5 \pm 0.2$   \\
    \vaeabbr\ (8)    & \revise{$49.7 \pm 4.3$}  &\revise{ $56.8 \pm 10.3$}  & \revise{$ 44.0 \pm 10.5$}  & \revise{$61.8 \pm 16.8$}   \\
    \vaeabbr\ (16)    & \revise{$49.3 \pm 5.3$ } & \revise{$67.7 \pm 11.2$}  & \revise{$50.9 \pm 17.4$}  &\revise{ $54.0 \pm 25.7$}   \\
    \vaeabbr\ (32)   & \revise{ $41.3 \pm 5.6$}  & \revise{$62.6 \pm 5.9$}  &\revise{ $23.1 \pm 15.5$}  & \revise{$85.2 \pm 3.7$ }  \\
    \midrule
    \revise{\static{} \vaeabbr\ (8)}    & \revise{$51.0 \pm 4.3$}  &\revise{ $53.9 \pm 11.6$}  & \revise{$45.5 \pm 11.3$}  & \revise{$70.6 \pm 13.9$}   \\
    \revise{\static{} \vaeabbr\ (16)}    & \revise{$49.8 \pm 6.1$ } & \revise{$ 67.9 \pm 8.4$}  & \revise{$52.0 \pm 18.6$}  &\revise{ $ 64.8 \pm 14.7$}   \\
    \revise{\static{} \vaeabbr\ (32) }   & \revise{ $45.9 \pm 6.1$}  & \revise{$ 61.1 \pm 6.0$}  &\revise{ $21.9 \pm 14.8$}  & \revise{$78.8 \pm 3.9$ }  \\
    \midrule
    \original{} \infonceabbr\ (8)    & $34.6 \pm 0.0$  & $59.3 \pm 2.9$  & $30.8 \pm  1.6$  & $68.7 \pm 2.3$    \\
    \original{} \infonceabbr\ (16)    & $35.7 \pm 0.0$ & $31.7 \pm 0.9$  & $63.9 \pm 0.4$ & $78.1 \pm 3.1 $    \\
    \original{} \infonceabbr\ (32)    & $47.9\pm 2.0$  & $16.9 \pm 0.6$  & $45.6 \pm 0.6$ & $85.6 \pm 2.3 $    \\
    \infonceabbr\ (8)   & \revise{ $50.8 \pm 3.8$ } &\revise{ $60.5 \pm 11.9$}  & \revise{ $50.0 \pm 3.4$}  & \revise{$65.6 \pm 15.0$ }   \\
    \revise{  \infonceabbr\ (16) }   & \revise{ $62.3 \pm 6.7$} & \revise{$27.8 \pm 20.1$ } & \revise{$55.4 \pm 19.7$} & \revise{$55.7 \pm 8.5$}    \\
    \revise{\infonceabbr\ (32) }   &\revise{ $64.0 \pm 3.2$ } & \revise{$44.1 \pm 7.9$ } & \revise{$64.8 \pm 6.2$ }& \revise{$81.5 \pm 5.2$ }   \\
    \midrule
    \revise{\static{} \infonceabbr\ (8) }   & \revise{ $51.0 \pm 5.0$ } &\revise{ $ 68.5 \pm 10.4$}  & \revise{ $49.2 \pm 13.4$}  & \revise{$75.9 \pm 7.1$ }   \\
    \revise{\static{} \infonceabbr\ (16) }   & \revise{ $21.2 \pm 4.1$} & \revise{$28.3 \pm 20.9$ } & \revise{$54.3 \pm 21.3$} & \revise{$56.6 \pm 15.7$}    \\
    \revise{\static{} \infonceabbr\ (32) }   &\revise{ $35.4 \pm 7.0$ } & \revise{$48.5 \pm 8.9$ } & \revise{$ 67.8 \pm 10.5$ }& \revise{$77.0 \pm 6.4$ }   \\
    \midrule
    \original{} \brownianabbr\ (8)    & $54.7 \pm 1.9$ & $56.0 \pm 3.5$  & $77.0 \pm 4.7$  & $82.7 \pm  0.4$    \\
    \original{} \brownianabbr\ (16)    & $ 61.4 \pm 1.0 $  & $63.8 \pm 4.7$  & $43.3 \pm 2.4$  & $34.9 \pm 1.5$    \\
    \original{} \brownianabbr\ (32)    & $44.3 \pm 1.4$  & $50.2 \pm 1.2$  & $14.8\pm 1.0$  & $87.1 \pm 0.1$    \\
    \brownianabbr\ (8)   & \revise{$50.1 \pm 3.4$ } & \revise{$61.8 \pm 8.9$ } & \revise{$53.8 \pm 17.3$ } & \revise{ $71.0 \pm 6.9$ }   \\
    \revise{ \brownianabbr\ (16)}   & \revise{$40.7 \pm 6.1$ } & \revise{$51.3 \pm 20.5$ } & \revise{$37.7 \pm 15.4$ } & \revise{$37.3 \pm 16.6$   } \\
    \brownianabbr\ (32)    & \revise{$43.4 \pm 7.4$}  & \revise{$41.9 \pm 15.1$ } & \revise{$ 49.2 \pm 3.7$  } & \revise{$53.2 \pm 7.6$ }   \\
    \midrule
    \revise{\static{} \brownianabbr\ (8) }   & \revise{$51.5 \pm 6.6$ } & \revise{$65.2 \pm 10.1$ } & \revise{$54.2 \pm 19.3$ } & \revise{ $54.9 \pm 17.9$ }   \\
    \revise{\static{} \brownianabbr\ (16) }   & \revise{$42.8 \pm 5.3$ } & \revise{$66.5 \pm 15.1$ } & \revise{$22.8 \pm 19.1$ } & \revise{$45.0 \pm 14.0$   } \\
    \static{} \brownianabbr\ (32)    & \revise{$46.3 \pm 4.2$}  & \revise{$66.7 \pm 10.3$ } & \revise{$23.4 \pm 12.6$  } & \revise{$30.1 \pm 12.4$ }   \\
    \midrule
    \original{} \modelabbr\ (8) & $52.3 \pm 2.5$ & $  76.7 \pm 7.5$ & $81.8 \pm 1.0$ &	$41.7 \pm 1.0$   \\
    \original{} \modelabbr\ (16) & $ 57.9 \pm 1.0$ &  $ 63.1 \pm 6.2$ & $  88.0 \pm 1.3$ &	$64.1 \pm 1.0$   \\
    \original{} \modelabbr\ (32) & $36.5 \pm 2.8$  & $59.1 \pm 5.5$ & $83.6 \pm 1.3$ &	$76.4 \pm 1.0$   \\
    \modelabbr\ (8) & \revise{$49.2 \pm 10.5$ }& \revise{$64.3 \pm 14.7$ } & \revise{$48.3 \pm 23.8$ }&	\revise{$56.0 \pm 10.4$ }  \\
    \modelabbr\ (16) & \revise{$42.5 \pm 7.0$} &  \revise{$66.2 \pm 17.7$ } & \revise{$ 58.8 \pm 11.7$} &	\revise{$66.6 \pm 10.0$ }  \\
    \revise{ \modelabbr\ (32)} &\revise{ $36.9 \pm 7.7$ } &\revise{ $69.7 \pm 12.3$ } & \revise{$35.4 \pm 16.5 $} &	\revise{$77.0 \pm 3.5$}   \\
    \midrule
    \static{} \modelabbr\ (8) & \revise{$ 50.5 \pm 11.2$} & \revise{$ 69.7 \pm 12.3$ } & \revise{$ 58.1 \pm 19.6 $ }&	\revise{$55.7 \pm 11.3$ }  \\
    \static{} \modelabbr\ (16) & \revise{$ 48.5 \pm 5.5$} &  \revise{$67.9 \pm 19.0$ } & \revise{$67.3 \pm 8.9$} &	\revise{$70.3 \pm 7.5$ }  \\
    \static{} \modelabbr\ (32) &\revise{ $ 52.1 \pm 6.4 $ } &\revise{ $ 71.5 \pm 10.5$ } & \revise{$ 49.1 \pm 18.0$} &	\revise{$82.4 \pm 3.3$}   \\
    \bottomrule
  \end{tabular}
  \vspace{5pt}
  \captionof{table}{\small Ordering in forced long text generation. ROC Stories and TM-2 omitted because they are not applicable.
  \revise{The label \original{} are the experiments from the original camera-ready version.}
  \label{tab:full_ordering}
  }
  
    \hspace{1em}
    \end{spacing}
\end{table*}

\section{Original text-infilling and human evaluation results}

\revise{
Due to the difficulty of re-running the human evaluation results, we have decided not to include an updated version of the human evaluation results. 
This impacts our text-infilling setting and long text generation setting. 
For transparency, we have included the text and results from the original version below. 
}

\paragraph{Original text for text infilling experiment}
We also report human evaluations on how coherent the generated sentence is as a fill-in sentence. 
Participants were asked to rank the generated fill-in sentence from ILM, LM, and \model\ on a scale of 1-5 (not reasonable to very reasonable). 
Appendix~\ref{sec:human_expt_setup} includes more details on the human experiment setup.

We evaluate the text coherence with the BLEU score \citep{papineni2002bleu}, ROUGE \citep{lin-2004-rouge}, BLEURT \citep{sellam2020bleurt} and BERTScore \citep{zhang2019bertscore} between the generated and ground truth infill sentence. Due to space constraints, the last three metrics are in the Appendix, Table~\ref{tab:bertscore}.
We also report human evaluations on how coherent the generated sentence is as a fill-in sentence. 
Participants were asked to rank the generated fill-in sentence from ILM, LM, and \model\ on a scale of 1-5 (not reasonable to very reasonable). 
Appendix~\ref{sec:human_expt_setup} includes more details on the human experiment setup.

The BLEU scores are summarized in Table~\ref{tab:bleu_scores}.
\model\ generates fill-in sentences that much more closely overlap with the ground truth than ILM and LM.
The ablated methods perform worse to varying degrees. 
On average, \vaeabbr\ performs worse than LM and \infonceabbr. 
This suggests that the interpolating embeddings learned via the variational objective yields embeddings which hurt the autoregressive model's ability to generate locally coherent text.

\paragraph{Original text for forced long text human evaluations}
The third metric is human evaluations, which measures the long-tail text quality. We examine the long-tail behavior as the complete 1000-token generation is long.

[...]  These results are additionally confirmed by our human evaluation experiments: human evaluators rank \model's extrapolation ability better on all latent dimension settings than that of GPT2 (Table~\ref{tab:human_eval_long_text_generation}).

\begin{minipage}[t]{\textwidth}
    \small
    \footnotesize
    \begin{spacing}{0.7}
    \hspace{1em}
    
    \begin{minipage}[b]{0.5\textwidth}
    \centering
  \begin{tabular}{l|c}
    \toprule
    \multicolumn{1}{c}{\bf Method}  &\multicolumn{1}{c}{\bf BLEU ($\uparrow$)}
\\ 
    \midrule\midrule
    LM & $1.54 \pm 0.02$ \\  
    ILM &  $3.03 \pm 0.11$  \\
    \midrule
    \vaeabbr\ (8)  & $0.75 \pm 0.17$  \\
    \vaeabbr\ (16)  & $0.62 \pm 0.07$  \\
    \vaeabbr\  (32) &  $0.03 \pm 0.0$ \\
    \midrule
    \infonceabbr\  (8) & $2.9 \pm 0.3$  \\
    \infonceabbr\ (16)  & $0.9 \pm 0.0$  \\
    \infonceabbr\= (32)  &  $1.0 \pm 0.1$ \\
    \midrule
    \modelabbr\ (8)  & $ 3.80 \pm 0.06$  \\
    \modelabbr\ (16) & $4.30 \pm 0.02$  \\
    \modelabbr\ (32)  &  $\graycell \bf 5.4 \pm 0.11$ \\
    \bottomrule
  \end{tabular}
  \captionof{table}{\small BLEU on ground truth infill and generated sentence. \label{tab:bleu_scores}}
    \end{minipage}
    \begin{minipage}[b]{0.50\textwidth}
    \footnotesize
        \centering
        \begin{minipage}[b]{\textwidth}
        \centering
            \begin{tabular}{l|c}
            \toprule
            \multicolumn{1}{c}{\bf Method} &\multicolumn{1}{c}{\bf Human}\\ 
            \midrule\midrule
            LM  & $2.4 \pm 0.06$ \\  
            ILM &  $\graycell \bf 3.77 \pm 0.07$  \\
            \midrule
            \modelabbr\ (8)  & $ \bf  3.64 \pm 0.07$  \\
            \bottomrule
            \end{tabular}
            \small
            \captionof{table}{\small  Human evaluations on text infilling. Scores were ranked between 1 and 5. Higher is better. \label{tab:human_eval_text_infilling}}
        \end{minipage}
        \vspace{1em}
  \begin{minipage}[b]{\textwidth}
    \centering
    \begin{tabular}{l|c}
        \toprule
        \multicolumn{1}{c}{\bf Method} &\multicolumn{1}{c}{\bf Human}
        \\ 
        \midrule\midrule
        GPT2  & $2.8 \pm 0.06$ \\  
        \midrule
        \modelabbr\ (8)  & $\graycell \bf 3.6 \pm 0.07$  \\
        \modelabbr\ (16)  & $ 3.4 \pm 0.07$  \\
        \modelabbr\ (32)  & $ 3.3 \pm 0.07$  \\
        \bottomrule
    \end{tabular}
    \footnotesize
    \captionof{table}{ \small Human evaluations on tail end quality in forced long text generation.  Scores were ranked between 1 and 5. Higher is better.  \label{tab:human_eval_long_text_generation}}
    \end{minipage}
    \end{minipage}
    \end{spacing}
\end{minipage}

\section{Original generation examples } \label{sec:app_generation_examples}

The following generation examples were generated from the  original paper manuscript.

We provide example outputs from the different models on text infilling. We denote the prefix sentences as ``LEFT CONTEXT'', the suffix sentences as ``RIGHT SENTENCES'', the generated output from the models as ``GENERATED'' and the ground truth infill sentence as ``GT''.

\subsection{Text infilling \label{sec:examples_text_infilling}}

\paragraph{LM} See Table~\ref{tab:app_lm_text_infilling}.

\begin{minipage}{\textwidth}
\captionof{table}{Example of text infilling with \citet{donahue2020ilm}'s \textbf{LM} model on ROCStories.} \label{tab:app_lm_text_infilling}
\small
\begin{tabular}{ p{13cm} } 
\toprule
LEFT CONTEXT: John couldn't wash his clothes because his washing machine was broken. \\
GENERATED:  He had an idea.\\
GT:  John didn't know how to fix washing machines. \\
RIGHT CONTEXT:  He bought a book to tell him how to repair it. He changed the fan belt. When he plugged it back in, it worked.
\\
\noalign{\vskip 2mm}   
\midrule
LEFT CONTEXT: I found a stray cat yesterday. I gave it some milk. It came back again today. I was going to give it some milk but didn't. \\
GENERATED:  I guess I need to get something.\\
GT:  I had run out of milk.\\
RIGHT CONTEXT: 
\\
\noalign{\vskip 2mm}   
\midrule
LEFT CONTEXT: Toni took a trip to the local fair with her family. She immediately laid eyes on a pretty balloon. The balloon was red and star shaped. Unfortunately it slipped out of her hands. \\
GENERATED:  Her swollen fingers had horrible rashes. \\ 
GT:  She was sad to see it go.\\ 
RIGHT CONTEXT: 
\\
\noalign{\vskip 2mm}   
\midrule
LEFT CONTEXT:  Miles was a roofer. He loved his job and was very good at it.  \\ 
GENERATED:  He dreamed of finding a good title. \\
GT:  However, one day he was not paying attention and fell off the roof. \\
RIGHT CONTEXT:  Luckily, he only broke his leg. Miles decided that he needed to be more careful.
\\
\noalign{\vskip 2mm}   
\midrule
LEFT CONTEXT:  Lindsey was feeling especially antsy at school. \\ 
GENERATED:  The school wasn't paying much attention to her.  \\
GT:  She did not want to go to Lacrosse practice after school.\\
RIGHT CONTEXT:  She tried to fake a sickness, but nobody believed her. Lindsey ended up skipping practice without an excuse. She hoped she wouldn't get in trouble. 
\\
\noalign{\vskip 2mm}   
\midrule
LEFT CONTEXT: Last night Jack went on a blind date. When he arrived at the restaurant a he heard someone yell his name. He spotted a blonde waving from across the room. \\ 
GENERATED: She was a waiter.  \\ 
GT:  She was nothing like the photo she sent.\\ 
RIGHT CONTEXT:  But he figured he would try to make the most of it.
\\
\noalign{\vskip 2mm}   
\midrule
LEFT CONTEXT: Rico was working really hard on his essay for English. After working on it all night, he decided to take a short nap. He slept in and was late for for school. When he got to school he noticed that his essay was gone. \\ 
GENERATED:  teacher found the essay in his bag and he passed it.\\ 
GT:  Rico's teacher let him turn in the essay the next day. \\ 
RIGHT CONTEXT: 
\\
\bottomrule
\end{tabular}
\end{minipage}

\paragraph{ILM} See Table~\ref{tab:app_ilm_text_infilling}.

\begin{minipage}{\textwidth}
\captionof{table}{Example of text infilling with \citet{donahue2020ilm}'s \textbf{ILM} model on ROCStories.} \label{tab:app_ilm_text_infilling}
\small
\begin{tabular}{ p{13cm} } 
\toprule
LEFT CONTEXT: \\
GENERATED: My 98 year old friend and I played blackjack yesterday. \\ 
GT: Last week's family game night was intense. \\ 
RIGHT CONTEXT:  We were playing Monopoly and nobody made any headway for hours. Everyone was trying their hardest to win and the game kept going. It wasn't until we finally decided to check the rules that we knew why. There were many different pieces missing.
\\
\noalign{\vskip 2mm}   
\midrule
LEFT CONTEXT: Tom was jealous of his brother. His brother was older and stronger. His brother went wherever he wanted. \\
GENERATED:  Tom decided to use steroids. \\
GT:  One day his brother was grounded for staying out too late. \\ 
RIGHT CONTEXT:  This made Tom really happy.
\\
\noalign{\vskip 2mm}   
\midrule
LEFT CONTEXT: His first time at the opera. He only went after his girlfriend begged. He sat for three hours in sheer boredom. Finally it was over. \\ 
GENERATED:  He turned on the tv and heard her thanking him. \\ 
GT:  He learned he didn't like the opera.\\
RIGHT CONTEXT: 
\\
\noalign{\vskip 2mm}   
\midrule
LEFT CONTEXT: My best friend played a video game online. One day, she met a nice boy there. They talked every day and fell in love. They finally met in person and hit it off. \\
GENERATED:  Now, they are best friends. \\
GT:  The two became a very loving couple. \\ 
RIGHT CONTEXT: 
\\
\noalign{\vskip 2mm}   
\midrule
LEFT CONTEXT: The family decided to go to the park for Veteran's Day.\\ 
GENERATED:  They asked permission from the park manager. \\
GT:  We had a barbecue outside. \\ 
RIGHT CONTEXT:  In addition, everyone decided to take a swim in the water. Before we left, we had a huge cake. We then swam for an hour at the lake.
\\
\noalign{\vskip 2mm}   
\midrule
LEFT CONTEXT: Jennifer always wanted to have the birthday of her dreams. Her parents surprised her with a big birthday party at the park. All of her friends and family came to celebrate her special day. She was so happy as she blew out the candles on her cake. \\ 
GENERATED:  Her parents had made a big surprise to her by turning the cake into a carnival. \\ 
GT:  It was the best day of birthday she could have imagined. \\
RIGHT CONTEXT: 
\\
\noalign{\vskip 2mm}   
\midrule
LEFT CONTEXT: Terry's daughter was in extreme pain in her mouth. \\ 
GENERATED:  Terry and her husband decided to stop the baby from teething. \\ 
GT:  After an evaluation, Terry realized that her daughter was teething. \\
RIGHT CONTEXT:  Once Terry realized she was teething, Terry administered medication. After the medication, Terry placed teething gel on the baby's teeth. Terry got the baby to calm down.
\\
\bottomrule
\end{tabular}
\end{minipage}

\paragraph{\model} See Table~\ref{tab:app_cl8_text_infilling}.

\begin{minipage}{\textwidth}
\captionof{table}{Example of text infilling with \textbf{\model} (d=8) on ROCStories.} 
\small
\begin{tabular}{ p{13cm} } 
\label{tab:app_cl8_text_infilling}\\
\toprule
LEFT CONTEXT:My day on Sunday this week went as expected. My family and I attended Church from 8 to 11 in the morning. \\
GENERATED: We went to a local church in the afternoon . \\
GT: Afterwards, we went to eat dinner at a restaurant. \\
RIGHT CONTEXT: The rest of my day was spent doing homework assignments. I then went to sleep feeling happy about my life. \\ 
\\
\noalign{\vskip 2mm}   
\midrule
LEFT CONTEXT: His first time at the opera. He only went after his girlfriend begged. He sat for three hours in sheer boredom. Finally it was over.  \\ 
GENERATED: He was so happy he didn't want to leave. \\ 
GT: He learned he didn't like the opera. \\ 
RIGHT CONTEXT:
\\
\noalign{\vskip 2mm}   
\midrule
LEFT CONTEXT: 
My best friend played a video game online. One day, she met a nice boy there. They talked every day and fell in love. They finally met in person and hit it off.  \\ 
GENERATED: Until that day, they were married.  \\ 
GT: The two became a very loving couple.  \\ 
RIGHT CONTEXT:
\\
\noalign{\vskip 2mm}   
\midrule
LEFT CONTEXT: 
The other day at the clinic I had to help put a dog down. He seemed really sad and lonely. Like he knew what was going to happen.  \\ 
GENERATED: He was going to die soon . \\ 
GT: As we laid it down and it took its final breaths it stared at me.  \\ 
RIGHT CONTEXT: I stayed calm, but cried after we were finished. \\ 
\\
\noalign{\vskip 2mm}   
\midrule
LEFT CONTEXT: 
Tom was jealous of his brother. His brother was older and stronger. His brother went wherever he wanted.  \\ 
GENERATED: Tom was jealous of his brother.  \\ 
GT: One day his brother was grounded for staying out too late.  \\ 
RIGHT CONTEXT: 
\\
\noalign{\vskip 2mm}   
\midrule
LEFT CONTEXT: 
Jays habit of buying expensive beer was catching up to him. He was spending more money on beer than food. He needed to find another source of income to support this habit. A friend recommended he try out Amazon MTurk.  \\ 
GENERATED: He found the site and bought a few beers. \\ 
GT: Jay become slightly less poor from Amazon Mturk. \\ 
RIGHT CONTEXT:\\
\\
\noalign{\vskip 2mm}   
\midrule
LEFT CONTEXT: 
John couldn't wash his clothes because his washing machine was broken.  \\ 
GENERATED: John went to the store to buy a new one .  \\ 
GT: John didn't know how to fix washing machines.  \\ 
RIGHT CONTEXT: He bought a book to tell him how to repair it. He changed the fan belt. When he plugged it back in, it worked. 
\\
\bottomrule
\end{tabular}
\end{minipage}

\begin{minipage}{\textwidth}
\captionof{table}{Example of text infilling with \textbf{\model} (d=16) on ROCStories.} 
\label{app:app_cl16_text_infilling}
\begin{tabular}{ p{13cm} } 
\toprule
LEFT CONTEXT: Tom was jealous of his brother. His brother was older and stronger. His brother went wherever he wanted.  \\ 
GENERATED:  Tom's brother was very jealous of his brother.  \\ 
GT:  One day his brother was grounded for staying out too late. \\ 
RIGHT CONTEXT:  This made Tom really happy.
\\
\noalign{\vskip 2mm}   
\midrule
LEFT CONTEXT: Jackie was 11 and had to get braces. She was worried about what her friends would think. She tried to hide them when she first got them.  \\ 
GENERATED:  But she was too embarrassed to tell them .   \\ 
GT:  Eventually her friends saw them and she was embarrassed.  \\ 
RIGHT CONTEXT:  Her friends noticed she was embarrassed and decided to comfort her.
\\
\noalign{\vskip 2mm}   
\midrule
LEFT CONTEXT: 
Sally was going to surprise the office with a cake. Sally felt that a cake would be a good way to make them smile. She went to the supermarket to pick up the cake.  \\ 
GENERATED:  She bought the cake and was very happy with it. \\ 
GT:  At the office she gathered the employees around the conference table.  \\ 
RIGHT CONTEXT:  She then brought out a cake and they all felt better.
\\
\noalign{\vskip 2mm}   
\midrule
LEFT CONTEXT: Lars was playing XBOX. His controller stopped working during a game. Lars didn't have a car so he had to walk all the way to the store. The store was being remodeled when he got there so he went to another.  \\ 
GENERATED:   Video games were all over the floor.  \\ 
GT:  Lars wasn't able to find a controller.  \\ 
RIGHT CONTEXT: 
\\
\noalign{\vskip 2mm}   
\midrule
LEFT CONTEXT: Emma had been working as a dishwasher. Her hands cracked and bled from the hot soapy water. Then her mom noticed and concocted a special salve for her. Emma used the salve every night before bed.  \\
GENERATED:  Soon she was able to keep her hands clean and happy. \\
GT:  Her hands got better. \\
RIGHT CONTEXT:
\\
\noalign{\vskip 2mm}   
\midrule
LEFT CONTEXT: Jerome dribbled the ball quickly. Sam tried to grab it from him, but wasn't fast enough. He chased Jerome down the court.  \\ 
GENERATED:  He scored a point with a long shot .  \\ 
GT:  Jerome pushed Sam backwards and threw the ball.  \\ 
RIGHT CONTEXT:  Jerome scored points for his team.
\\
\noalign{\vskip 2mm}   
\midrule
LEFT CONTEXT: Class started in 10 minutes and I had a math assignment due. My older friend volunteered to do it for me. Unfortunately, my teacher found out about the copying.  \\ 
GENERATED:  She was very upset with me for doing the assignment.  \\ 
GT:  She crumbled my paper and threw it in the trash. \\ 
RIGHT CONTEXT:  My teacher gave me an F for the assignment.
\\
\bottomrule
\end{tabular}
\end{minipage}

\begin{minipage}{\textwidth}
\captionof{table}{Example of text infilling with \textbf{\model} (d=32) on ROCStories.} 
\label{app:app_cl32_text_infilling}
\begin{tabular}{ p{13cm} } 
\toprule
LEFT CONTEXT: I went to the park to play frisbee with my dog.  \\ 
GENERATED: We played all day . \\ 
GT:  I tossed the frisbee to my dog and he would catch it in his mouth. \\ 
RIGHT CONTEXT:  I accidentally threw the frisbee too far. The frisbee landed into the pond. But my dog went and got it.
\\
\noalign{\vskip 2mm}   
\midrule
LEFT CONTEXT: I was tired of working at Walmart. The hours were bad. The store music was awful. \\ 
GENERATED:  I was very bored . \\ 
GT:  I handed my two weeks in to the manager. \\ 
RIGHT CONTEXT:  I then found another job and was happy.
\\
\noalign{\vskip 2mm}   
\midrule
LEFT CONTEXT: Sam bought a new SUV. It was all wheel drive. He figured he would take it off road. He hit a few hard bumps and broke his suspension.  \\ 
GENERATED:  Unfortunately he had to pay a lot of money for it. \\ 
GT:  Sheepishly, he brought it to the dealership for repair. \\ 
RIGHT CONTEXT: 
\\
\noalign{\vskip 2mm}   
\midrule
LEFT CONTEXT: Missy got drunk and went to get a tattoo. She decided to get a tattoo on her forehead. The next day, Missy was horrified at what she had done.  \\ 
GENERATED:  Her tattoo was on her forehead! \\ 
GT:  Missy scraped up her money to pay for a tattoo removal procedure. \\ 
RIGHT CONTEXT:  After much wasted money, the tattoo was gone.
\\
\noalign{\vskip 2mm}   
\midrule
LEFT CONTEXT: Jake was going on a road trip to see his family. He Got in the car and drove. \\ 
GENERATED:  Jake's family was driving down the road . \\ 
GT:  The car's tires exploded due to too much air. \\ 
RIGHT CONTEXT:  Jake hitchhiked for 30 miles. When Jake got to his family he was happy his trip was over.
\\
\noalign{\vskip 2mm}   
\midrule
LEFT CONTEXT: Bob decided to start a business.  \\ 
GENERATED: However, he did not know the market at all . \\ 
GT:  He opened up a grocery store and was doing very well. \\ 
RIGHT CONTEXT:  After a year, his profits dropped and he had to declare bankruptcy. Bob was sad to see his business fail. Bob worked hard and reopened his business.
\\
\noalign{\vskip 2mm}   
\midrule
LEFT CONTEXT: I hadn't seen my girlfriend in a while. She got a new job so it's hard to talk. The job takes up all of her time.  \\ 
GENERATED:  I had to ask her to go out with me . \\ 
GT:  Finally she called me to hang out. \\ 
RIGHT CONTEXT:  I was really happy to see her and we made plans.
\\
\bottomrule
\end{tabular}
\end{minipage}

\paragraph{\vae} See Table~\ref{tab:app_vae_text_infilling}.

\begin{minipage}{\textwidth}
\captionof{table}{Example of text infilling with \textbf{\vae} on ROCStories.} \label{tab:app_vae_text_infilling}
\small
\begin{tabular}{ p{13cm} } 
\toprule
LEFT CONTEXT:  I went to the park to play frisbee with my dog. \\
GENERATED:  served served served \textbf{[...]} \\
GT:  I tossed the frisbee to my dog and he would catch it in his mouth. \\
RIGHT CONTEXT:  I accidentally threw the frisbee too far. The frisbee landed into the pond. But my dog went and got it.
\\
\noalign{\vskip 2mm}   
\midrule
LEFT CONTEXT: Tim had a dentist appointment today. He was going to get his wisdom teeth pulled.  \\ 
GENERATED: .  . \\ 
GT:  His dentist numbed his gums. \\ 
RIGHT CONTEXT:  His wisdom teeth popped out instantly. It was less hassle than he thought it would be.
\\
\noalign{\vskip 2mm}   
\midrule
LEFT CONTEXT: Tom was studying for the big test. He then fell asleep do to boredom.  \\ 
GENERATED:  The The The The The The The The The \textbf{[...]}  \\ 
GT:  He slept for five hours.  \\ 
RIGHT CONTEXT:  He woke up shocked. Tom hurried to study as much as possible before the test.
\\
\bottomrule
\end{tabular}
\end{minipage}

\paragraph{\infonce}

 See Table~\ref{tab:app_infonce_text_infilling}.

\begin{minipage}{\textwidth}
\captionof{table}{Example of text infilling with \textbf{\infonce} on ROCStories.} \label{tab:app_infonce_text_infilling}
\small
\begin{tabular}{ p{13cm} } 
\toprule
LEFT CONTEXT: Tom was jealous of his brother. His brother was older and stronger. His brother went wherever he wanted.  \\ 
GENERATED:  he wasn't like him. he wasn't like him.............................................................  isn't...  doesn't...  doesn't..  don't.  don't.  don't.  don't.  don't.  don't.  don't.  don't.  don't.  don't.  don't.  don't.  don't.  don't.  don't.  don't.  don't.  don't.  don  don't.  don  don't  don't.  don  don't  don't  don't  don't  don't  don't  don't  don't  don  don  don  don  \textbf{[...]}  \\ 
GT:  One day his brother was grounded for staying out too late.  \\ 
RIGHT CONTEXT:  This made Tom really happy.
\\
\noalign{\vskip 2mm}   
\midrule
LEFT CONTEXT: 
The family decided to go to the park for Veteran's Day.  \\ 
GENERATED: I didn't I knew why why why why \textbf{[...]}  \\ 
GT:  We had a barbecue outside.  \\ 
RIGHT CONTEXT:  In addition, everyone decided to take a swim in the water. Before we left, we had a huge cake. We then swam for an hour at the lake.
\\
\bottomrule
\end{tabular}
\end{minipage}

\subsection{Forced long text generation}

\begin{figure*}[t]
\begin{minipage}{\textwidth}
    \centering
    \centering
    \small
    \newcommand{\gw}{140mm}
    \includegraphics[width=\gw]{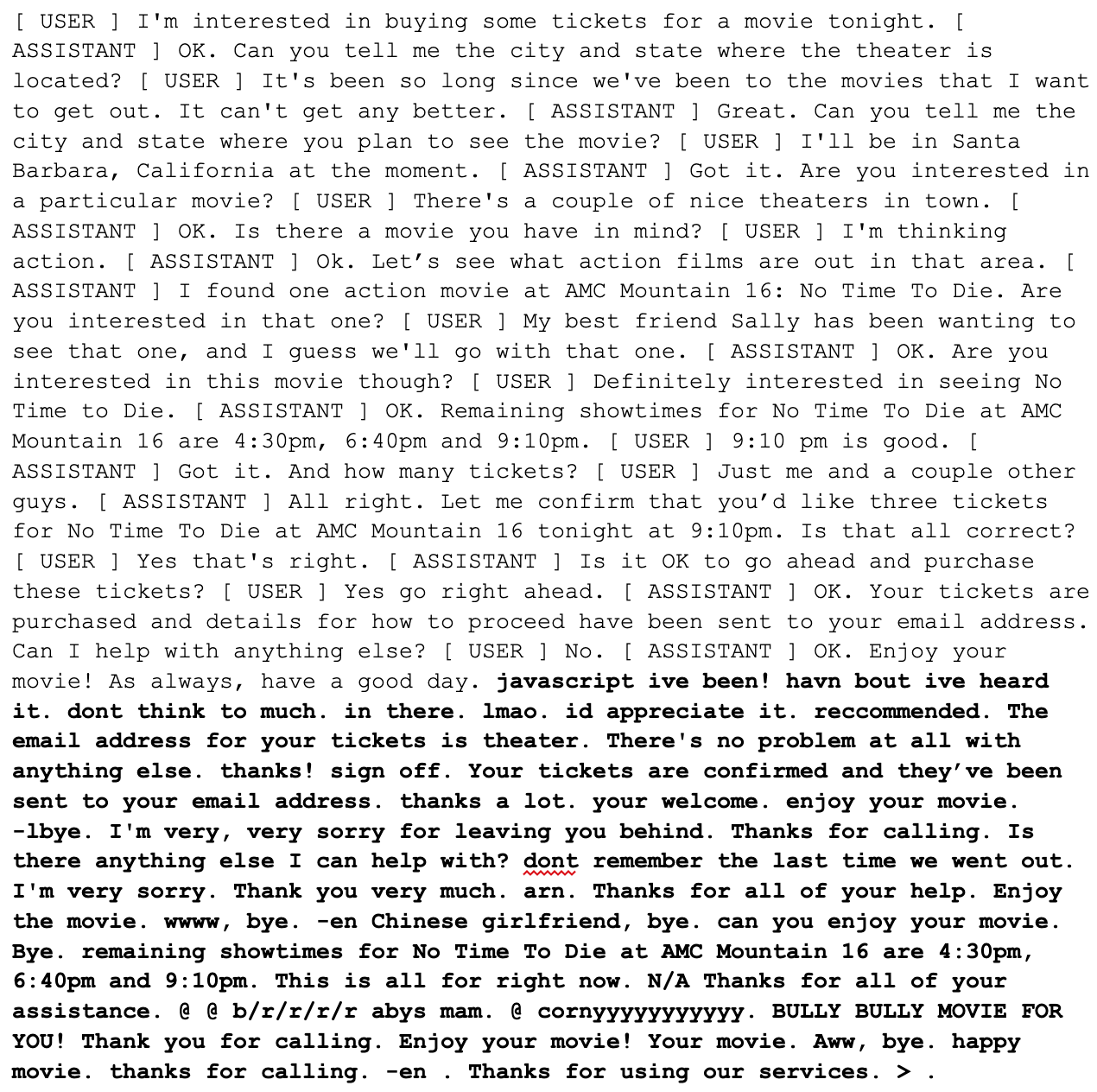}
    \caption{
    \small
    Example 1 of GPT2 forced long text generation.
    }
    \label{fig:gpt2_long_text_example_1}
\end{minipage}
\end{figure*}

\begin{figure*}[t]
\begin{minipage}{\textwidth}
    \centering
    \centering
    \small
    \newcommand{\gw}{140mm}
    \includegraphics[width=\gw]{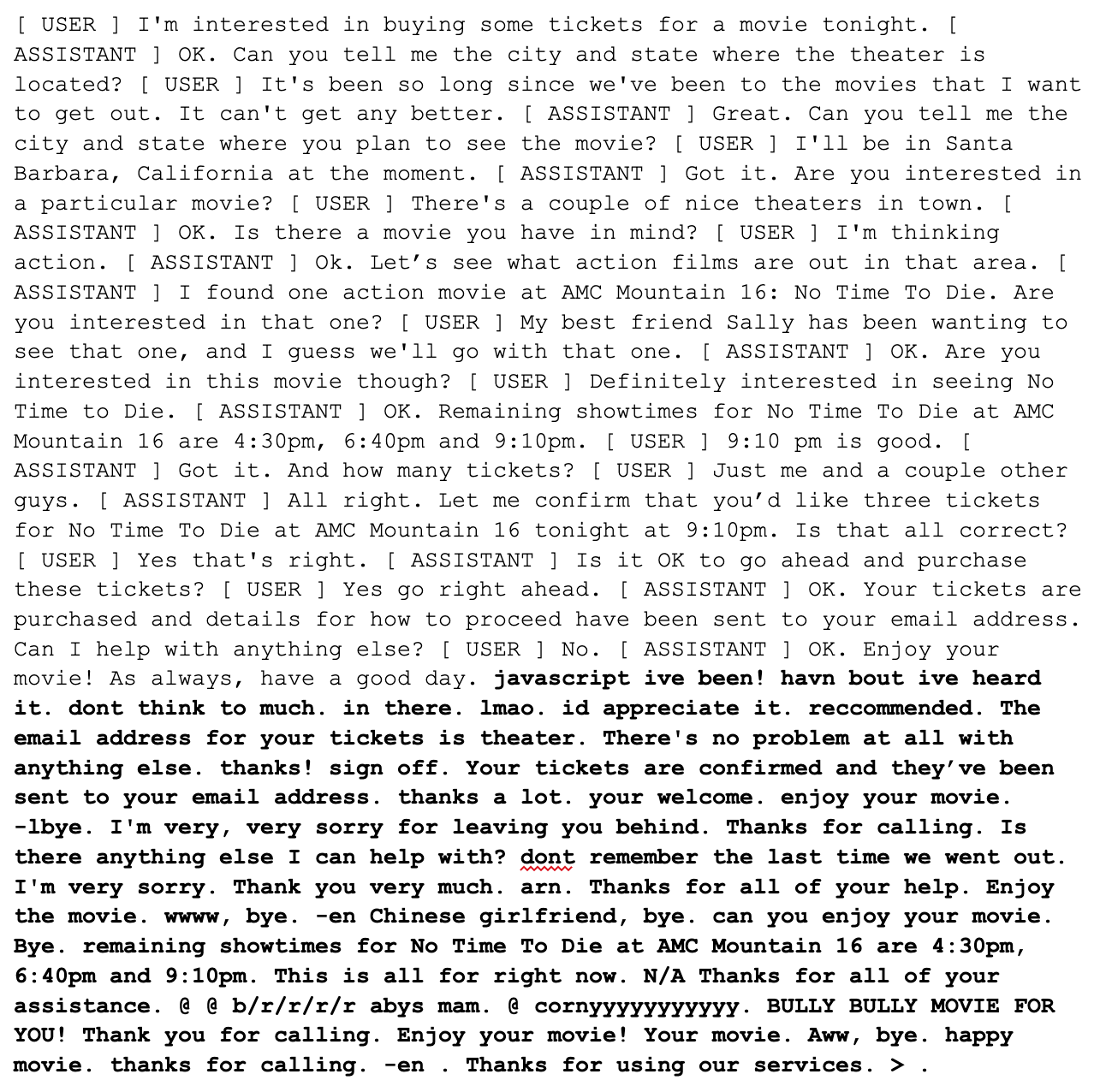}
    \caption{
    \small
    Example 2 of GPT2 forced long text generation.
    }
    \label{fig:gpt2_long_text_example_2}
\end{minipage}
\end{figure*}

We include examples of GPT2 long text generation in Figures~\ref{fig:gpt2_long_text_example_1} and \ref{fig:gpt2_long_text_example_2}.

\begin{figure*}[t]
\begin{minipage}{\textwidth}
    \centering
    \centering
    \small
    \newcommand{\gw}{140mm}
    \includegraphics[width=\gw]{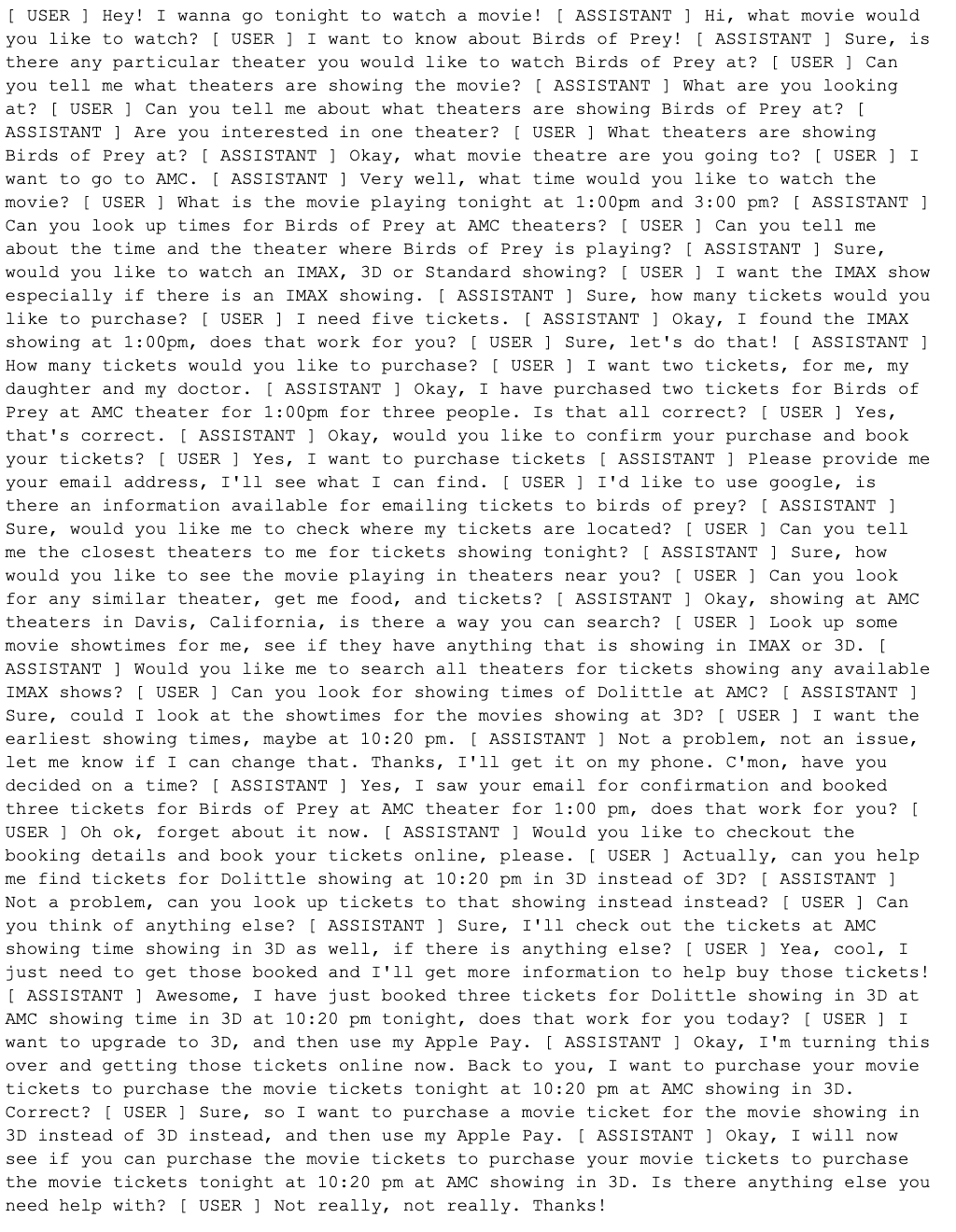}
    \caption{
    \small
    Example 1 of \model\ forced long text generation.
    }
    \label{fig:cl8_long_text_example_1}
\end{minipage}
\end{figure*}

\begin{figure*}[t]
\begin{minipage}{\textwidth}
    \centering
    \centering
    \small
    \newcommand{\gw}{140mm}
    \includegraphics[width=\gw]{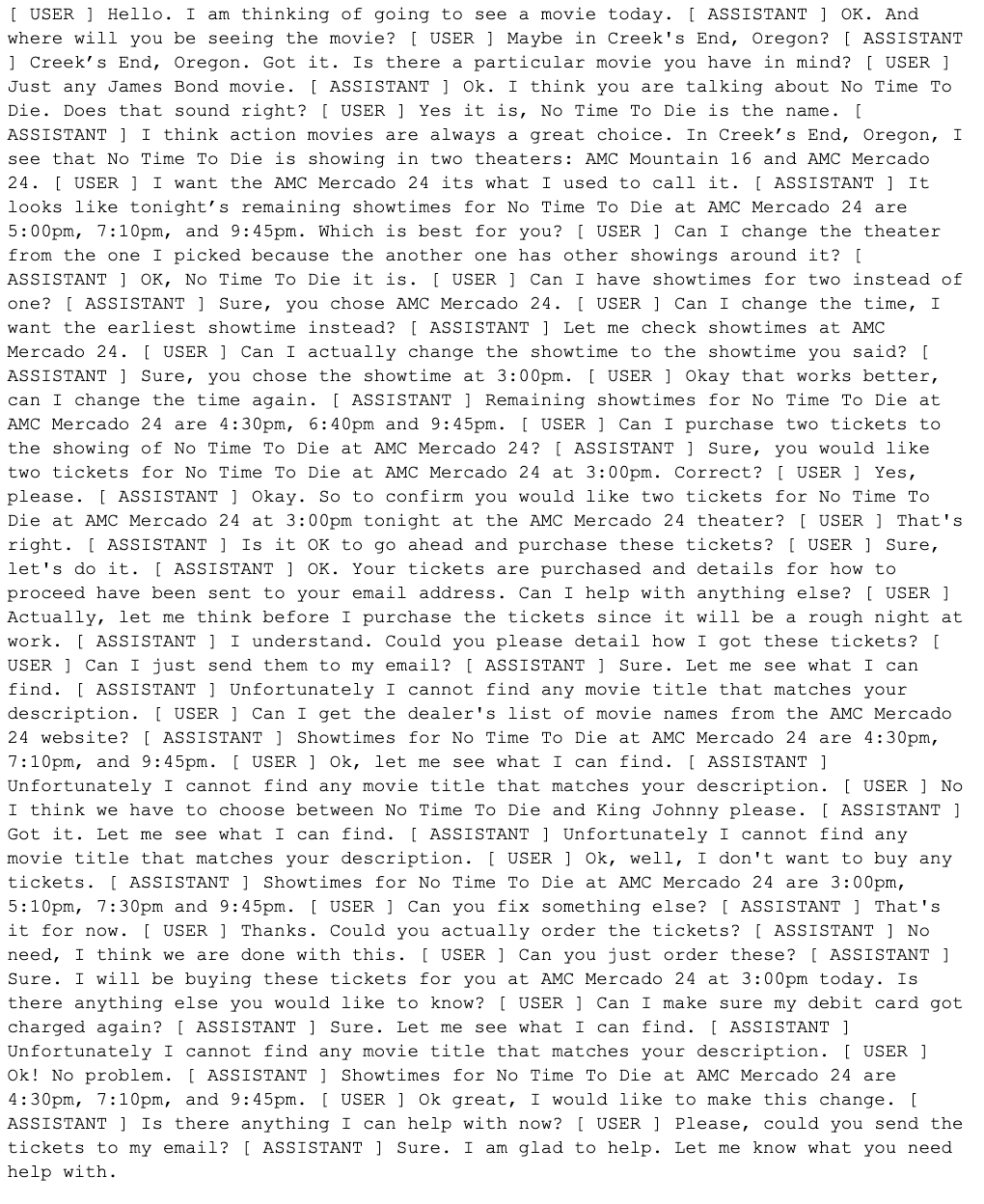}
    \caption{
    \small
    Example 2 of \model\ forced long text generation.
    }
    \label{fig:cl8_long_text_example_2}
\end{minipage}
\end{figure*}

\section{Human experimental setup} \label{sec:human_expt_setup}

We use Mechanical Turk to collect human evaluation for long text generation and text infilling models. Examples from all the systems are selected randomly and each example is evaluated by $10$ annotators. 

\subsection{Long Text Generation Evaluation}

The annotators are provided with a dialogue between a movie assistant and a user generated by GPT and Time Control, where the goal of the assistant is to help the user to book a movie ticket. We show the last few generated interactions between the assistant and the user to the annotators, and ask them to assess whether the provided snippet could be a reasonable ending for such a conversation. We use a 5-point Likert scale where $1$ corresponds to a ``Very Unreasonable Ending'' and $5$ corresponds to a ``Very Reasonable Ending''.

\subsection{Text Infilling Evaluation}

\revise{Note: These results were generated from the original ICLR camera-ready (v1).}

Given the preceding and following context, we ask annotators to evaluate whether the generated fill-in sentence is a reasonable completion. We evaluate generated fill-in sentence by ILM, LM and Time Control on a scale of 1-5.

\begin{table}[ht]
  \centering
  \begin{tabular}{lc|ccc|ccc|c}
    \toprule
    \multicolumn{1}{c}{\bf }  &\multicolumn{1}{c}{\bf } &\multicolumn{3}{c}{\bf BERTScore} &\multicolumn{3}{c}{\bf ROUGE}&\multicolumn{1}{c}{\bf BLEURT}
\\ 
    \multicolumn{1}{c}{\bf Method}  &\multicolumn{1}{c}{\bf Latent dim $d$} &\multicolumn{1}{c}{\bf Precision} &\multicolumn{1}{c}{\bf Recall}&\multicolumn{1}{c}{\bf F1}&\multicolumn{1}{c}{\bf 1-F1}&\multicolumn{1}{c}{\bf 2-F1}&\multicolumn{1}{c}{\bf L-F1}&\multicolumn{1}{c}{\bf }
\\ 
    \midrule\midrule
    LM  & - & $0.45 $ &  $0.50 $ & $0.47 $ & $22.6 $ & $ 1.8$ & $ 14.0$ & $0.31 $\\  
    ILM &  - &  $0.50 $&  $0.51 $ &  $0.50 $ & $21.6 $ & $ 1.8$ & $ 20.7$ & $ 0.33$ \\
    \midrule
    VAE &  $8$  & $ 0.21$ &  $0.26 $ &  $ 0.21$ & $ 1.1$ & $0.9 $ & $1.1 $ & $0.27 $ \\
    VAE  &  $16$  & $ 0.17$  & $0.25$  & $ 0.18 $ & $1.4 $ & $ 0.1$ & $1.5 $ & $0.26 $ \\
    VAE  &  $32$  & $ 0.10$ & $ 0.10 $  & $ 0.12 $ & $ 0.2$ & $0.6 $ & $ 0.2$ & $0.26 $ \\
    \midrule
    InfoNCE &  $8$  & $ 0.22 $&  $ 0.29 $ &  $ 0.23 $ & $ 7.7$ & $1.3$ & $ 7.4$ & $ 0.18$ \\
    InfoNCE &  $16$  & $ 0.18$  & $0.28 $  & $ 0.20 $ & $ 1.9$ & $0.0 $ & $1.9 $ & $ 0.18$ \\
    InfoNCE &  $32$  & $ 0.20$ & $ 0.28 $  & $ 0.21  $ & $ 3.0$ & $1.0 $ & $ 5.8$ & $0.18 $ \\ 
    \midrule
    \modelabbr\ (Ours) &  $8$  & $0.51 $&  $0.51 $ &  $0.51 $ & $17.5 $ & $1.6 $ & $ 16.1$ & $ 0.30$ \\
    \modelabbr\ (Ours) &  $16$  & $0.47$  & $0.49$  & $0.49$ & $15.9 $ & $ 1.5$ & $14.3 $ & $ 0.34$ \\
    \modelabbr\ (Ours) &  $32$  & $0.50$ & $ 0.50$  & $ 0.50 $  & $ 18.4$ & $ 1.4$ & $17.1 $ & $0.32$ \\
    \bottomrule
  \end{tabular}
  \vspace{5pt}
  \caption{BERTScore \citep{zhang2019bertscore}, ROUGE, and BLEURT \citep{sellam2020bleurt} on ground truth infilled sentence and the generated sentence.}
  \label{tab:bertscore}
\end{table}

\section{Example trajectories of the learned latent space}
\begin{figure}
    \centering
    \newcommand{\gw}{60mm}
    \newcommand{\plotr}{0.45}
    \begin{subfigure}[b]{\plotr\columnwidth}
      \centering
        \includegraphics[width=\gw]{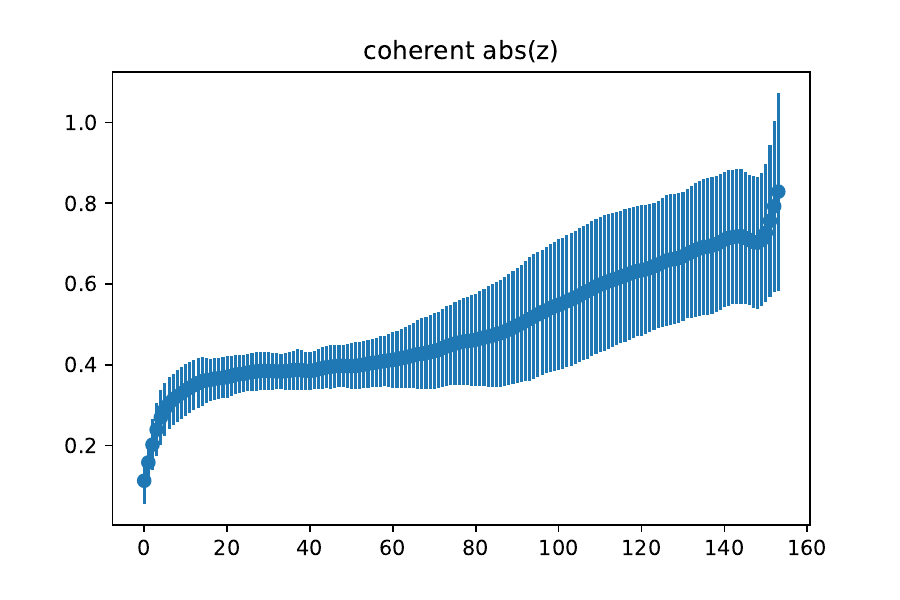}
        \caption{Coherent, \model\ d=8.}
    \end{subfigure}
    \begin{subfigure}[b]{\plotr\columnwidth}
      \centering
    \vspace{1em}
        \hspace{-2em}
        \includegraphics[width=\gw]{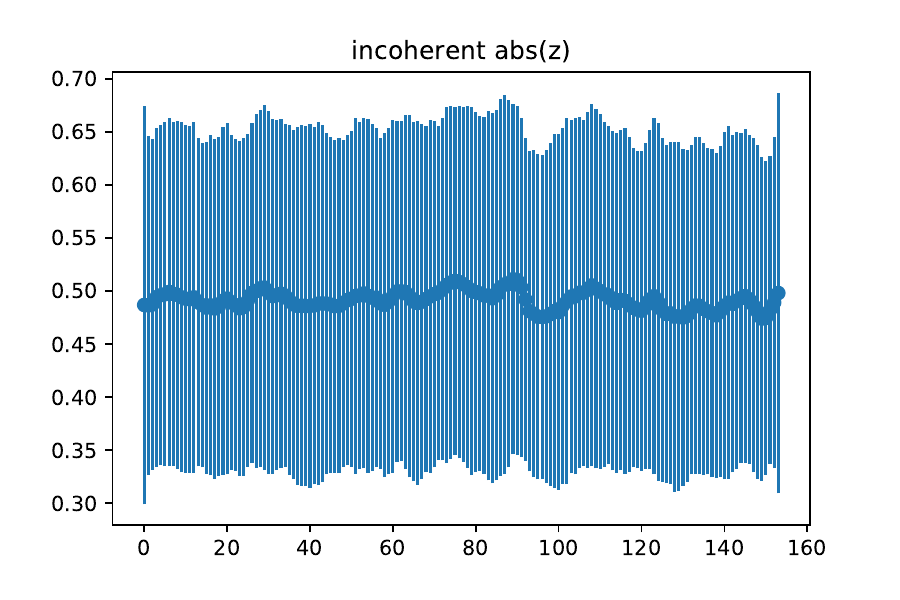}
        \caption{Incoherent, \model\ d=8.}
    \end{subfigure}
    \begin{subfigure}[b]{\plotr\columnwidth}
      \centering
        \includegraphics[width=\gw]{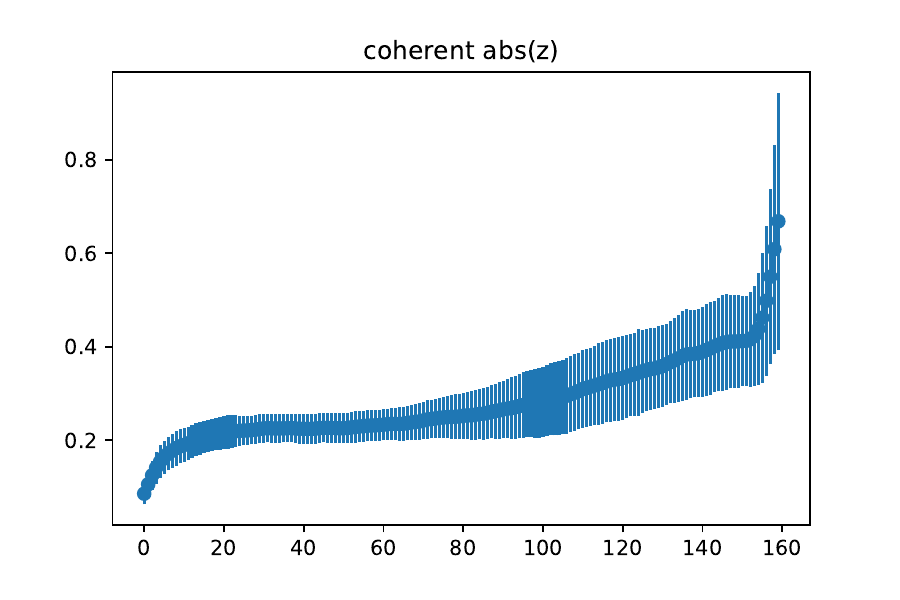}
        \caption{Coherent, \model\ d=16.}
    \end{subfigure}
    \begin{subfigure}[b]{\plotr\columnwidth}
      \centering
    \vspace{1em}
        \hspace{-2em}
        \includegraphics[width=\gw]{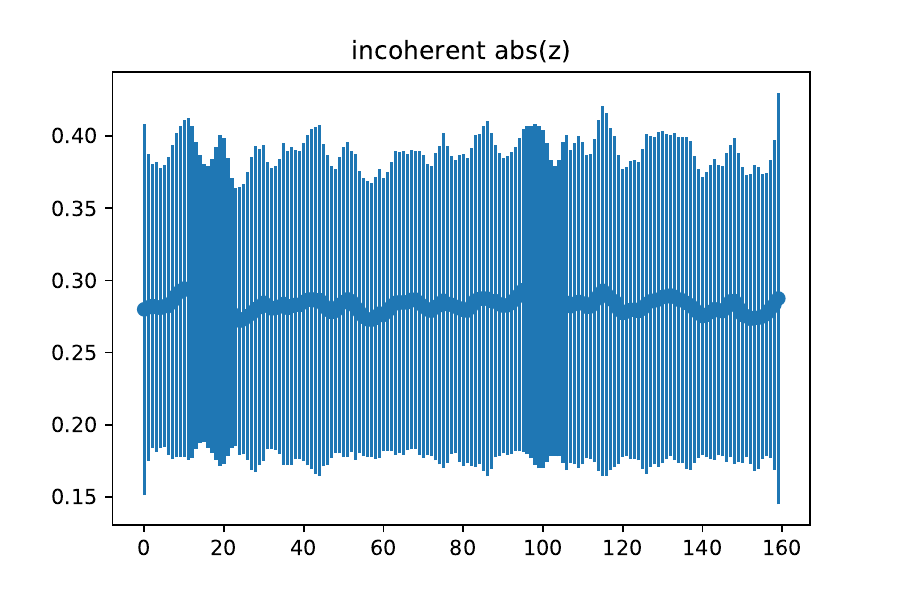}
        \caption{Incoherent, \model\ d=16.}
    \end{subfigure}
    \begin{subfigure}[b]{\plotr\columnwidth}
      \centering
        \includegraphics[width=\gw]{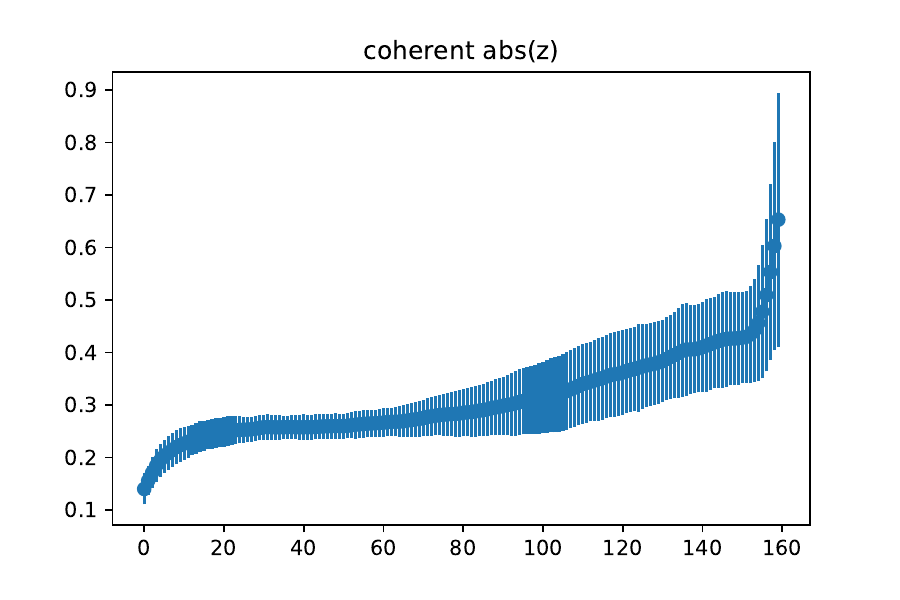}
        \caption{Coherent, \model\ d=32.}
    \end{subfigure}
    \begin{subfigure}[b]{\plotr\columnwidth}
      \centering
    \vspace{1em}
        \hspace{-2em}
        \includegraphics[width=\gw]{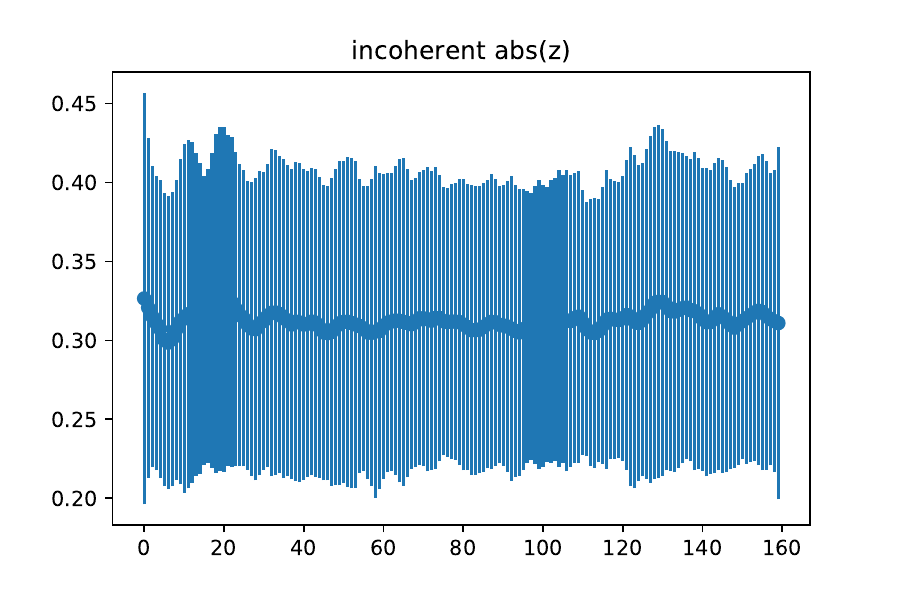}
        \caption{Incoherent, \model\ d=32.}
    \end{subfigure}
    \caption{\model's latent trajectories over coherent vs. incoherent (randomly scrambled) held-out Wikisection documents. The encoder learns Brownian bridge-like latent trajectories over the coherent documents. The incoherent documents map to noisy trajectories that don't evolve over time.} \label{fig:example_wikisection_latent_trajectories_tc}
    \vspace{-0.5em}
    
\end{figure}

\begin{figure}
    \centering
    \newcommand{\gw}{60mm}
    \newcommand{\plotr}{0.45}
    \begin{subfigure}[b]{\plotr\columnwidth}
      \centering
        \includegraphics[width=\gw]{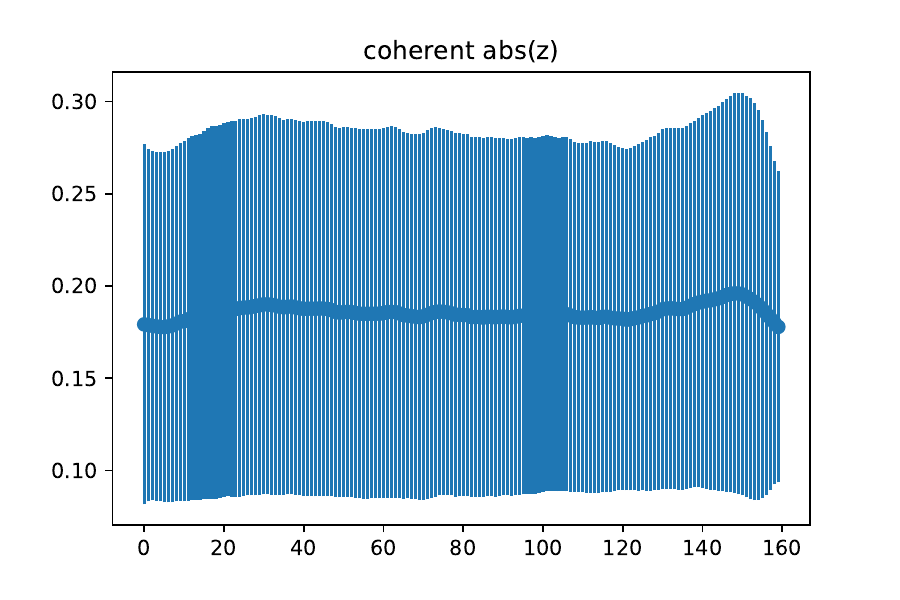}
        \caption{Coherent, \infonce\ d=8.}
    \end{subfigure}
    \begin{subfigure}[b]{\plotr\columnwidth}
      \centering
    \vspace{1em}
        \hspace{-2em}
        \includegraphics[width=\gw]{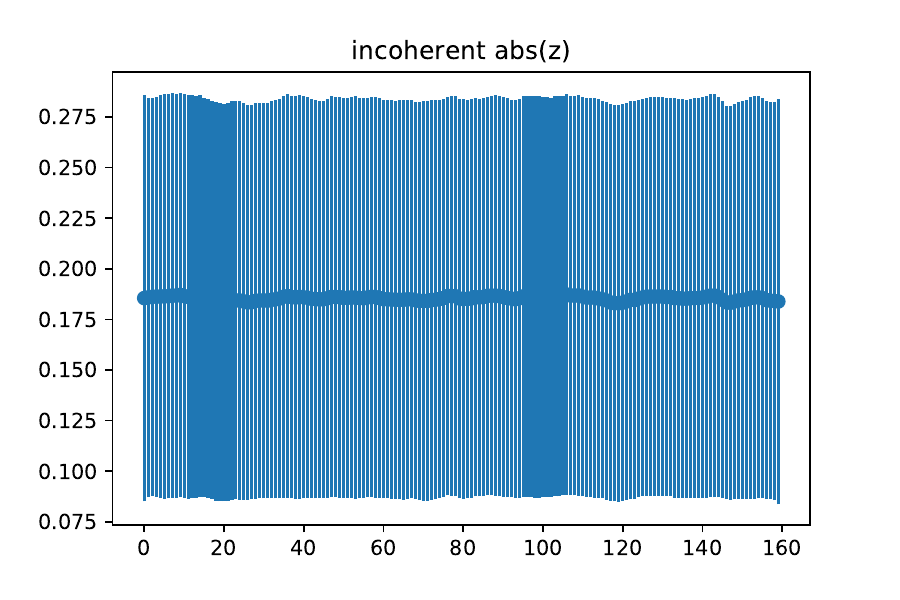}
        \caption{Incoherent, \infonce\ d=8.}
    \end{subfigure}
    \begin{subfigure}[b]{\plotr\columnwidth}
      \centering
        \includegraphics[width=\gw]{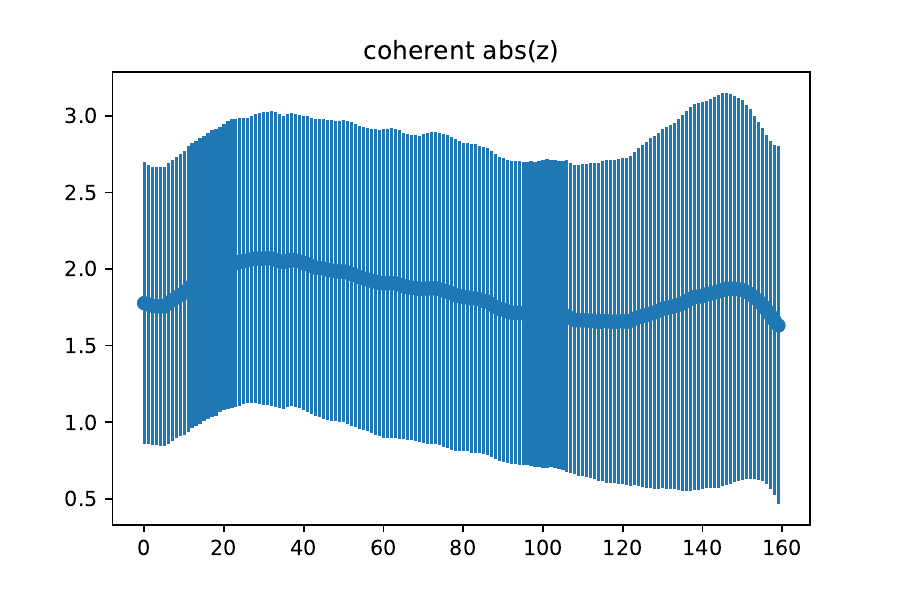}
        \caption{Coherent, \infonce\ d=16.}
    \end{subfigure}
    \begin{subfigure}[b]{\plotr\columnwidth}
      \centering
    \vspace{1em}
        \hspace{-2em}
        \includegraphics[width=\gw]{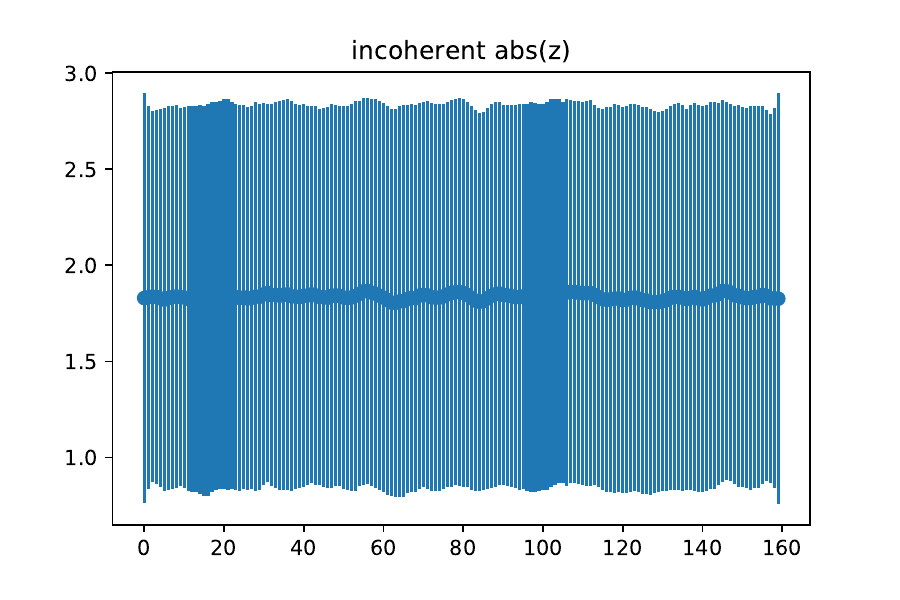}
        \caption{Incoherent, \infonce\ d=16.}
    \end{subfigure}
    \begin{subfigure}[b]{\plotr\columnwidth}
      \centering
        \includegraphics[width=\gw]{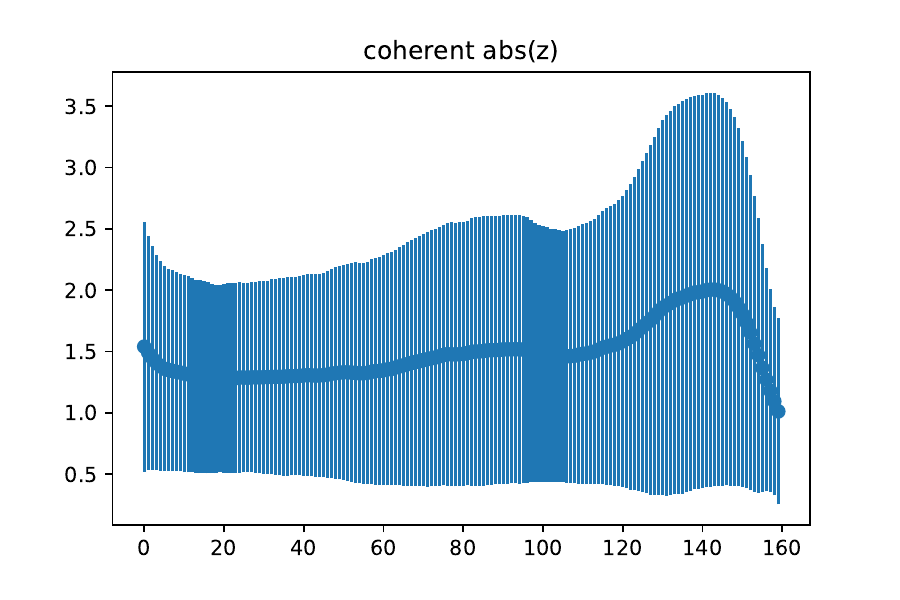}
        \caption{Coherent, \infonce\ d=32.}
    \end{subfigure}
    \begin{subfigure}[b]{\plotr\columnwidth}
      \centering
    \vspace{1em}
        \hspace{-2em}
        \includegraphics[width=\gw]{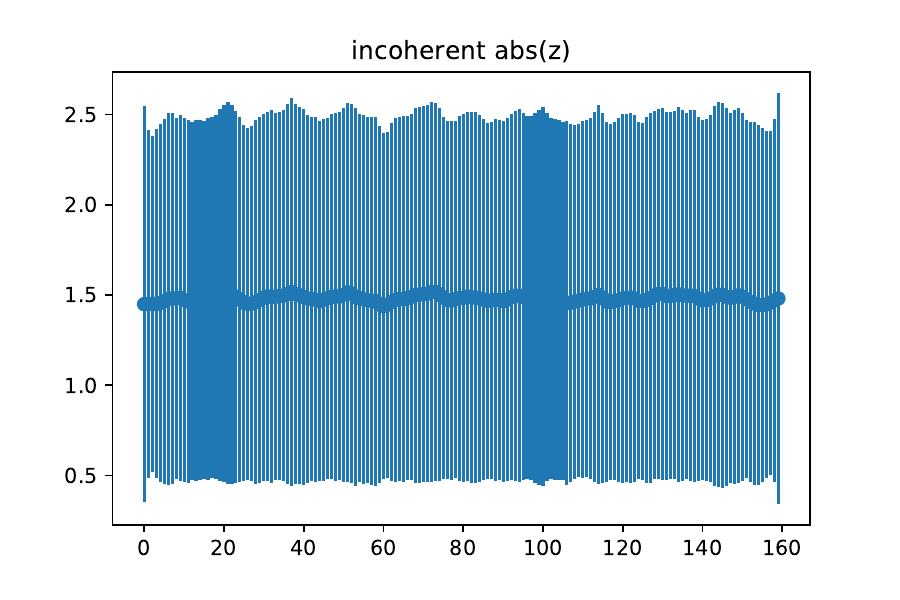}
        \caption{Incoherent, \infonce\ d=32.}
    \end{subfigure}
    \caption{\infonce's latent trajectories over coherent vs. incoherent (randomly scrambled) held-out Wikisection documents.} \label{fig:example_wikisection_latent_trajectories_infonce}
    \vspace{-0.5em}
\end{figure}

\begin{figure}
    \centering
    \newcommand{\gw}{60mm}
    \newcommand{\plotr}{0.45}
    \begin{subfigure}[b]{\plotr\columnwidth}
      \centering
        \includegraphics[width=\gw]{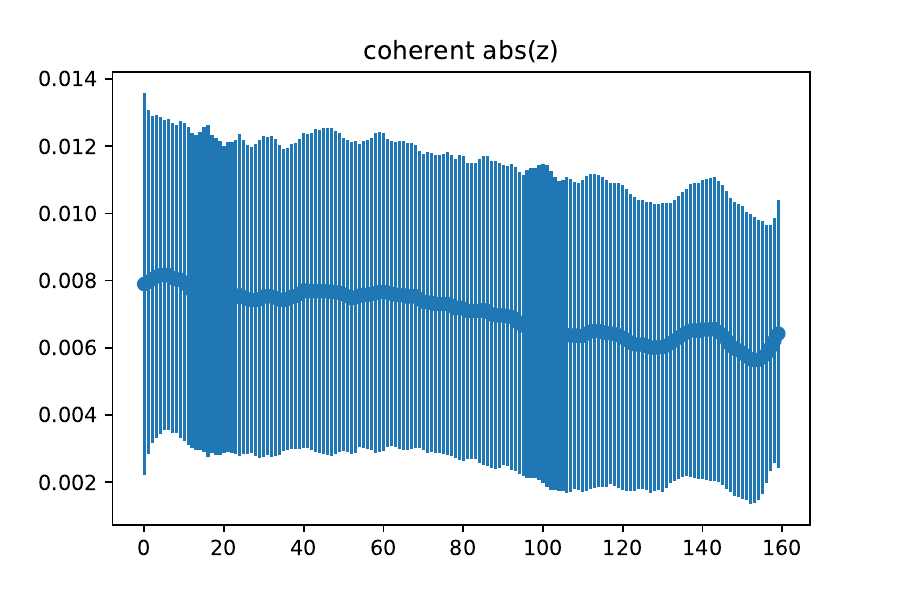}
        \caption{Coherent, \brownian\ d=8.}
    \end{subfigure}
    \begin{subfigure}[b]{\plotr\columnwidth}
      \centering
    \vspace{1em}
        \hspace{-2em}
        \includegraphics[width=\gw]{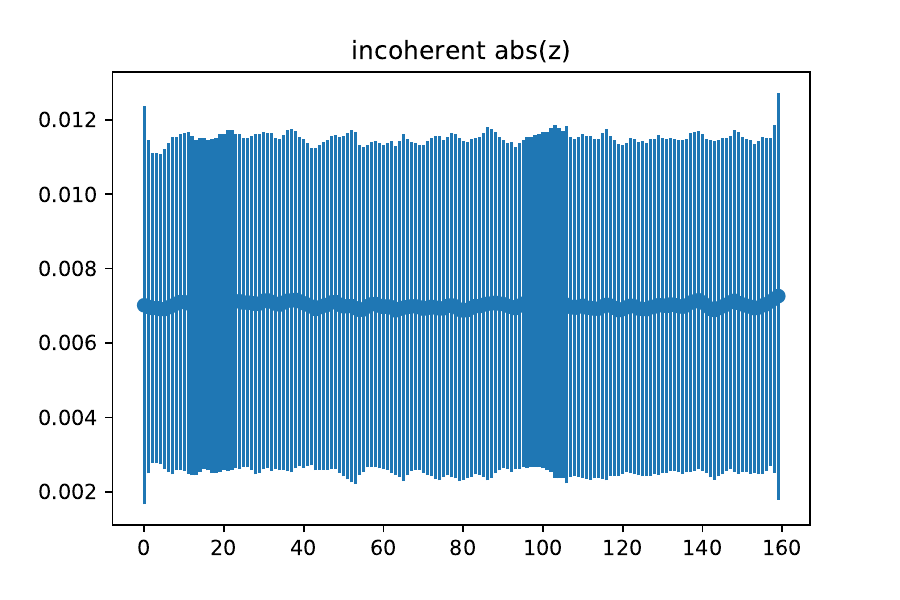}
        \caption{Incoherent, \brownian\ d=8.}
    \end{subfigure}
    \begin{subfigure}[b]{\plotr\columnwidth}
      \centering
        \includegraphics[width=\gw]{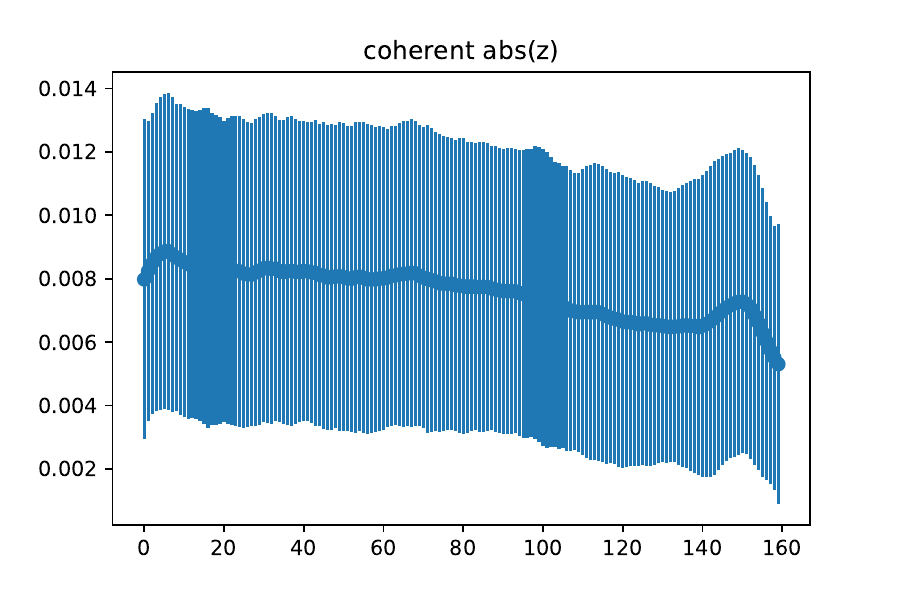}
        \caption{Coherent, \brownian\ d=16.}
    \end{subfigure}
    \begin{subfigure}[b]{\plotr\columnwidth}
      \centering
    \vspace{1em}
        \hspace{-2em}
        \includegraphics[width=\gw]{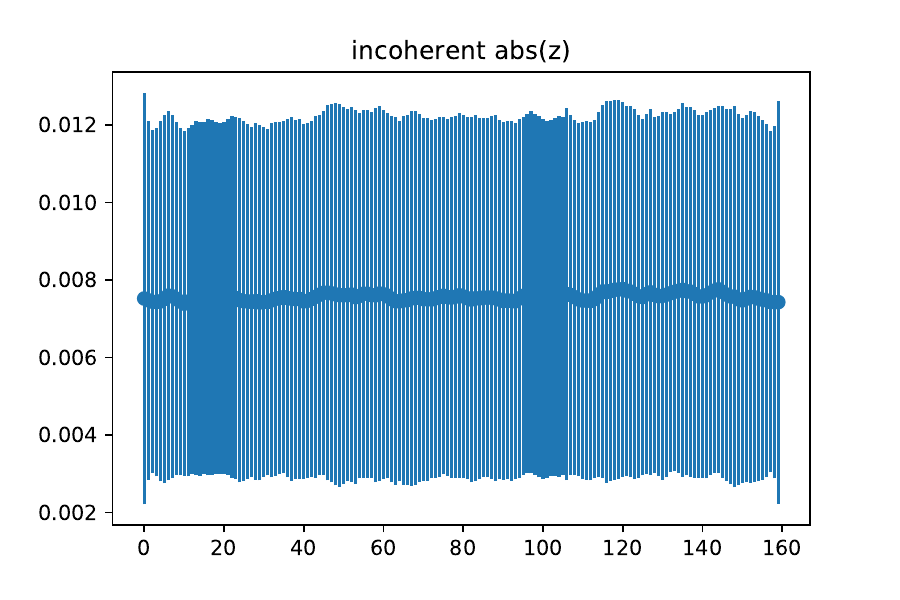}
        \caption{Incoherent, \brownian\ d=16.}
    \end{subfigure}
    \begin{subfigure}[b]{\plotr\columnwidth}
      \centering
        \includegraphics[width=\gw]{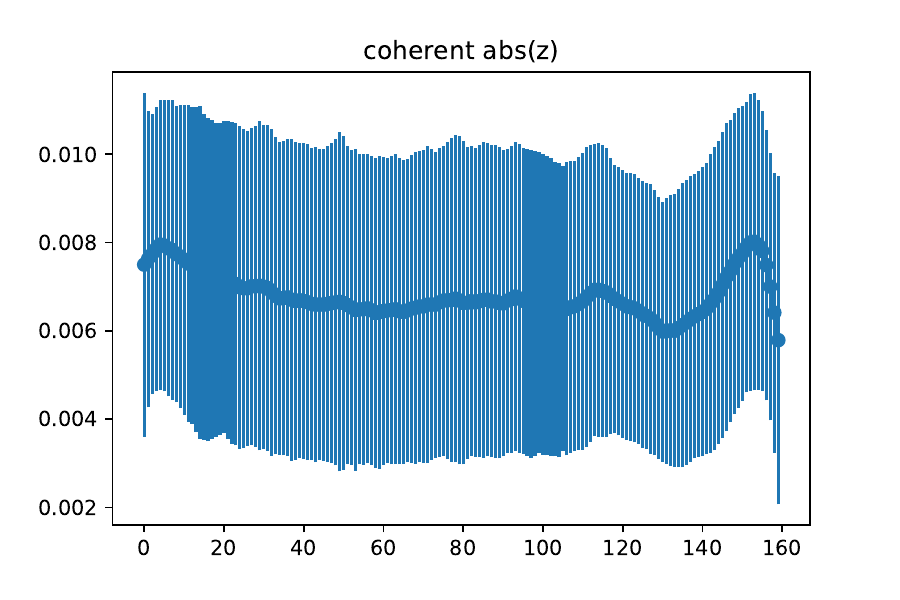}
        \caption{Coherent, \brownian\ d=32.}
    \end{subfigure}
    \begin{subfigure}[b]{\plotr\columnwidth}
      \centering
    \vspace{1em}
        \hspace{-2em}
        \includegraphics[width=\gw]{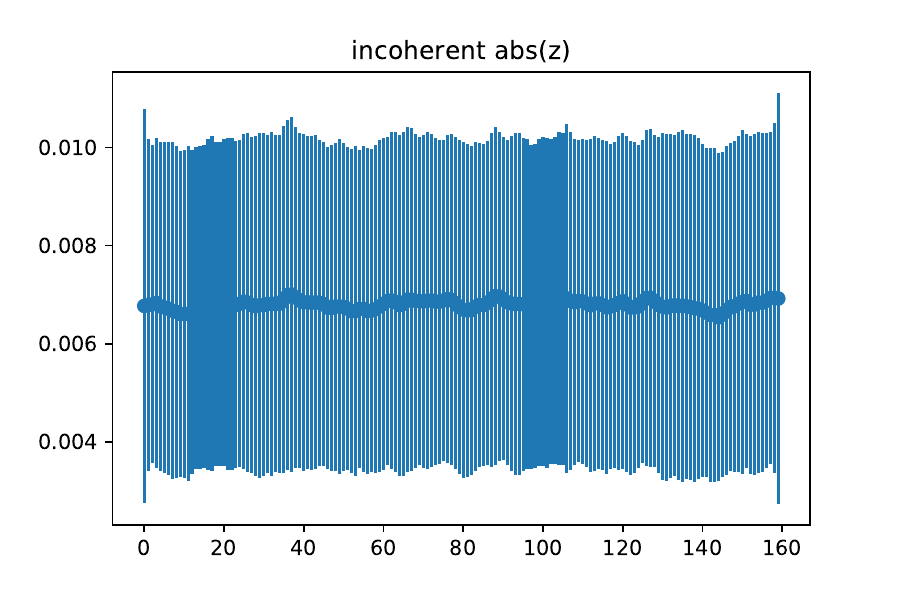}
        \caption{Incoherent, \brownian\ d=32.}
    \end{subfigure}
    \caption{\brownian's latent trajectories over coherent vs. incoherent (randomly scrambled) held-out Wikisection documents.} \label{fig:example_wikisection_latent_trajectories_brownian}
    \vspace{-0.5em}
\end{figure}

\begin{figure}
    \centering
    \newcommand{\gw}{60mm}
    \newcommand{\plotr}{0.45}
    \begin{subfigure}[b]{\plotr\columnwidth}
      \centering
        \includegraphics[width=\gw]{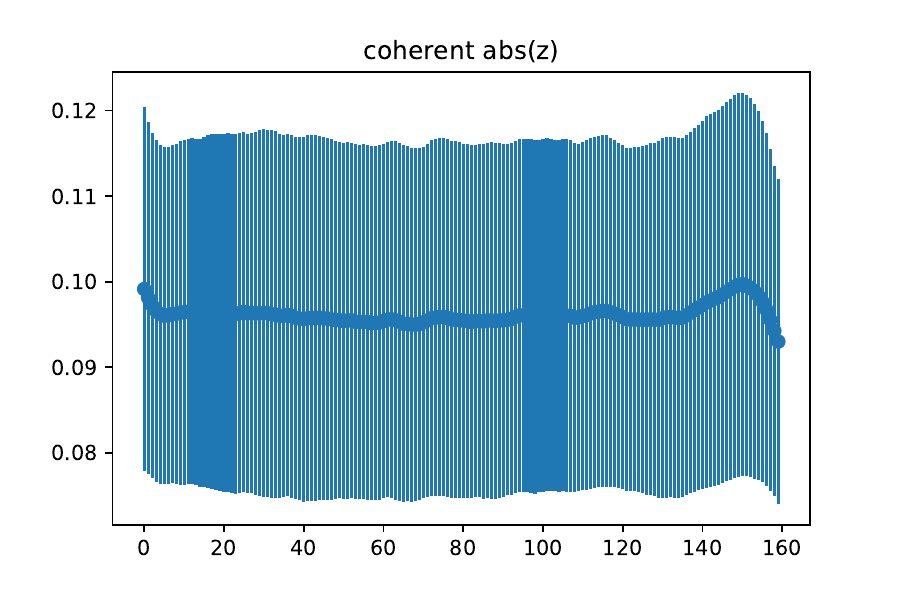}
        \caption{Coherent, \vae\ d=8.}
    \end{subfigure}
    \begin{subfigure}[b]{\plotr\columnwidth}
      \centering
    \vspace{1em}
        \hspace{-2em}
        \includegraphics[width=\gw]{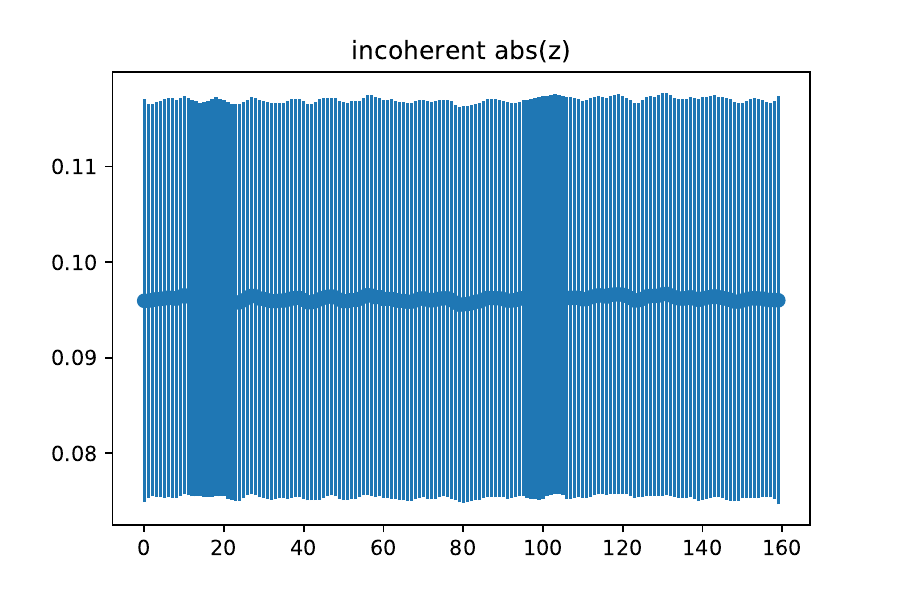}
        \caption{Incoherent, \vae\ d=8.}
    \end{subfigure}
    \begin{subfigure}[b]{\plotr\columnwidth}
      \centering
        \includegraphics[width=\gw]{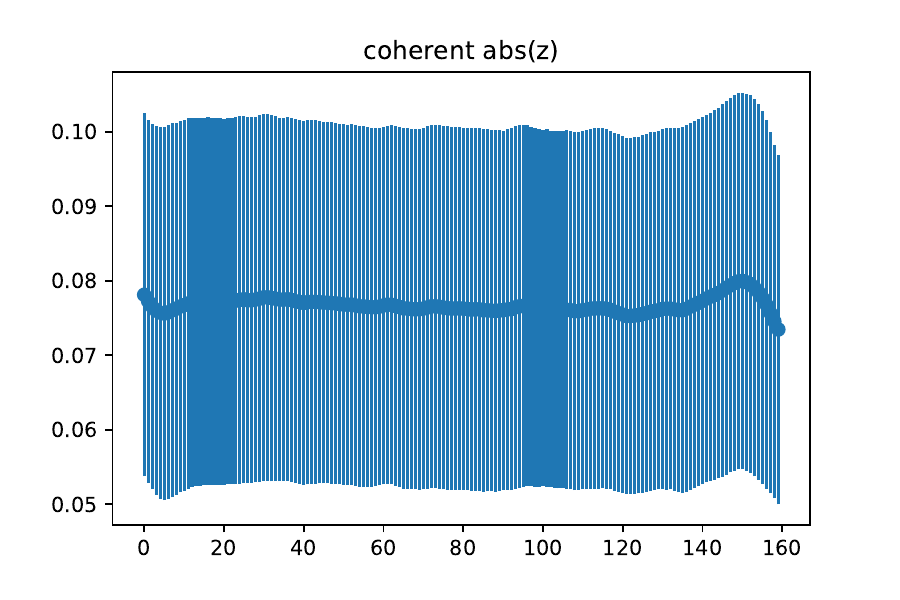}
        \caption{Coherent, \vae\ d=16.}
    \end{subfigure}
    \begin{subfigure}[b]{\plotr\columnwidth}
      \centering
    \vspace{1em}
        \hspace{-2em}
        \includegraphics[width=\gw]{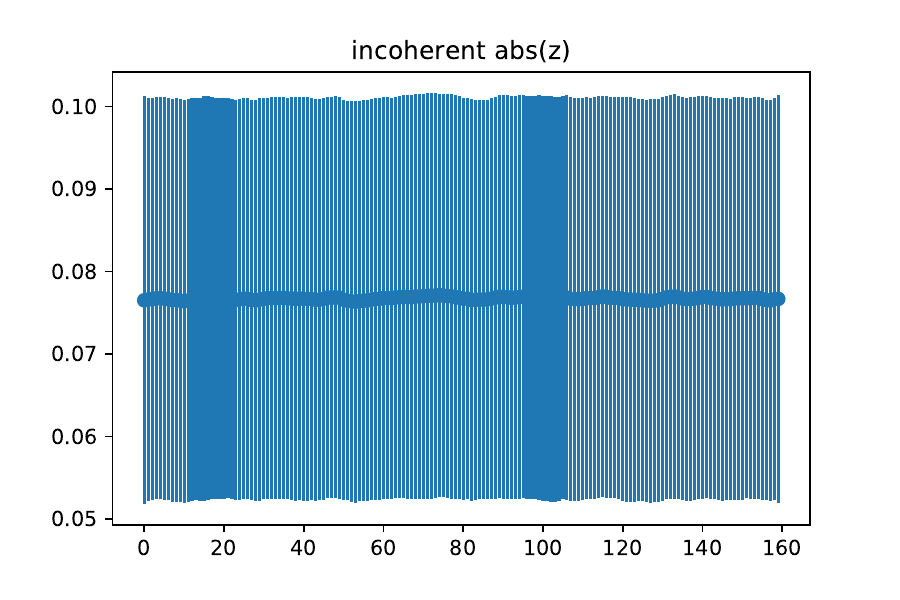}
        \caption{Incoherent, \vae\ d=16.}
    \end{subfigure}
    \begin{subfigure}[b]{\plotr\columnwidth}
      \centering
        \includegraphics[width=\gw]{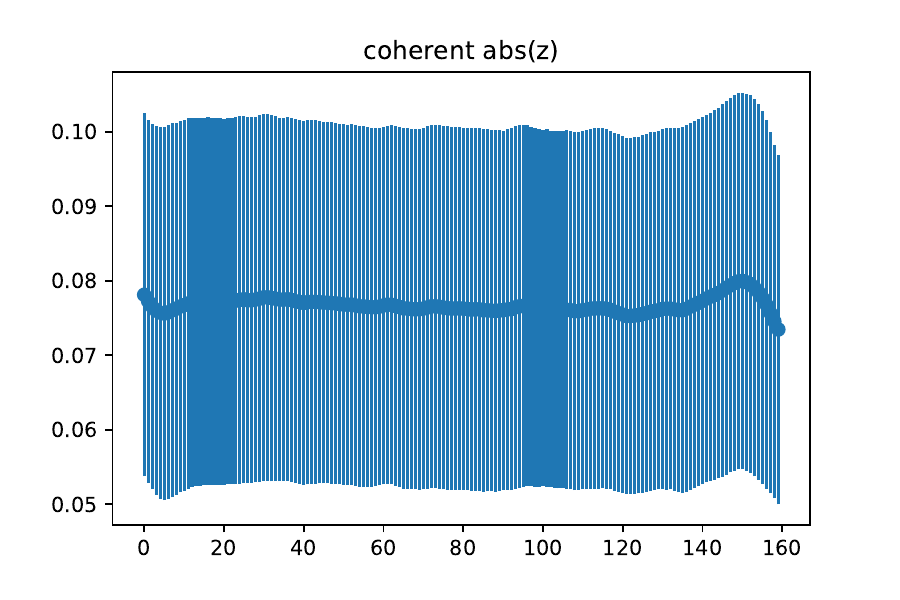}
        \caption{Coherent, \vae\ d=32.}
    \end{subfigure}
    \begin{subfigure}[b]{\plotr\columnwidth}
      \centering
    \vspace{1em}
        \hspace{-2em}
        \includegraphics[width=\gw]{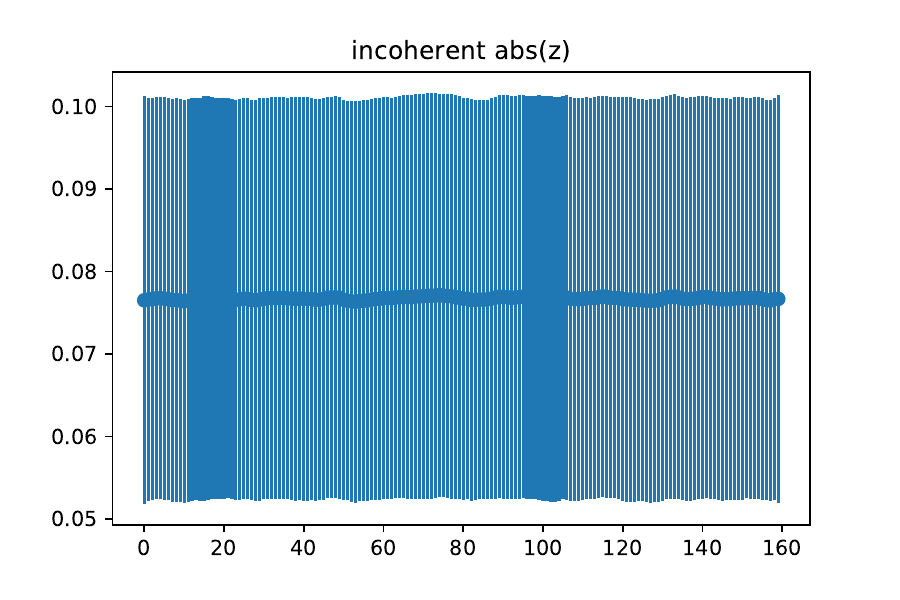}
        \caption{Incoherent, \vae\ d=32.}
    \end{subfigure}
    \caption{\vae's latent trajectories over coherent vs. incoherent (randomly scrambled) held-out Wikisection documents.} \label{fig:example_wikisection_latent_trajectories_vae}
    \vspace{-0.5em}
\end{figure}

See Figures~\ref{fig:example_wikisection_latent_trajectories_tc}-\ref{fig:example_wikisection_latent_trajectories_vae} for example latent trajectories over coherent vs. incoherent (randomly scrambled) held-out Wikisection documents. These trajectories show the recovered latent structure by \model\ and the three ablations from our work.
Figure~\ref{fig:example_wikisection_latent_trajectories_tc} are latent trajectories from \model. 
Figure~\ref{fig:example_wikisection_latent_trajectories_infonce} are latent trajectories from \infonce. 
Figure~\ref{fig:example_wikisection_latent_trajectories_brownian} are latent trajectories from \brownian.
Figure~\ref{fig:example_wikisection_latent_trajectories_vae} are latent trajectories from \vae.

\section{Why does \vae\ perform well on the Recipe domain? \label{sec:vae_good_on_recipe}}
In Table~\ref{tab:ordering}, \vaeabbr\ does much better than \modelabbr. Here we investigate potential reasons why. 

\begin{figure}
    \centering
    \newcommand{\gw}{60mm}
    \newcommand{\plotr}{0.45}
    \begin{subfigure}[b]{\plotr\columnwidth}
      \centering
        \includegraphics[width=\gw]{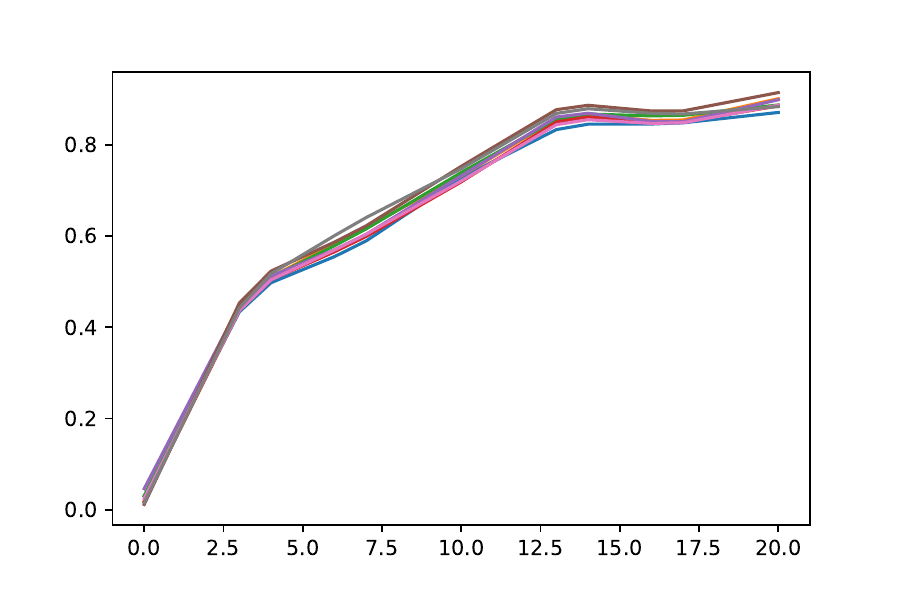}
        \caption{Latent trajectories across dimensions for \modelabbr\ (d=8).}
    \end{subfigure}
    \begin{subfigure}[b]{\plotr\columnwidth}
      \centering
    \vspace{1em}
        \hspace{-2em}
        \includegraphics[width=\gw]{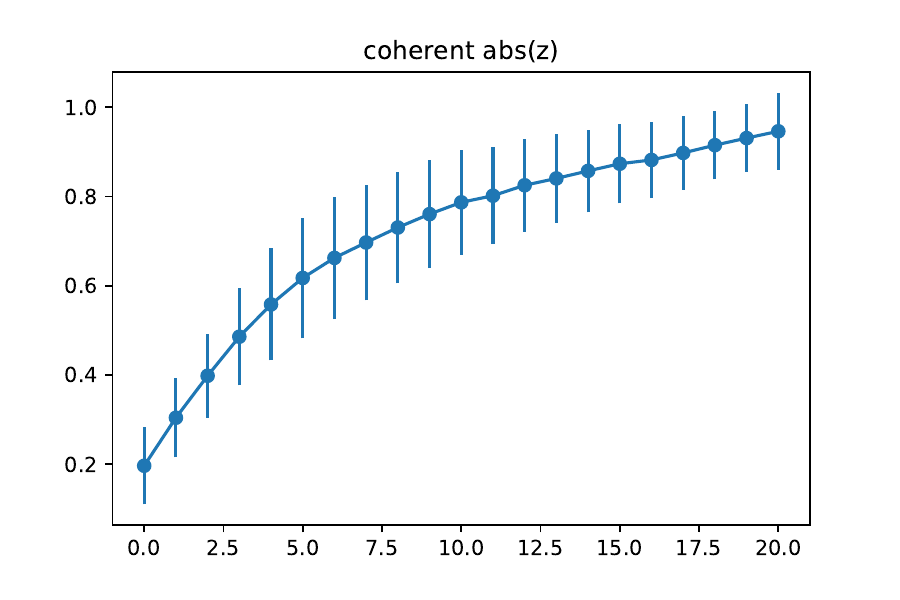}
        \caption{$\mu \pm \sigma$ in latent trajectories across dimensions for \modelabbr\ (d=8).}
    \end{subfigure}
    \begin{subfigure}[b]{\plotr\columnwidth}
      \centering
        \includegraphics[width=\gw]{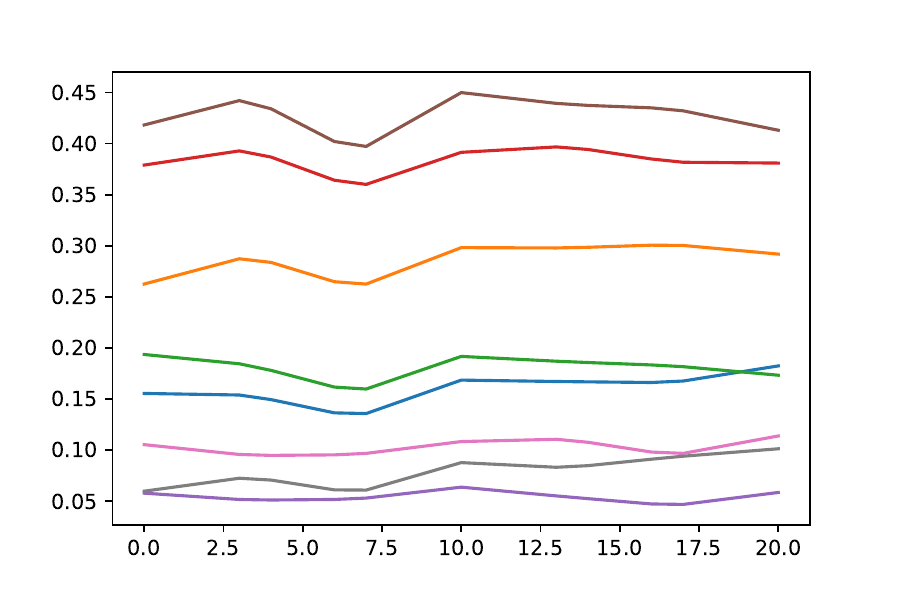}
        \caption{Latent trajectories across dimensions for \vaeabbr\ (d=8).}
    \end{subfigure}
    \begin{subfigure}[b]{\plotr\columnwidth}
      \centering
    \vspace{1em}
        \hspace{-2em}
        \includegraphics[width=\gw]{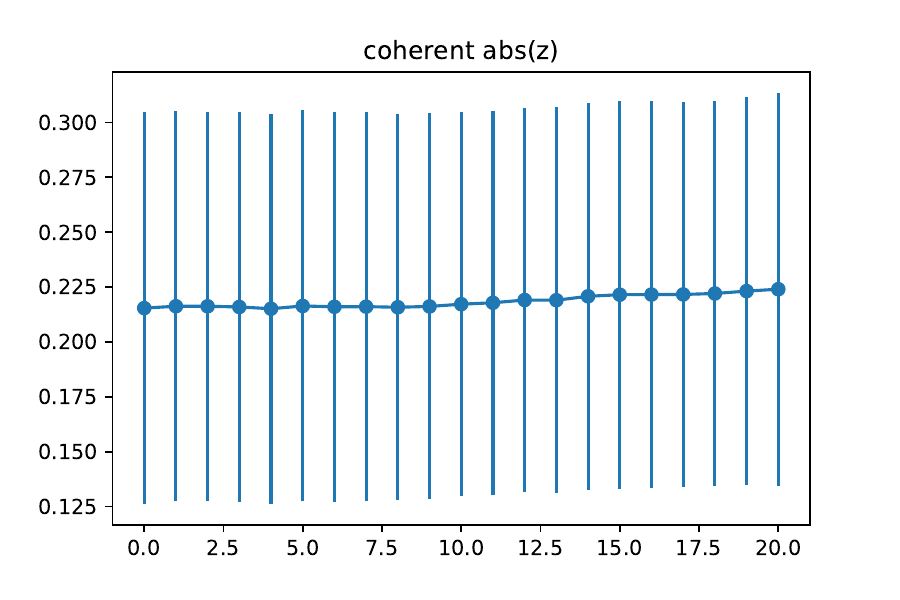}
        \caption{$\mu \pm \sigma$ in latent trajectories across dimensions for \vaeabbr\ (d=8).}
    \end{subfigure}
    \caption{Comparing the latent structure recovered by the \vaeabbr baseline and \modelabbr\ on held-out Recipe documents. Noticeably, \modelabbr\ learns temporally relevant activations whereas \vaeabbr\ latent does not correlate with time. \label{fig:why_is_vae_good_latents}}
    \vspace{-0.5em}
    
\end{figure}

We looked into comparing the latent structure recovered by the \vaeabbr\ baseline and \modelabbr; see Figure~\ref{fig:why_is_vae_good_latents}. What we found is the Time Control is best at extracting time over the course of the document (ie. its latents recover a Bridge process correlated with time), whereas the VAE doesn’t elicit strong temporal structure. We hypothesize that temporal structure is not all you need to succeed in the Recipe domain.

\section{Additional experiments}
On discourse coherence, we ran ALBERT on $k=1$ since ALBERT is trained on this setting. We found that it got slightly above random chance, unlike the most of the baselines we compare against. The discourse accuracy is $50.1\pm 9.0$ on Wikisection, $55.8 \pm 4.0$ on TicketTalk and $60.8 \pm 2.8$ on TM-2.

\section{Previous exploratory threads }

\subsection{Success and failure modes in fine-tuning GPT2 \label{sec:gpt2_app}}
We found that fine-tuned GPT2 was able to replicate certain aspects of a document corpus, such as section header ordering (~92\% accurate). However, when going from document-level to section-level statistics, we noticed that GPT-2 seemed to be either undershooting or overshooting the section lengths. These results are reported in Table~\ref{tab:length_matching}.

\subsection{Investigating the importance of distances between sampled sentences \label{sec:exploratory_t}}
\begin{figure}
    \centering
    \newcommand{\gw}{60mm}
    \newcommand{\plotr}{0.45}
    \begin{subfigure}[b]{\plotr\columnwidth}
      \centering
    \vspace{1em}
        \hspace{-2em}
        \includegraphics[width=\gw]{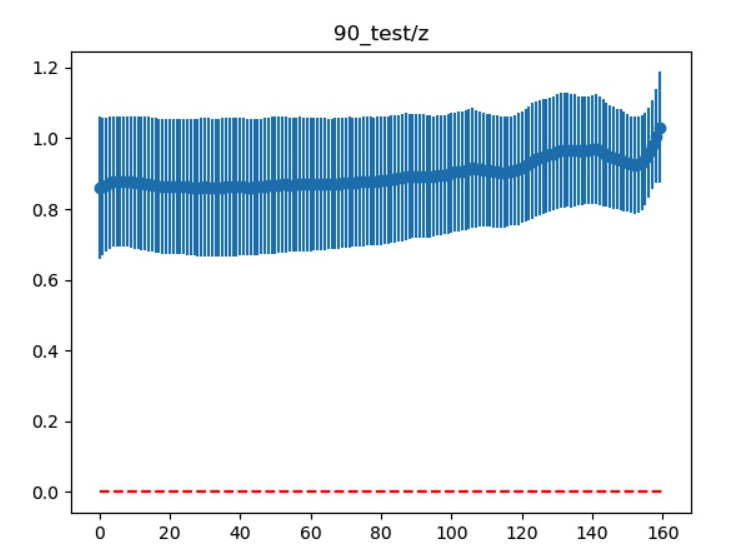}
        \caption{k=1}
    \end{subfigure}
    \begin{subfigure}[b]{\plotr\columnwidth}
      \centering
        \includegraphics[width=\gw]{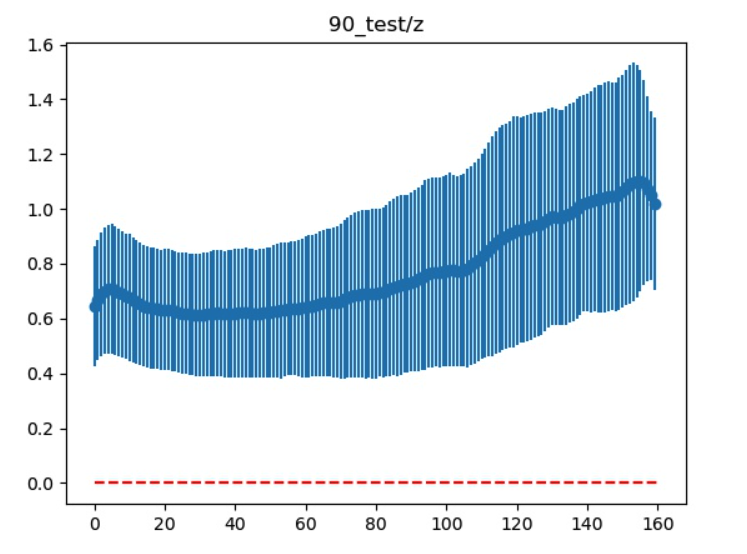}
        \caption{k=5}
    \end{subfigure}
    \begin{subfigure}[b]{\plotr\columnwidth}
      \centering
    \vspace{1em}
        \hspace{-2em}
        \includegraphics[width=\gw]{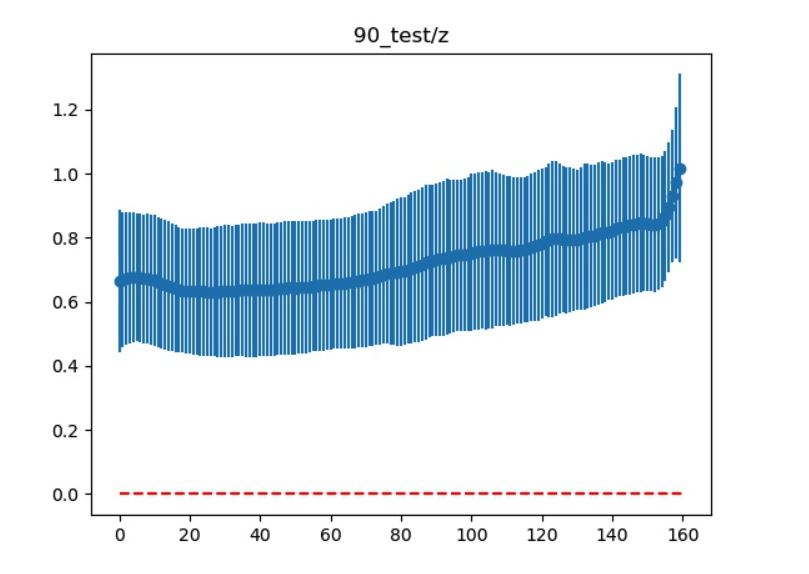}
        \caption{k=10}
    \end{subfigure}
    \caption{Recovered latent trajectories on held-out Wikisection documents where we used pair-wise contrasts with varying time distances; this was on \modelabbr, d = 8. Notice how the recovered latent structure varies depending on $k$, the distance between sampled sentences. \label{fig:exploratory_t}}
    \vspace{-0.5em}
\end{figure}

In earlier iterations of the work, we did explore learning with pairwise contrasts with fixed t-distances, e.g. distances of length 1, 5, and 10 sentences. We observed that the resulting latent trajectories elicited different fits to Brownian bridge dynamics, and the quality in fit varied in t across domains; we've include some examples in Figure~\ref{fig:exploratory_t} on Wikisection.

\end{document}

%% file: math_commands.tex
\usepackage{amsmath,amsfonts,bm}

\def\eqref#1{equation~\ref{#1}}

\def\1{\bm{1}}

\DeclareMathAlphabet{\mathsfit}{\encodingdefault}{\sfdefault}{m}{sl}
\SetMathAlphabet{\mathsfit}{bold}{\encodingdefault}{\sfdefault}{bx}{n}

